\documentclass[11pt]{article}

\usepackage[final]{acl}

\usepackage{times}
\usepackage{latexsym}
\usepackage[T1]{fontenc}
\usepackage[utf8]{inputenc}
\usepackage{microtype}
\usepackage{inconsolata}
\usepackage{graphicx}
\graphicspath{{./}}
\usepackage{booktabs}
\usepackage{multirow}
\usepackage{array}
\usepackage{makecell}
\usepackage{amsmath}
\usepackage{amssymb}
\usepackage{xcolor}
\usepackage{colortbl}
\usepackage{xspace}
\usepackage{enumitem}
\usepackage{adjustbox}
\usepackage{placeins}
\usepackage{listings}
\usepackage{tabularx}
\usepackage{makecell}
\usepackage{array}
\usepackage{booktabs}

\lstdefinestyle{promptstyle}{
  basicstyle=\ttfamily\scriptsize,
  breaklines=true,
  breakatwhitespace=true,
  breakindent=0pt,
  columns=fullflexible,
  keepspaces=true,
  showstringspaces=false,
  upquote=true,
  xleftmargin=0pt,
  xrightmargin=0pt,
  aboveskip=0.4em,
  belowskip=0.4em,
}
\newcommand{\framework}{EvalMem\xspace}
\newcommand{\datasetname}{DynaMem-Bench\xspace}
\newcommand{\memscan}{AgenticRAG\xspace}
\newcommand{\memwiki}{MemWiki\xspace}

\newcommand{\defectset}{\mathcal{D}}

\definecolor{hmlow}{HTML}{F7FBFF}
\definecolor{hmmid}{HTML}{6BAED6}
\definecolor{hmhigh}{HTML}{08306B}

\title{\framework: An Operation-Level Diagnostic Framework for \\ Long-Term Memory Systems}

\author{
\begin{tabular}{@{}ccccc@{}}
\textbf{Zeyu Liu\textsuperscript{1,*}} &
\textbf{Jian Zhong\textsuperscript{1,*}} &
\textbf{Rongduo Han\textsuperscript{1}} &
\textbf{Ziyang Wu\textsuperscript{1}} &
\textbf{Shunye Tang\textsuperscript{1}} \\
\textbf{Chenghao He\textsuperscript{1}} &
\textbf{Yaxuan Yang\textsuperscript{1}} &
\textbf{Yihang Qiu\textsuperscript{1}} &
\textbf{Ailing Wang\textsuperscript{2}} &
\textbf{Xiao Liang\textsuperscript{2}} \\
\textbf{Guohuan Xie\textsuperscript{3}} &
\textbf{Xiaokang Xue\textsuperscript{4}} &
\textbf{Gongchen Li\textsuperscript{1}} &
\textbf{Haining Zhang\textsuperscript{1,\textdagger}} &
\textbf{Wei Wang\textsuperscript{1,\textdagger}} \\
\multicolumn{5}{c}{
  \normalfont
  \textsuperscript{1}Nankai University
  \quad \textsuperscript{2}vivo AI Lab
} \\
\multicolumn{5}{c}{
  \normalfont
  \textsuperscript{3}Tsinghua University
  \quad \textsuperscript{4}Zhejiang University
} \\
\multicolumn{5}{c}{
  \small\normalfont
  Contact: \texttt{liuzeyu@mail.nankai.edu.cn}
} \\
\multicolumn{5}{c}{
  \small\normalfont
  \textsuperscript{*}Equal contribution;
  \textsuperscript{\textdagger}Corresponding authors;
  Code: \href{https://github.com/ZeyuuLiu/EvalMem}
  {github.com/ZeyuuLiu/EvalMem}
}
\end{tabular}
}

\begin{document}
\maketitle



\begin{abstract}
Long-horizon interactions with LLM-based assistants require memory systems that preserve and update user states, preferences, and interaction histories.
Existing evaluations report end-to-end QA accuracy and cannot determine whether errors arise from encoding, retrieval, or generation.
We introduce \textbf{\framework}, an operation-level diagnostic framework with three parallel Examiners.
For each query, the Encoding Examiner checks whether the target fact is stored, the Retrieval Examiner assesses whether the native retriever returns usable evidence, and the Generation Examiner tests whether the model can answer from oracle evidence.
Their outputs form fine-grained multi-label defect codes.
To improve store-level diagnosis, we adapt agentic RAG with a recall-first strategy that searches using both the query and source evidence, increasing recall of present evidence on \textsc{LoCoMo} from $70.2\%$ to $95.6\%$.
Evaluations of seven memory systems on \textsc{LoCoMo}, \textsc{LongMemEval}-S, and dynamic \datasetname identify retrieval as the most frequently attributed failure layer; in default \textsc{LoCoMo}, retrieval defects reach 22.1\%, compared with $7.7\%$ for encoding and $6.5\%$ for generation.
Guided by this diagnosis, \textbf{\memwiki}, a search-friendly auxiliary structure built from each system's memory export, improves mean accuracy by $2.5$ and $2.3$ percentage points on \textsc{LoCoMo} and \textsc{LongMemEval}-S, respectively.

\end{abstract}

\section{Introduction}
\label{sec:intro}

\begin{figure}[t]
  \centering
  \includegraphics[width=\columnwidth]{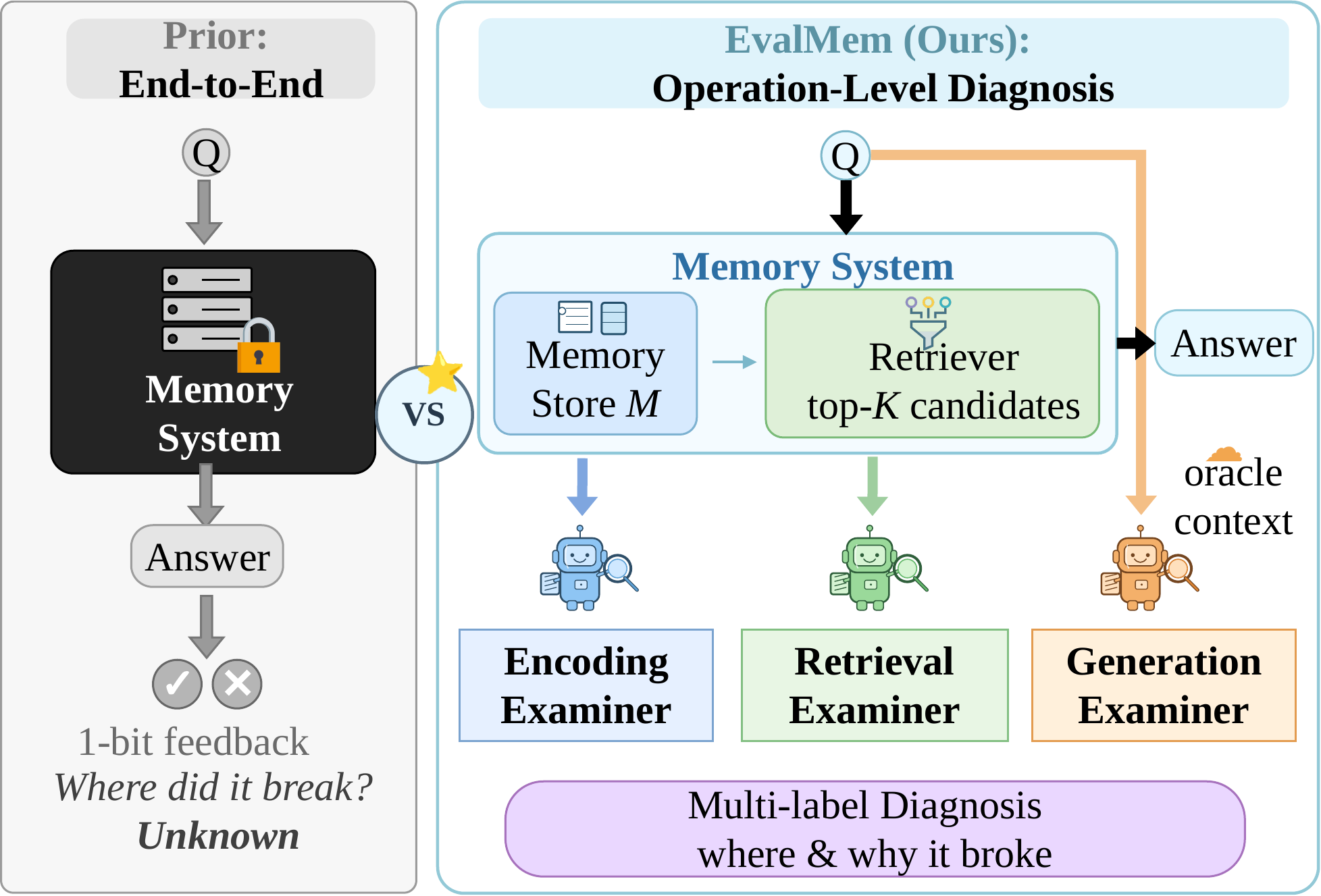}
  \caption{\framework{} at a glance. A query is examined by three parallel Examiners over Encoding, Retrieval, and Generation; their decisions are reconciled into a multi-label defect set instead of a single end-to-end score.}
  \label{fig:overview}
  \vspace{-4mm}
\end{figure}

Long-horizon interaction is becoming a central setting for LLM-based assistants, where personal assistants, tutors, health-support tools, and companion agents must maintain user states, preferences, plans, and past events across sessions~\citep{park2023generative, packer2023memgpt, wang2023voyager}.
Long-term memory systems support this capability by adding external storage and retrieval layers on top of LLMs~\citep{li2025memos,kang2025memory,wang2025mem}.
Although their implementations vary widely, from hierarchical or temporal summaries to operating-system-style memory management, modular memory blocks, user profiles, and agentically constructed records~\citep{kang2025memory,li2026timem,li2025memos,hu2026evermemos,tao2026membox,wang2025mem,yan2025general, xu2025mem}, they share a common loop: encode past interactions, retrieve relevant records for a new query, and generate an answer from retrieved evidence.
This shared loop makes unified attribution-oriented evaluation possible, yet current evaluation remains less informative than memory-system design demands~\citep{ding2026memground}.

Existing benchmarks have expanded long-term memory evaluation along several directions~\citep{liu2024lost}.
Static long-history benchmarks such as \textsc{LoCoMo} and \textsc{LongMemEval} test factual recall, temporal reasoning, multi-hop reasoning, knowledge updates, and abstention~\citep{maharana2024evaluating,wu2024longmemeval}; capability-oriented benchmarks examine hallucination, continual learning, memory operations, and broader memory competencies~\citep{chen2025halumem,ai2025memorybench,wang2026evomembench}; and interactive benchmarks such as \textsc{AMemGym} argue for on-policy evaluation, where the assistant participates in the interaction rather than being tested only on a fixed dialogue tape~\citep{jiayang2026amemgym}.
Despite this progress, most protocols still report end-to-end answer correctness: a wrong answer does not reveal whether the relevant fact was not stored, not retrieved, or not used~\citep{es2024ragas, lyu2025crud}.
Many datasets also rely on fixed histories, making it difficult to reflect how an assistant's own dialogue choices shape memory-writing trajectory~\citep{jiayang2026amemgym, liu2024agentbench}.
Failures caused by state evolution, stale records, or retrieval over the system's actual memory store thus remain difficult to diagnose.

Building a diagnostic evaluation framework is challenging.
It must abstract over heterogeneous storage formats, retrieval mechanisms, and generation pipelines; allow coupled failures across memory construction, retrieval, and answer generation; and reliably observe the memory store.
If the evaluator fails to find a fact that was actually stored, it may incorrectly attribute a retrieval error to memory construction~\citep{meng2022locating, manakul2023selfcheckgpt}.

To address these challenges, we propose \framework, an operation-level diagnostic framework summarized in Figure~\ref{fig:overview}.
\framework treats heterogeneous memory systems as an encoding--retrieval--generation pipeline and runs three independent Examiners in parallel for each query.
The Encoding Examiner checks whether the target fact is present in the memory store, the Retrieval Examiner inspects whether the system's native retriever returns usable evidence, and the Generation Examiner tests whether the model can answer from oracle evidence.
An Attribution Agent merges their decisions into a fine-grained multi-label defect set, allowing co-occurring failures while suppressing retrieval defects when the source fact is absent from the store.
To make the Encoding Examiner reliable when diagnostic retrieval is imperfect, we adapt agentic RAG with a recall-first principle: the probe searches the original memory store using both the query and its corresponding source evidence, reducing observation misses from $29.8\%$ to $4.4\%$ on \textsc{LoCoMo} samples with present evidence.
We further construct \datasetname for on-policy dynamic evaluation and introduce \memwiki, a search-friendly auxiliary structure built from each system's own memory export, to test whether diagnosis can guide system improvement.
Our contributions are threefold:

\noindent\textbf{(1) Fine-grained attribution for memory-system evaluation.}
We introduce \framework, which uses Encoding, Retrieval, and Generation Examiners to produce an $11$-code multi-label diagnosis.

\noindent\textbf{(2) Reliable and dynamic diagnostic evaluation.}
We adapt agentic RAG into a bounded recall-first diagnostic routine and construct \datasetname for on-policy state evolution.

\noindent\textbf{(3) Retrieval as the most frequently attributed failure layer.}
Across seven systems and three datasets, retrieval defects consistently exceed encoding and generation defects ($22.1\%$ vs.\ $7.7\%$/$6.5\%$ on \textsc{LoCoMo}, $18.3\%$ vs.\ $8.5\%$/$5.4\%$ on \textsc{LongMemEval}-S, and $25.9\%$ vs.\ $9.7\%$/$5.0\%$ on \datasetname); guided by this diagnosis, \memwiki improves mean accuracy by $+2.5$\,pp on \textsc{LoCoMo} and $+2.3$\,pp on \textsc{LongMemEval}-S.

\section{\framework}
\label{sec:framework}

\begin{figure*}[t]
  \centering
  \includegraphics[width=0.8\textwidth]{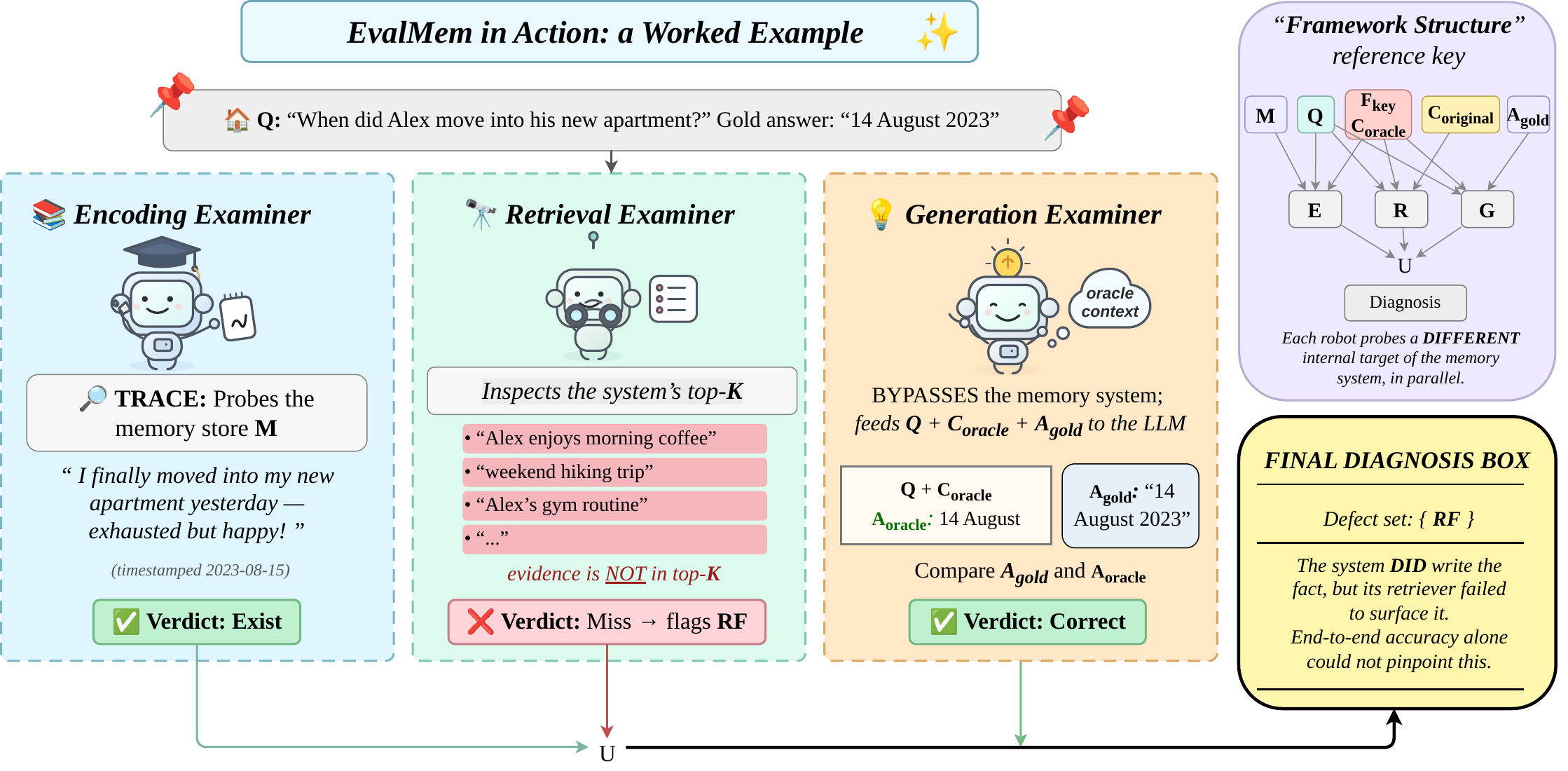}
  \caption{Overall view of \framework: the tested system runs natively, then Encoding, Retrieval, and Generation Examiners map task-aware states into a multi-label defect set.}
  \label{fig:framework}
  \vspace{-4mm}
\end{figure*}

\framework is an operation-level diagnostic framework for long-term memory systems.
As shown in Figure~\ref{fig:framework}, a tested system first runs natively: given dialogue history $H$ and query $Q$, it builds or updates memory store $\mathcal{M}$, retrieves $C_{\text{original}}$, and generates $\hat{A}$.
\framework then diagnoses this completed run with benchmark-side annotations: gold answer $A_{\text{gold}}$, task type $\tau$, oracle evidence $C_{\text{oracle}}$, and key facts $F_{\text{key}}$ distilled from that evidence.
We first define task types and inputs, then present the three Examiners and Attribution Agent, the recall-first agentic-RAG routine used inside Encoding diagnosis, \datasetname for on-policy dynamic evaluation, and \memwiki as a diagnosis-guided actionability check.

\subsection{Task Types and Diagnostic Inputs}
\label{sec:framework:basics}

Attribution is task-dependent, so we distinguish POS and NEG queries.
A \textbf{POS} question is answerable from the original dialogue, so the system should store the relevant fact, retrieve it, and use it to answer; it tests whether the system can \emph{remember and retrieve}.
A \textbf{NEG} question lacks sufficient supporting evidence, so the ideal behavior is refusal rather than a concrete answer; it tests whether the system can \emph{avoid unsupported answering}.

Each evaluated query is represented as
\[
x=(H,Q,A_{\text{gold}},\tau,F_{\text{key}},C_{\text{oracle}}).
\]
$H$ is the dialogue history available before $Q$.
$A_{\text{gold}}$ is the expected answer or refusal, and $\tau\in\{\text{POS},\text{NEG}\}$ denotes the task type.
$C_{\text{oracle}}$ is the clean, complete evidence context.
$F_{\text{key}}$ is the set of atomic evidence facts required by $Q$: for POS it supports $A_{\text{gold}}$; for NEG, no valid supporting fact exists, so it is empty or used only contrastively.
After processing $H$, the tested system exposes $\mathcal{M}$; when asked $Q$, it returns $C_{\text{original}}$ and $\hat{A}$.
EvalMem can be integrated when a system exposes (i) an inspectable memory bank, (ii) the context actually passed to the generation model, and (iii) its final answer. The benchmark must provide, for each test question, the corresponding evidence text, i.e., the dialogues from the original conversational dataset.

\subsection{Three Examiners and Attribution}
\label{sec:framework:examiners}

\framework runs three Examiners in parallel.
Each Examiner isolates one operation, returns a local state, and maps it to a task-aware defect subset.
The Attribution Agent reconciles these subsets into a multi-label diagnosis while suppressing downstream defects that merely follow from upstream failures.

\paragraph{Encoding Examiner $\mathcal{E}_{\text{enc}}$.}
Inputs: $(Q,\tau,F_{\text{key}},\mathcal{M})$.
The Encoding Examiner asks whether the information required by $Q$ is present in $\mathcal{M}$ under an oracle diagnostic view.
This judgement is independent of the system's native retrieval result.
Let $P$ denote $\tau=\text{POS}$ and $N$ denote $\tau=\text{NEG}$.
All existential predicates range over $\mathcal{M}$.
Predicates $\mathrm{Supp}$, $\mathrm{Ambig}$, and $\mathrm{Wrong}$ are evaluated against $F_{\text{key}}$ and indicate full support, unresolved-reference support, and wrong-value support.
$\mathrm{Dirty}(m,Q)$ indicates an unsupported pseudo-fact relevant to a NEG query.
The encoding state is
\begin{equation}
  \small
  S_{\text{enc}} =
  \begin{cases}
    \textsc{Exist}, & P\wedge \exists m:\mathrm{Supp}(m),\\
    \textsc{Dirty}, & N\wedge \exists m:\mathrm{Dirty}(m,Q),\\
    \textsc{Corrupt}_{\text{Ambig}}, & P\wedge \exists m:\mathrm{Ambig}(m),\\
    \textsc{Corrupt}_{\text{Wrong}}, & P\wedge \exists m:\mathrm{Wrong}(m),\\
    \textsc{Miss}, & \text{otherwise}.
  \end{cases}
  \label{eq:enc_state}
\end{equation}
For POS, \textsc{Exist} means the key fact is stored correctly; \textsc{Miss} means no corresponding record is found; and the two \textsc{Corrupt} states mean a related key is present but unusable.
For NEG questions, \textsc{Miss} is the healthy state, while \textsc{Dirty} means the store contains an unsupported pseudo-fact.

\paragraph{Retrieval Examiner $\mathcal{E}_{\text{ret}}$.}
Inputs: $(Q,\tau,F_{\text{key}},C_{\text{original}})$.
The Retrieval Examiner asks whether the native retrieval step returns the key information in an effective context.
Write $C=C_{\text{original}}$ in the following equations.
For POS questions, $F_{\text{key}}\preceq C$ means that all key facts are semantically supported by the retrieved context.
For NEG, $\mathrm{Mislead}_{\tau_{\text{H}}}(c,Q)$ marks a relevant-looking but insufficient, stale, or unsupported record with similarity above $\tau_{\text{H}}$.
The retrieval state is
\begin{equation}
  \small
  S_{\text{ret}} =
  \begin{cases}
    \textsc{Hit},   & P\wedge F_{\text{key}}\preceq C, \\
    \textsc{Noise}, & N\wedge \exists c \in C:
      \mathrm{Mislead}_{\tau_{\text{H}}}(c,Q), \\
    \textsc{Miss},  & \text{otherwise}.
  \end{cases}
  \label{eq:ret_state}
\end{equation}
For POS, \textsc{Hit} means key information appears in the retrieved top-$K$ context; \textsc{Miss} means it does not.
For NEG, \textsc{Noise} means retrieval returns a high-risk misleading record when no valid support exists.
For POS hits, retrieval quality is further measured by the best evidence rank $r(F_{\text{key}})$ and the signal-to-noise ratio
\begin{equation}
  \mathrm{SNR}(C_{\text{original}})=
  \frac{|\mathrm{evid}(F_{\text{key}})\cap C_{\text{original}}|_{\text{tok}}}
       {|C_{\text{original}}|_{\text{tok}}}.
  \label{eq:snr}
\end{equation}
A \textsc{Hit} is still retrieval-defective if the evidence is ranked too late ($r(F_{\text{key}})>\tau_{\text{rank}}$) or diluted by noise ($\mathrm{SNR}<\tau_{\text{snr}}$).
Conversely, a POS \textsc{Miss} is counted as \textsc{RF} only when the Encoding Examiner confirms that the fact exists in $\mathcal{M}$; otherwise it is treated as upstream storage failure.

\paragraph{Generation Examiner $\mathcal{E}_{\text{gen}}$.}
Inputs: $(Q,\tau,C_{\text{oracle}})$.
The Generation Examiner asks whether the answer model can solve the query when memory and retrieval are no longer bottlenecks.
It feeds clean oracle context $C_{\text{oracle}}$ and compares the oracle-path answer with $A_{\text{gold}}$:
Let $\hat{A}_{o}=g(Q,C_{\text{oracle}})$ and $J=\mathrm{Judge}$.
\begin{equation}
  S_{\text{gen}} =
  \begin{cases}
    \textsc{Pass}, & J(\hat{A}_{o},A_{\text{gold}},\tau)=1,\\
    \textsc{Fail}, & \text{otherwise}.
  \end{cases}
  \label{eq:gen_state}
\end{equation}
For POS questions, \textsc{Pass} means that the model can answer when the complete evidence is given; for NEG questions, it means that the model correctly refuses without support.
When $S_{\text{gen}}=\textsc{Fail}$, a subclassifier assigns the code: for POS questions, \textsc{GF} (Generation Faithfulness) when the model ignores the oracle evidence, and \textsc{GRF} (Generation Reasoning Failure) when the model reads the evidence but reasons incorrectly; for NEG questions, \textsc{GH} (Generation Hallucination) when the model fabricates an answer instead of refusing.

\paragraph{Multi-label union attribution.}
The final attribution is not a single root cause.
For sample $i$, the Attribution Agent outputs
\begin{equation}
  \defectset_{\text{total}}^{(i)}
  =
  \defectset_{\text{enc}}
  \cup
  \defectset_{\text{ret}}
  \cup
  \defectset_{\text{gen}}.
  \label{eq:union}
\end{equation}
The healthy path is \textsc{Exist}/\textsc{Hit}/\textsc{Pass} for POS and \textsc{Miss}/\textsc{Miss}/\textsc{Pass} for NEG.
The Attribution Agent also applies an upstream masking rule: if a POS fact is absent from the store, retrieval \textsc{Miss} is not counted as \textsc{RF}; for NEG, \textsc{Noise} is counted as \textsc{NIR} when the store is clean, while a retrieved pseudo-fact already written into the store is attributed to \textsc{DMP}.
The $11$ defect-code definitions are listed in Table~\ref{tab:defect_codes} and Appendix~\ref{app:defect_codes}.

For reported defect statistics, we apply an \emph{error gate}: only incorrectly answered samples contribute defect codes, and every rate takes the total number of questions in the benchmark as its denominator (the \emph{full-query denominator}). Since attribution is multi-label, one incorrect sample may be assigned multiple defect codes.


\begin{table}[t]
\centering
\footnotesize
\setlength{\tabcolsep}{3pt}
\begin{tabular}{@{}lp{0.31\columnwidth}lp{0.34\columnwidth}@{}}
\toprule
\textbf{Code} & \textbf{Full name} & \textbf{Layer} & \textbf{Condition} \\
\midrule
\textsc{EM}   & Encoding Missing & Enc. & POS store \textsc{Miss} \\
\textsc{EA}   & Encoding Ambiguous & Enc. & ambiguous value \\
\textsc{EW}   & Encoding Wrong & Enc. & wrong value \\
\textsc{DMP}  & Dirty Memory Pollution & Enc. & NEG dirty memory \\
\textsc{RF}   & Retrieval Failure & Ret. & POS \textsc{Miss} with $S_{\text{enc}}\!=\!\textsc{Exist}$ \\
\textsc{LATE} & Retrieval Late & Ret. & hit rank $>\tau_{\text{rank}}$ \\
\textsc{NOI}  & Retrieval Noise & Ret. & hit SNR $<\tau_{\text{snr}}$ \\
\textsc{NIR}  & Noise-Induced Retrieval & Ret. & NEG misleading retrieval \\
\textsc{GH}   & Generation Hallucination & Gen. & NEG fails to refuse \\
\textsc{GF}   & Generation Faithfulness & Gen. & POS ignores oracle \\
\textsc{GRF}  & Generation Reasoning Failure & Gen. & POS reasoning error \\
\bottomrule
\end{tabular}
\caption{The $11$ defect codes used by \framework, grouped by layer. The full attribution for query $i$ is the multi-label union $\defectset_{\text{total}}^{(i)}$.}
\label{tab:defect_codes}
\vspace{-4mm}
\end{table}

\paragraph{Recall-first store diagnosis with adapted agentic RAG.}
The Encoding Examiner can only be trusted if it observes the memory store reliably.
Weak evaluator-side search may mislabel a stored but hard-to-find fact as \textsc{Miss}, turning a retrieval problem into a false encoding defect.
This risk is acute because the same evidence may appear as an atomic fact, dialogue turn, compressed summary, or document page.

To reduce such observation misses, \framework adapts agentic RAG into a read-only, recall-first diagnostic routine inside the Encoding Examiner.
\label{sec:framework:agenticrag}
Unlike answer-oriented agentic RAG, this routine is used only for evidence-existence checking: it may use both $Q$ and $F_{\text{key}}$ to inspect $\mathcal{M}$, but it never modifies the store or supplies context to the tested generator.

The routine extracts anchors from $(Q,F_{\text{key}})$, plans complementary views over the same memory export (original-query, temporal, entity, topical paraphrase, sparse keyword, and gap views), adapts retrieval budget to memory granularity, retrieves candidates with provenance metadata, and reranks them by direct support for $F_{\text{key}}$.
Fusion uses reciprocal rank fusion,
\begin{equation}
  s_{\text{rrf}}(d)=\sum_{v\in\mathcal{V}}\frac{1}{k_{\text{rrf}}+r_v(d)},
  \label{eq:rrf}
\end{equation}
because dense and sparse scores are not directly comparable.
The reranker prioritizes candidates expressing the entity, relation, value, and temporal constraints in $F_{\text{key}}$.

Since the target fact may genuinely be absent, the routine uses bounded convergence rather than stopping when an answer seems sufficient:
\begin{equation}
  \begin{aligned}
  &\underbrace{|\mathrm{pool}_t|\!\ge\!K_{\max}}_{\text{pool-full}}
  \;\vee\;
  \underbrace{\mathrm{Jacc}(R_t,R_{t-1})\!\ge\!\tau_{\text{J}}}_{\text{converged}}\\[-1pt]
  &\qquad\vee\;\underbrace{t\!\ge\!T_{\max}}_{\text{max-iter}} .
  \end{aligned}
\label{eq:agenticrag_stop}
\end{equation}
The final high-recall pool is judged only by the Encoding Examiner.

\subsection{\datasetname: On-Policy Dynamic Evaluation}
\label{sec:framework:dyn}

\datasetname evaluates memory on trajectories produced by the tested system rather than fixed dialogue replay.
Its construction keeps the semantic world and evaluation questions fixed for comparability while letting the actual dialogue unfold on policy for each system.
The full construction pipeline is illustrated in Appendix~\ref{app:dyn_construct}, Figure~\ref{fig:dyn_construct}.

\paragraph{Offline blueprint.}
We define personas and user questions that depend on evolving personal states, then derive a state schema $\Sigma$ and trajectory $\sigma_0,\ldots,\sigma_T$.
The trajectory specifies stable facts, updates, cascaded changes, and rollback cases, serving as the benchmark's hidden world state.
For each session $t$, we derive a must-expose set $E_t$: facts that must be naturally revealed because they first appear or change in $\sigma_t$.
From the same trajectory, we generate session themes, grounded seed utterances, POS/NEG probes, gold answers, $F_{\text{key}}$, oracle contexts, and trap answers for stale or unsupported cases.

\paragraph{On-policy interaction.}
Each system interacts with the user simulator independently.
Session $t$ starts with the same seed utterance grounded in $E_t$ for all systems; later user turns are role-played from the persona, current state $\sigma_t$, session theme, remaining unexposed items $E_t^{\text{rem}}$, dialogue history, and the system's previous response.
The assistant follows its native policy and updates memory normally.
An exposure tracker removes mentioned facts, and a force-expose routine verbalizes any remaining required items near the final turn.
Thus, systems produce different dialogues, but must-expose constraints guarantee the same observed fact set.

\paragraph{Read-only probes and validation.}
At the end of each session, scheduled evaluation questions are asked through a read-only channel, so evaluation does not write into the memory store or affect later sessions.
The probes include new POS questions about recently exposed facts, state-tracking POS questions whose answers change with the current state, and NEG questions about unmentioned, outdated, contradictory, or superficially similar but unsupported facts; all probes are judged by the same three Examiners.
We validate the benchmark by checking that all required facts are exposed, that each question is answerable from its oracle evidence, that NEG subtypes and cascaded state changes behave as designed, and by human and model meta-evaluation.
The resulting \datasetname has controlled state evolution, must-expose fairness constraints, read-only probes, and $962$ validated questions over $10$ personas.

\subsection{Diagnosis-Guided Actionability Check with \memwiki}
\label{sec:framework:memwiki}

\memwiki is not part of the core scoring protocol.
It checks whether retrieval defects identified by \framework are actionable: do the same stored facts become more useful when their retrieval surface is reorganized?
\memwiki builds a search-friendly auxiliary index from each system's memory export without oracle context, gold answers, $F_{\text{key}}$, evidence labels, or replacement of the native retriever.

\paragraph{Auxiliary records from raw memory.}
\memwiki organizes the exported memory $\mathcal{M}$ into a wiki-style structure.
Original records become source pages; repeated entities, topics, and events are aggregated into hub pages; and dynamic facts are versioned rather than overwritten.
The structure is flattened into auxiliary set $\mathcal{A}$ with four retrieval-oriented views: \textbf{factual}, \textbf{tag}, \textbf{query-shaped}, and \textbf{aggregate} records, each pointing back to source records.

\paragraph{Integration with native retrieval.}
At query time, the tested system retrieves over the union of its original memory records and auxiliary records.
Let $\mathrm{Retr}(Q,\cdot)$ denote the native retriever, and let $\mathrm{src}(a)$ return the original records pointed to by auxiliary record $a$.
The final candidate pool is
\begin{equation}
  \begin{aligned}
  C_{\text{final}} &= \mathrm{Rerank}\!\Bigl(
  \mathrm{Retr}(Q,\mathcal{M}\cup\mathcal{A})\\[-1pt]
  &\qquad\cup\bigcup_{a\in H_{\mathcal{A}}}\mathrm{src}(a)
  \Bigr),
  \end{aligned}
  \label{eq:memwiki_merge}
\end{equation}
where $H_{\mathcal{A}}$ is the subset of retrieved candidates that are auxiliary records.
This implements storage-surface injection, source expansion, and reranking while keeping the original evidence and native retriever fixed during this retrieval-side intervention.

\section{Experiments}
\label{sec:experiments}

We evaluate \framework on three datasets (\textsc{LoCoMo}~\citep{maharana2024evaluating}, \textsc{LongMemEval}-S~\citep{wu2024longmemeval}, and \datasetname) under three LLM backbone settings (\textsc{gpt-4o-mini}, \textsc{gpt-4.1}, and \textsc{Qwen3-30B-Instruct}~\citep{yang2025qwen3}). \S\ref{sec:exp:setup} describes the experimental setup. \S\ref{sec:exp:loco} reports end-to-end accuracy together with operation-level defect rates, including cross-dataset and backbone comparisons. \S\ref{sec:exp:mechanism} further examines whether the diagnostic components are reliable and actionable through three mechanism studies: agentic-RAG reliability, the \memwiki intervention, and top-$K$ sensitivity.

\subsection{Setup}
\label{sec:exp:setup}

\paragraph{Datasets.}
\textsc{LoCoMo}~\citep{maharana2024evaluating} contains $1986$ QA pairs over $10$ multi-session dialogues ($1540$ POS and $446$ NEG). \textsc{LongMemEval}-S~\citep{wu2024longmemeval} contains $500$ items ($470$ POS and $30$ NEG) with approximately $115$K-token haystacks across about $50$ sessions. \datasetname (\S\ref{sec:framework:dyn}) contains $962$ validated questions over $10$ personas ($790$ POS and $172$ NEG).

\paragraph{Memory systems.}
We evaluate seven released memory systems through dedicated adapters: TiMem~\citep{li2026timem}, O-Mem~\citep{wang2025mem}, EverMemOS~\citep{hu2026evermemos}, MemOS~\citep{li2025memos}, MemBox~\citep{tao2026membox}, MemoryOS~\citep{kang2025memory}, and GAM~\citep{yan2025general}. These systems cover hierarchical-temporal memory, system-level multi-tier memory management, engram-lifecycle modeling, memory-block abstraction, agentic deep-research memory, and user-profile-centric memory.

\paragraph{Models and hyper-parameters.}
Unless otherwise specified, both the memory-system-side LLM calls and answer generation use \textsc{gpt-4o-mini}. \textsc{gpt-4.1} and \textsc{Qwen3-30B-Instruct} are used for the cross-backbone study. The EvalMem-side agents, including the three Examiners, use \textsc{gpt-5-mini}. The user simulator for \datasetname uses \textsc{gpt-4o}. To keep retrieval comparisons controlled, we standardize the embedding encoder used by each system's embedding-based retrieval/index and by the corresponding diagnostic retrieval to Qwen3-Embedding-0.6B~\citep{zhang2025qwen3} ($1024$-d). This replaces only the encoder (and rebuilds its vectors/index entries as needed); memory construction, index type/structure, retrieval logic, answer backbone, and EvalMem judges remain unchanged. Appendix~\ref{app:native_embedding} reports an embedding-only rerun with each system's released/native encoder: systems whose encoder changes shift accuracy by $1.2$--$6.8$\,pp, while Retrieval remains the most frequently attributed layer; TiMem uses the standardized encoder natively. Temperatures are all fixed at $\tau\!=\!0.0$ for answer generation and judging. Unless otherwise stated, retrieval uses Top-$K\!=\!10$; we analyze this choice in \S\ref{sec:exp:mechanism} and Appendix~\ref{app:topk}. We set $\tau_{\text{rank}}\!=\!5$, $\tau_{\text{snr}}\!=\!0.20$, and $\tau_H\!=\!0.55$. These values were selected with system-stratified human calibration on fixed pilot samples; Appendix~\ref{app:threshold_calibration} reports one-at-a-time sensitivity sweeps and calibration macro-F1, with each selected value fixed before aggregating test results. The agentic-RAG diagnostic procedure uses $T_{\max}\!=\!4$, pool cap $50$, and Jaccard convergence threshold $0.85$.

\subsection{End-to-End Results across Datasets and Backbones}
\label{sec:exp:loco}

Table~\ref{tab:main_results} reports per-system end-to-end accuracy, the three layer-aggregate defect rates, and the full $11$-code distribution for three main configurations: \textsc{LoCoMo} with the default backbone, \textsc{LoCoMo} with \textsc{gpt-4.1}, and \datasetname with the default backbone. 
All code and layer rates follow the definition in
\S\ref{sec:framework:examiners}: unless otherwise noted, only incorrectly
answered samples contribute defect codes, the denominator is the total number of
questions in the corresponding benchmark, and one sample may be assigned multiple defect codes.
Appendix~\ref{app:full_matrix} provides the complete $7\!\times\!3\!\times\!3$ matrix over seven systems, three datasets, and three backbones. We use \textsc{LongMemEval}-S as an auxiliary generalization set. Because this dataset is relatively small and contains only $30$ NEG samples, we report only accuracy and layer-level aggregate defect rates for it.


\providecommand{\BR}[1]{\textbf{\textcolor{red}{#1}}}
\providecommand{\UB}[1]{\textcolor{blue}{\underline{#1}}}
\definecolor{hmsub}{HTML}{ECECEC}

\begin{table*}[t]
\centering
\scriptsize
\setlength{\tabcolsep}{2.0pt}
\renewcommand{\arraystretch}{0.92}
\resizebox{0.82\textwidth}{!}{%
\begin{tabular}{l|ccccc|ccccc|cccc|c}
\toprule
\multirow{2}{*}{\textbf{System}}
 & \multicolumn{5}{c|}{\textbf{Encoding} $\downarrow$}
 & \multicolumn{5}{c|}{\textbf{Retrieval} $\downarrow$}
 & \multicolumn{4}{c|}{\textbf{Generation} $\downarrow$}
 & \multirow{2}{*}{\textbf{Acc} $\uparrow$} \\
\cmidrule(lr){2-6}\cmidrule(lr){7-11}\cmidrule(lr){12-15}
 & \textsc{EM} & \textsc{EA} & \textsc{EW} & \textsc{DMP} & \textbf{Enc}
 & \textsc{RF} & \textsc{LATE} & \textsc{NOI} & \textsc{NIR} & \textbf{Ret}
 & \textsc{GF} & \textsc{GRF} & \textsc{GH} & \textbf{Gen}
 & \\
\midrule
\rowcolor{hmsub}\multicolumn{16}{c}{\textit{Set 1:\ \textsc{LoCoMo}\,+\,\textsc{gpt-4o-mini}}\quad $N\!=\!1986$\quad mean Acc $=68.1$,\ mean Ret $=22.1$}\\
\midrule
TiMem      & 3.3 & 1.8 & 1.2 & 0.8 & 7.1 & \BR{9.1}  & \BR{2.7} & \BR{3.5} & \UB{1.7} & \BR{17.0} & \BR{2.1} & 2.6 & \UB{1.3} & \BR{6.0} & \BR{75.3} \\
O-Mem      & \BR{2.1} & \BR{1.3} & \BR{1.0} & \BR{0.5} & \BR{4.9} & 10.5 & \UB{3.3} & \UB{3.7} & 2.1 & \UB{19.6} & 2.4 & \UB{2.5} & \BR{1.2} & \UB{6.1} & \UB{74.9} \\
EverMemOS  & 4.6 & 2.0 & 1.3 & \UB{0.7} & 8.6 & \UB{10.4} & 3.4 & 4.0 & 2.2 & 20.0 & 2.5 & \BR{2.4} & 1.4 & 6.3 & 72.4 \\
MemOS      & 4.2 & 1.6 & 1.4 & 0.9 & 8.1 & 11.0 & 3.6 & 4.4 & 2.0 & 21.0 & 2.6 & 2.7 & \BR{1.2} & 6.5 & 70.1 \\
MemBox     & \UB{2.4} & \UB{1.4} & \UB{1.1} & 1.0 & \UB{5.9} & 14.7 & 4.5 & 4.3 & 2.2 & 25.7 & 2.7 & 2.6 & 1.6 & 6.9 & 63.2 \\
MemoryOS   & 4.7 & 2.3 & 1.5 & 0.8 & 9.3 & 13.7 & 4.9 & 5.0 & 2.4 & 26.0 & \UB{2.3} & 2.8 & 1.5 & 6.6 & 60.8 \\
GAM        & 5.2 & 2.5 & 1.3 & 1.1 & 10.1 & 14.3 & 4.4 & 5.6 & \BR{1.4} & 25.7 & 2.8 & 2.7 & 1.6 & 7.1 & 60.2 \\
\midrule
\textit{mean} & 3.8 & 1.8 & 1.3 & 0.8 & 7.7 & 12.0 & 3.8 & 4.4 & 2.0 & 22.1 & 2.5 & 2.6 & 1.4 & 6.5 & 68.1 \\
\midrule
\rowcolor{hmsub}\multicolumn{16}{c}{\textit{Set 2:\ \textsc{LoCoMo}\,+\,\textsc{gpt-4.1}}\quad $N\!=\!1986$\quad mean Acc $=70.9$,\ mean Ret $=20.2$}\\
\midrule
O-Mem      & \BR{1.5} & \BR{1.0} & \BR{0.7} & \BR{0.5} & \BR{3.7} & 9.5  & \UB{2.9} & \UB{3.3} & 1.8 & \UB{17.5} & \BR{1.4} & 1.9 & \UB{0.9} & \UB{4.2} & \BR{78.1} \\
TiMem      & 2.5 & 1.4 & 1.0 & \UB{0.6} & 5.5 & \BR{8.1}  & \BR{2.6} & \BR{3.1} & \UB{1.6} & \BR{15.4} & \UB{1.5} & \UB{1.8} & \BR{0.8} & \BR{4.1} & \UB{77.8} \\
EverMemOS  & 3.4 & 1.5 & 1.0 & 0.7 & 6.6 & \UB{9.4}  & 3.0 & 3.6 & 1.8 & 17.8 & 1.6 & \BR{1.7} & 1.1 & 4.4 & 75.2 \\
MemOS      & 3.0 & 1.5 & 1.1 & \BR{0.5} & 6.1 & 10.0 & 3.2 & 3.8 & 1.7 & 18.7 & 1.8 & 2.0 & \UB{0.9} & 4.7 & 73.1 \\
MemBox     & \UB{2.4} & \UB{1.1} & \UB{0.9} & 1.0 & \UB{5.4} & 13.6 & 4.0 & 4.3 & 2.1 & 24.0 & 1.9 & 2.1 & 1.2 & 5.2 & 65.7 \\
MemoryOS   & 3.7 & 1.8 & 1.2 & 0.8 & 7.5 & 13.1 & 4.3 & 4.6 & 2.4 & 24.4 & 1.7 & 2.0 & 1.1 & 4.8 & 63.8 \\
GAM        & 4.5 & 2.0 & 1.2 & 0.9 & 8.6 & 13.0 & 4.2 & 5.1 & \BR{1.5} & 23.8 & 2.0 & 2.2 & 1.0 & 5.2 & 62.8 \\
\midrule
\textit{mean} & 3.0 & 1.5 & 1.0 & 0.7 & 6.2 & 11.0 & 3.5 & 4.0 & 1.8 & 20.2 & 1.7 & 2.0 & 1.0 & 4.6 & 70.9 \\
\midrule
\rowcolor{hmsub}\multicolumn{16}{c}{\textit{Set 3:\ \datasetname\,+\,\textsc{gpt-4o-mini}}\quad $N\!=\!962$\quad mean Acc $=60.8$,\ mean Ret $=25.9$}\\
\midrule
TiMem      & \UB{3.5} & \UB{2.0} & \BR{1.4} & \BR{0.7} & \UB{7.6} & \BR{9.3}  & \BR{3.3} & \BR{3.8} & \BR{2.6} & \BR{19.0} & \BR{1.5} & \UB{1.9} & \BR{0.9} & \BR{4.3} & \BR{71.2} \\
O-Mem      & \BR{3.1} & \BR{1.8} & 1.7 & \UB{0.9} & \BR{7.5} & \UB{10.9} & \UB{3.6} & \UB{4.2} & \UB{2.8} & \UB{21.5} & 1.7 & \BR{1.8} & \UB{1.0} & \UB{4.5} & \UB{68.9} \\
MemOS      & 4.3 & 2.1 & \UB{1.5} & 1.1 & 9.0 & 13.3 & 4.1 & 4.5 & 3.5 & 25.4 & 1.8 & 2.0 & 1.1 & 4.9 & 62.5 \\
EverMemOS  & 4.8 & 2.2 & 1.6 & 1.0 & 9.6 & 13.7 & 4.0 & 4.7 & 3.7 & 26.1 & \UB{1.6} & 2.1 & \UB{1.0} & 4.7 & 61.0 \\
MemoryOS   & 5.4 & 2.6 & 1.7 & 1.3 & 11.0 & 14.7 & 4.4 & 5.2 & 4.1 & 28.4 & 1.9 & 2.0 & 1.2 & 5.1 & 56.4 \\
MemBox     & 4.5 & 2.3 & 1.8 & 1.4 & 10.0 & 15.4 & 4.6 & 5.0 & 4.1 & 29.1 & 2.0 & 2.2 & 1.3 & 5.5 & 55.9 \\
GAM        & 6.5 & 2.9 & 2.1 & 1.5 & 13.0 & 16.9 & 5.1 & 5.6 & 4.4 & 32.0 & 2.1 & 2.4 & 1.4 & 5.9 & 49.6 \\
\midrule
\textit{mean} & 4.6 & 2.3 & 1.7 & 1.1 & 9.7 & 13.5 & 4.2 & 4.7 & 3.6 & 25.9 & 1.8 & 2.1 & 1.1 & 5.0 & 60.8 \\
\bottomrule
\end{tabular}%
}
\caption{Accuracy and per-layer defect rates for seven memory systems on three configurations: \textsc{LoCoMo}+\textsc{gpt-4o-mini} (Set 1), \textsc{LoCoMo}+\textsc{gpt-4.1} (Set 2, cross-backbone), and \datasetname+\textsc{gpt-4o-mini} (Set 3). Columns: encoding codes + \textbf{Enc} total, retrieval codes + \textbf{Ret} total, generation codes + \textbf{Gen} total, end-to-end \textbf{Acc}; defect codes are assigned only to incorrectly answered outputs, while each defect column reports a sample-level union rate with the full-query denominator. Defect codes may overlap. \BR{Bold red} = best per column within each Set; \UB{blue underline} = second-best (ties marked at every tied cell). \textsc{Qwen3}, \textsc{LongMemEval}-S, and the full $7\!\times\!3\!\times\!3$ matrix are in Appendix~\ref{app:full_matrix}.}
\label{tab:main_results}
\vspace{-3mm}
\end{table*}

\paragraph{Finding 1: retrieval is the most frequently attributed failure layer.}
Under the default backbone, the mean retrieval-defect incidence is consistently higher than both encoding and generation across all three datasets: $22.1$ vs.\ $7.7$/$6.5$ on \textsc{LoCoMo}, $18.3$ vs.\ $8.5$/$5.4$ on \textsc{LongMemEval}-S, and $25.9$ vs.\ $9.7$/$5.0$ on \datasetname. On the two datasets with reliable fine-grained code breakdowns, the largest retrieval sub-code is \textsc{RF}, consistent with the diagnostic observation that the relevant fact is present in the memory store but is not surfaced by the system's native retriever. 

\paragraph{Finding 2: system rankings depend on dataset characteristics.}
No single memory system dominates across all evaluation settings. Under the default backbone, TiMem achieves the highest accuracy on \textsc{LoCoMo} and \datasetname, whereas EverMemOS leads on \textsc{LongMemEval}-S. Under \textsc{gpt-4.1} on \textsc{LoCoMo}, O-Mem slightly exceeds TiMem by $0.3$\,pp. This pattern is consistent with the different emphases of the datasets: \textsc{LongMemEval}-S stresses factual recall from very long haystacks, whereas \datasetname emphasizes state evolution and stale-evidence handling. Therefore, a single aggregate leaderboard can obscure different capability profiles, and memory-system evaluation should report layer-level diagnostics together with cross-dataset results.

\paragraph{Finding 3: dynamic state changes expose stale-evidence failures.}
Compared with \textsc{LoCoMo} under the same default backbone, all seven systems obtain lower accuracy on \datasetname, with a mean drop of $-7.3$\,pp. The drop is mainly accompanied by higher retrieval and encoding defect rates ($+3.8$ and $+2.0$\,pp), while generation defects decrease by $1.5$\,pp. The retrieval sub-code with the largest increase is \textsc{NIR}, which rises from $2.0\%$ on \textsc{LoCoMo} to $3.6\%$ on \datasetname. In the dynamic benchmark audit, the rollback subset is also the hardest aggregate subset, with only $47.3\%$ accuracy (Appendix~\ref{app:dyn_audit}). These results suggest that current systems can often retain old evidence after state changes, but they do not reliably prevent stale or misleading records from affecting later answers.

\paragraph{Backbone effect.}
Replacing \textsc{gpt-4o-mini} with \textsc{gpt-4.1} increases mean accuracy by $+2.8$, $+4.7$, and $+3.7$\,pp on \textsc{LoCoMo}, \textsc{LongMemEval}-S, and \datasetname, respectively. Using \textsc{Qwen3-30B-Instruct} keeps mean accuracy within $-1.3$\,pp of the default backbone. System rankings remain highly stable across backbone settings (Spearman $\rho\!\ge\!0.929$), with rank flips mainly occurring between near-tied systems. Thus, the observed cross-system differences are not artifacts of a single backbone.

\begin{figure*}[!t]
\centering
\includegraphics[width=0.9\textwidth]{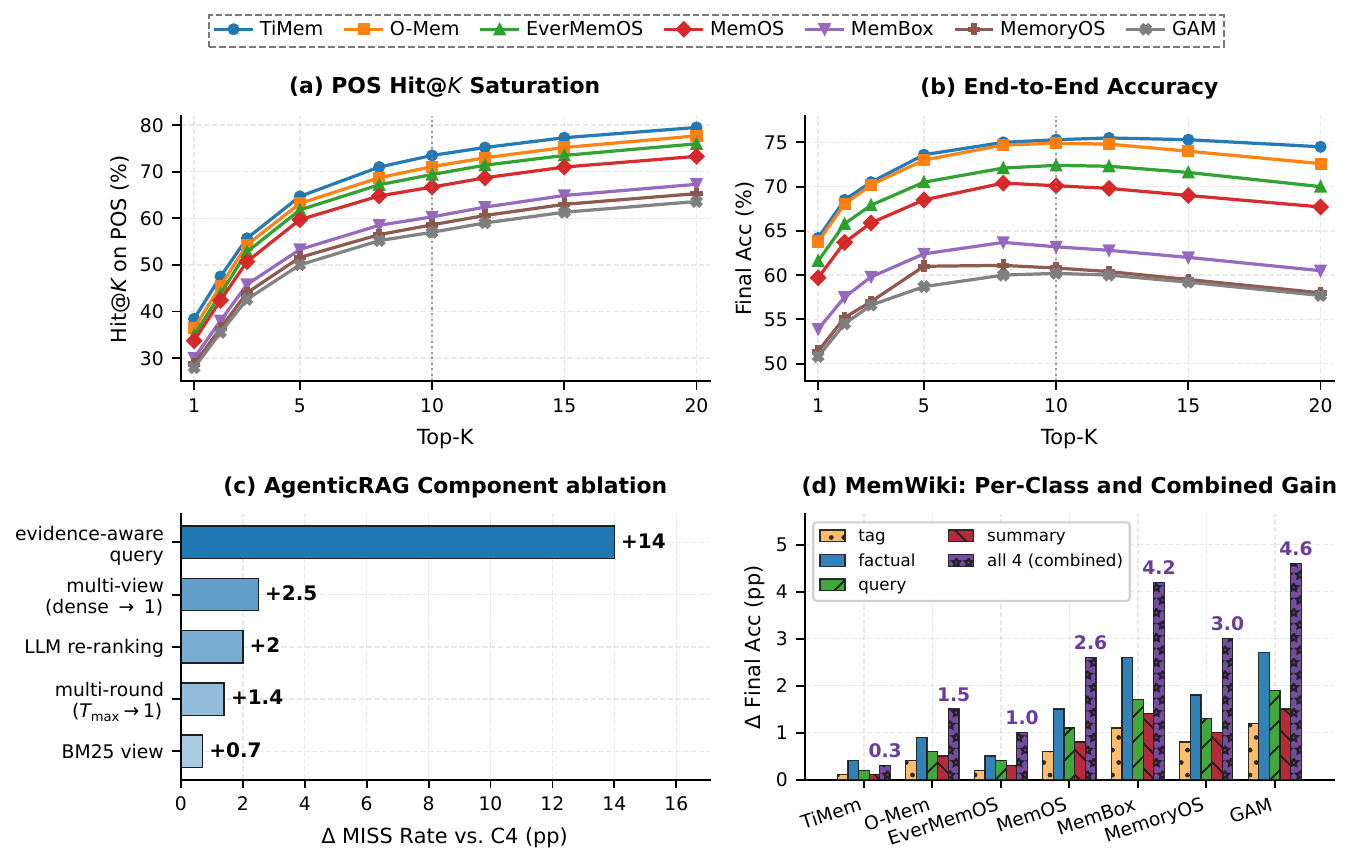}
\caption{Mechanism analyses on \textsc{LoCoMo} with \textsc{gpt-4o-mini}. \textbf{(a)} POS Hit@$K$ saturates around $K\!=\!10$. \textbf{(b)} Final accuracy peaks at $K\!\in\!\{8,10\}$. \textbf{(c)} Agentic-RAG reliability analysis: the evidence-aware query is the largest contributor to reducing diagnostic MISS, while the agentic-RAG pipeline mechanisms provide additional gains. \textbf{(d)} \memwiki per-class and combined intervention: factual records are the strongest single auxiliary class.}
\label{fig:mechanism}
\vspace{-4mm}
\end{figure*}

\subsection{Mechanism Studies: Diagnostic Reliability, Intervention, and Top-$K$}
\label{sec:exp:mechanism}

Figure~\ref{fig:mechanism} summarizes the three mechanism studies. Full per-cell numbers and ablations are reported in Appendix~\ref{app:mechanism_detail}.

\paragraph{Agentic-RAG reliability (panel c).}
The first mechanism study evaluates whether the agentic-RAG procedure used inside the Encoding Examiner can make store-level observation more reliable. This study concerns the evaluator-side diagnosis of whether relevant evidence exists in the memory store, rather than the tested system's native retrieval quality. On the \textsc{LoCoMo} POS subset, a vanilla diagnostic retriever using only $Q$ misses $29.8\%$ of evidence known to be present. Adding only the agentic-RAG pipeline lowers the MISS rate to $18.4\%$; adding only the evidence-aware query lowers it to $11.3\%$; using both further lowers it to $4.4\%$ (Appendix~\ref{app:memscan_detail}). A leave-one-out analysis from the full configuration shows that the evidence-aware query is the largest contributor: removing it raises MISS by $14.0$\,pp. The remaining agentic-RAG pipeline mechanisms contribute $0.7$--$2.5$\,pp each. These results support using the full diagnostic configuration to reduce false encoding attributions caused by incomplete evaluator-side search.

\paragraph{\memwiki intervention (panel d).}
\memwiki provides an actionability check for a diagnosis-guided retrieval-side intervention. On \textsc{LoCoMo} with \textsc{gpt-4o-mini}, enabling all four auxiliary structures improves mean final accuracy by $+2.5$\,pp and reduces the mean retrieval-defect rate by $3.2$\,pp. Gains are largest for systems with higher observed baseline retrieval-defect rates (GAM $+4.6$\,pp; MemBox $+4.2$\,pp) and smallest for TiMem ($+0.3$\,pp), whose native profile layer already supplies part of this surface. 
Repeating the intervention on \textsc{LongMemEval}-S yields $+2.3$\,pp accuracy and a $3.2$\,pp retrieval-defect reduction (Table~\ref{tab:memwiki_longmemeval}). McNemar tests and quota ablations are in Appendix~\ref{app:memwiki}.

\begin{table}[t]
\centering
\small
\setlength{\tabcolsep}{4pt}
\renewcommand{\arraystretch}{0.98}
\begin{tabular}{l|cccc}
\toprule
\textbf{System} & \textbf{Bsl Acc} & \textbf{+MW Acc} & $\Delta$ \textbf{Acc} & $\Delta$ \textbf{Ret} \\
\midrule
EverMemOS & 82.6 & 83.4 & $+0.8$ & $-1.4$ \\
O-Mem     & 81.4 & 82.8 & $+1.4$ & $-2.6$ \\
MemBox    & 77.0 & 79.6 & $+2.6$ & $-3.4$ \\
GAM       & 76.6 & 79.4 & $+2.8$ & $-3.6$ \\
TiMem     & 76.0 & 76.6 & $+0.6$ & $-1.4$ \\
MemOS     & 65.0 & 68.4 & $+3.4$ & $-4.4$ \\
MemoryOS  & 55.4 & 59.8 & $+4.4$ & $-5.4$ \\
\midrule
\textit{mean} & \textit{73.4} & \textit{75.7} & \textbf{+2.3} & \textbf{$-3.2$} \\
\bottomrule
\end{tabular}
\caption{\memwiki{} on \textsc{LongMemEval}-S (gpt-4o-mini). Mean Acc / Ret shifts mirror the \textsc{LoCoMo} result within $0.2$\,pp; per-system significance and Pearson $r(\Delta\text{Ret},\Delta\text{Acc})$ in Appendix~\ref{app:memwiki}.}
\label{tab:memwiki_longmemeval}
\vspace{-1.5mm}
\end{table}

\paragraph{Top-$K$ sensitivity (panels a, b).}
Increasing $K$ monotonically improves POS Hit@$K$, but it does not monotonically improve end-to-end accuracy. Hit@$K$ shows an elbow around $K\!=\!10$, with marginal gains below $1.0$\,pp per additional candidate beyond that point. Final accuracy peaks at $K\!\in\!\{8,10\}$ for the mean curve and for nearly all systems, then declines as larger contexts introduce more noise. Appendix~\ref{app:topk} shows the mechanism: from $K\!=\!10$ to $K\!=\!20$, RF decreases but NOI rises from $5.5\%$ to $12.5\%$, indicating that additional candidates increasingly become distractors. We therefore use $K\!=\!10$ as the retrieval budget.

\paragraph{Diagnostic stability and overhead.}
Holding the baseline systems' \textsc{LoCoMo} outputs fixed and rerunning the full EvalMem diagnostic pipeline five times yields layer-level standard deviations of at most $0.25$\,pp and Fleiss' $\kappa\!\ge\!0.89$; this measures diagnostic stochasticity rather than memory-system seed variance (Appendix~\ref{app:diagnostic_stability}). The mean per-example EvalMem diagnostic overhead is $8.48$ calls, $47.2$K input tokens, and $2.04$K output tokens, with critical-path latency of $13.0$/$28.9$ seconds at p50/p95. Because the three Examiners run in parallel before Attribution, the critical path is approximately $\max(t_E,t_R,t_G)+t_A$, not the sum of component latencies (Appendix~\ref{app:diagnostic_overhead}).

\section{Conclusion}
\label{sec:conclusion}


\framework turns long-term memory evaluation from end-to-end scoring into operation-level attribution. By combining Encoding, Retrieval, and Generation Examiners with task-aware multi-label rules, it diagnoses whether a failure arises because the relevant fact was not stored, not retrieved, or not used. A recall-first agentic-RAG diagnostic routine makes store-level observation more reliable, while \datasetname extends evaluation to on-policy state evolution. Across seven memory systems and three datasets, retrieval is the most frequently attributed failure layer in the evaluated settings: the reported Retrieval attribution frequency exceeds the Encoding and Generation frequencies in every setting, and dynamic interaction exposes stale-evidence failures that static replay can miss. Finally, \memwiki improves accuracy on \textsc{LoCoMo} and \textsc{LongMemEval}-S by reorganizing existing memories into a more searchable surface, providing an actionability check for retrieval-side reorganization.

\section*{Limitations}
\label{sec:limitations}

The three Examiners in \framework are realised as LLM-judge components, and the corresponding attributions are therefore subject to the standard caveats that apply to LLM-as-judge protocols; we report Cohen's $\kappa$ against trained annotators in Appendix~\ref{app:reliability_dyn_validation} so that this dependency is quantified. Our experiments cover three datasets and three answer-generation backbones drawn from a single language; broader coverage --- additional languages and domains --- would further consolidate the generality of the framework. We evaluate seven public memory systems under a controlled standardized configuration and an embedding-only native-encoder robustness check; how attribution behaves under system-specific tuning remains open for follow-up work.

\section*{Ethics Statement}
\label{sec:ethics}

All datasets used in this work are either publicly available (\textsc{LoCoMo}, \textsc{LongMemEval}-S) or generated through a controlled persona-driven simulator (\datasetname), and no human-subject data was collected during this study. The personas in \datasetname are synthetic and do not represent any real individual; they are derived from a small grid over commonly studied user-profile dimensions (age, education, communication style, change rate, NEG sensitivity) and are released alongside the benchmark to support reproducible evaluation. The memory systems we evaluate are released by their respective authors with open licences; we use them strictly within those licence terms and do not redistribute model weights. We anticipate the primary impact of \framework to be diagnostic: by making operation-level failure modes visible, the framework helps the community measure long-term memory progress more rigorously and avoid over-claiming based on opaque end-to-end metrics. We see no immediate dual-use risk specific to this work beyond the general considerations that apply to long-term memory deployments (privacy of accumulated user content, traceability of stored records); these considerations are inherited from the memory systems we evaluate rather than introduced by the framework itself.

\FloatBarrier



\section*{Acknowledgments}

This research is supported by National Natural Science Foundation of China
under Grant 62406163, Tianjin Science and Technology Plan Project under
Grant 25ZXSFSN00150, and the Fundamental Research Funds for the Central
Universities under Grant 63261039. We thank the anonymous reviewers for
their constructive feedback, which helped us clarify the scope, attribution
rules, and limitations of this work.  We also thank the developers and
maintainers of the open-source memory systems and benchmark resources used
in our experiments for making these resources available to the community.

\bibliography{custom}
\FloatBarrier

\appendix
\raggedbottom

\section{Additional Related Work}
\label{app:related}

\paragraph{From long-context evaluation to memory-system evaluation.}
Early long-context benchmarks primarily test whether a model can locate and reason over evidence already present in a long input sequence, as in \textsc{LongBench}, \textsc{Ada-LEval}, \textsc{XL}$^2$\textsc{Bench}, and \textsc{BABILong}~\citep{bai2024longbench,wang2024ada,ni2024xl,kuratov2024babilong}.
Long-term memory benchmarks shift the focus from passive reading to persistent interaction: the system must decide what to store, how to update it, and when to retrieve it across sessions.
Representative datasets such as \textsc{LoCoMo} and \textsc{LongMemEval} establish this setting for conversational recall, temporal reasoning, knowledge updates, and abstention~\citep{maharana2024evaluating, wu2024longmemeval}.
This distinction is important for \framework: our target is not long-context saturation by itself, but diagnosing failures in externalized memory loops after information has been written, reorganized, and later queried.

\paragraph{Static, incremental, and on-policy memory benchmarks.}
Recent benchmarks form a progression from fixed-history evaluation to increasingly interactive settings.
\textsc{LoCoMo} and \textsc{LongMemEval} largely evaluate systems against fixed or pre-recorded histories~\citep{maharana2024evaluating,wu2024longmemeval}.
\textsc{MemoryAgentBench} and \textsc{MemoryBench} broaden the task surface toward incremental multi-turn interaction, continual learning, and user feedback~\citep{hu2025evaluating,ai2025memorybench}.
\textsc{HaluMem} focuses on hallucination under memory extraction, updating, and question answering~\citep{chen2025halumem}, while \textsc{AMemGym} moves evaluation on-policy so that the tested assistant's own interaction trajectory affects later memory use~\citep{jiayang2026amemgym}.
Very recent work continues this move toward self-evolving settings~\citep{wang2026evomembench}.
\datasetname is closest in spirit to the interactive line, but differs in objective: rather than maximizing environmental realism alone, it introduces a controlled must-expose schedule so different systems remain comparable at the level of fact exposure while still generating their own dialogue trajectories.

\paragraph{From stage-aware evaluation to failure attribution.}
Several prior works already recognize that end-to-end answer accuracy is too coarse for understanding memory systems.
\textsc{LongMemEval} discusses indexing, retrieval, and reading~\citep{wu2024longmemeval}; \textsc{HaluMem} decomposes hallucinations across extraction, update, and question answering~\citep{chen2025halumem}; \textsc{AMemGym} reports writing, reading, and utilization metrics~\citep{jiayang2026amemgym}; and \textsc{MemoryAgentBench} evaluates broader memory capabilities through multi-turn interaction~\citep{hu2025evaluating}.
These studies motivate operation-aware evaluation, but they do not fully solve attribution for deployed memory systems, where a single failure can span multiple operations at once.
A fact may be encoded ambiguously, retrieved too late, and then ignored by the generator; collapsing such a case into one bucket hides the coupled nature of the error.
\framework therefore frames diagnosis as a parallel, multi-label attribution problem rather than a single-stage classification problem.

\paragraph{Memory architectures and retrieval-side optimization.}
Recent memory systems differ substantially in how they store, organize, and retrieve past interactions.
Hierarchical and temporal systems maintain multi-level summaries or time-aware indices~\citep{kang2025memory,li2026timem}; operating-system-style designs separate memory types and lifecycle operations~\citep{li2025memos,hu2026evermemos}; and other systems use modular memory blocks, user-centric profiles, or agentic memory construction~\citep{tao2026membox, wang2025mem,yan2025general,xu2025mem}.
Production-oriented systems similarly combine explicit memory stores with retrieval-augmented generation~\citep{chhikara2025mem0, rasmussen2025zep}.
This diversity makes direct architectural comparison difficult, and leaderboard accuracy may obscure which component is responsible for an observed gain or failure.
It also leaves retrieval under-examined: many memory systems invest heavily in what to store, but still expose the LLM to a relatively flat top-$K$ retrieved context.
Retrieval-augmented generation and agentic retrieval methods provide useful building blocks for high-recall search and iterative query refinement~\citep{lewis2020retrieval,asai2024self,trivedi2023interleaving}, but they are usually proposed as task-solving mechanisms rather than as diagnostic instruments for memory evaluation.
In contrast, \framework uses an adapted agentic RAG procedure as a post-hoc probe for store-level diagnosis, and \memwiki as a diagnosis-guided auxiliary structure to test whether retrieval defects can be turned into actionable system improvements.

\section{Theory and Definitions}
\label{app:theory}

\subsection{Adapted \memscan Pseudocode}
\label{app:memscan_alg}

\noindent
{\small
\begin{tabular}{p{0.92\columnwidth}}
\toprule
\textbf{Algorithm 1: adapted \memscan diagnostic routine (sketch)} \\
\midrule
\textbf{Input}: query $Q$, key fact $F_{\text{key}}$, store $\mathcal{M}$. \\
\textbf{Output}: candidate pool, stop reason. \\[0.2em]
$\mathrm{pool}\leftarrow\emptyset$; $\mathrm{prev}\leftarrow\emptyset$; $s\leftarrow 0$. \\
\textbf{for} $t=1\ldots\,T_{\max}$ \textbf{do} \\
\quad views $\leftarrow$ \textsc{plan}$(Q, F_{\text{key}}, \mathrm{diag})$ \\
\quad raw $\leftarrow$ \textsc{parallel-retrieve}(views, $\mathcal{M}$, top-$K$) \\
\quad fused $\leftarrow$ \textsc{rrf-fuse}(raw); ranked $\leftarrow$ \textsc{rerank}(fused) \\
\quad pool $\leftarrow$ \textsc{merge}(pool, ranked) \\
\quad \textbf{if} $|\mathrm{pool}|\!\ge\!50$ \textbf{return} pool, \textsc{pool-full} \\
\quad \textbf{if} \textsc{converged}(ranked, prev) \textbf{then} $s\!\leftarrow\!s\!+\!1$; \\
\quad \quad \textbf{if} $s\!\ge\!2$ \textbf{return} pool, \textsc{converged} \\
\quad prev $\leftarrow$ ranked \\
\textbf{return} pool, \textsc{max-iter} \\
\bottomrule
\end{tabular}
\par}

\vspace{0.5em}
\noindent
The adapted \memscan routine is a bounded multi-view recall procedure with deterministic convergence. The diagnosis vector $\mathrm{diag}$ summarises whether the previous round surfaced any candidate overlapping with $F_{\text{key}}$ and, if not, what sub-aspects of $F_{\text{key}}$ remain unanchored. Views span dense, sparse (BM25), and rewritten paraphrases derived from $F_{\text{key}}$; \textsc{rrf-fuse} is reciprocal-rank fusion. The two stop conditions (\textsc{pool-full} when the candidate pool exceeds $50$ records; \textsc{converged} when two consecutive rounds produce no new top-ranked candidates) guarantee termination in at most $T_{\max}\!=\!4$ rounds, with empirical mean of $1.74$ ($\sigma\!=\!0.81$) across $1540$ \textsc{LoCoMo} POS queries. We set $T_{\max}\!=\!4$ because $87.9\%$ of queries terminate via convergence or pool-full before reaching this cap (the remaining $12.1\%$ trigger \textsc{max-iter}); raising the cap to $T_{\max}\!=\!6$ would help only this long tail at near-linear cost while leaving the global MISS rate essentially unchanged.

\subsection{\memwiki Pseudocode}
\label{app:memwiki_alg}

\noindent
{\small
\begin{tabular}{p{0.92\columnwidth}}
\toprule
\textbf{Algorithm 2: \memwiki build (offline)} \\
\midrule
\textbf{Input}: raw memory $\mathcal{M}$ exported by the tested system. \\
\textbf{Output}: auxiliary record set $\mathcal{A}$. \\[0.2em]
$\mathcal{P}\!\leftarrow\!\emptyset$;\,\,$\mathcal{A}\!\leftarrow\!\emptyset$. \\
\textbf{for} each record $r\!\in\!\mathcal{M}$ \textbf{do} \\
\quad page $\leftarrow$ \textsc{source-page}$(r)$ \\
\quad page.tags $\leftarrow$ \textsc{normalize}(ent, topic, time) \\
\quad page.facts $\leftarrow$ \textsc{extract-atomic-facts}$(r)$ \\
\quad page.hq $\leftarrow$ \textsc{gen-hypo-questions}$(r)$ \\
\quad $\mathcal{P}\!\leftarrow\!\mathcal{P}\cup\{\text{page}\}$ \\
$\mathcal{P}\!\leftarrow\!\mathcal{P}\cup$ \textsc{aggregate-hubs}$(\mathcal{P})$\,\,\textit{// entity / topic / event} \\
$\mathcal{P}\!\leftarrow$ \textsc{apply-versions}$(\mathcal{P})$\,\,\textit{// append / rewrite} \\
\textbf{for} each page $p\!\in\!\mathcal{P}$ \textbf{do} \\
\quad $\mathcal{A}\!\leftarrow\!\mathcal{A}\cup$ \textsc{flatten}$(p,\,\{\textsc{fact},\textsc{tag},\textsc{query},\textsc{summary}\})$ \\
\quad each $a\!\in\!\mathcal{A}$ carries $\mathrm{src}(a)\!\subseteq\!\mathcal{M}$ \\
\textbf{return} $\mathcal{A}$ \\
\bottomrule
\end{tabular}
\par}

\vspace{0.5em}
\noindent
{\small
\begin{tabular}{p{0.92\columnwidth}}
\toprule
\textbf{Algorithm 3: \memwiki at query time (online)} \\
\midrule
\textbf{Input}: query $Q$, native retriever $\mathrm{Retr}$, budget $K$, quota $\alpha$. \\
\textbf{Output}: final context $C_{\text{final}}$. \\[0.2em]
$C_0\!\leftarrow\mathrm{Retr}(Q,\,\mathcal{M}\cup\mathcal{A})$\,\,\textit{// joint pool, native retriever} \\
$H_{\!\mathcal{A}}\!\leftarrow\!C_0\cap\mathcal{A}$;\,\,$E\!\leftarrow\!\bigcup_{a\in H_{\!\mathcal{A}}}\mathrm{src}(a)$ \\
$C_1\!\leftarrow$ \textsc{rerank}$(C_0\cup E)$\,\,\textit{// source-expansion before rerank} \\
$C_{\text{final}}\!\leftarrow$ \textsc{quota-cap}$(C_1,\,K,\,|\cdot\cap\mathcal{A}|\!\le\!\alpha K)$ \\
\textbf{return} $C_{\text{final}}$ \\
\bottomrule
\end{tabular}
\par}

\vspace{0.5em}
\noindent
Algorithm~2 is build-side and runs once per (system, persona). It is governed by three invariants: (i) the only readable input is $\mathcal{M}$ exported through the system's adapter --- the build never sees $A_{\text{gold}}$, $F_{\text{key}}$, $C_{\text{oracle}}$, or evidence labels; (ii) every auxiliary record $a$ stores a non-empty back-pointer $\mathrm{src}(a)\!\subseteq\!\mathcal{M}$, so an auxiliary hit can always be exchanged for the original evidence at query time; (iii) when a later page rewrites an earlier one, \textsc{apply-versions} writes a new version instead of overwriting --- a fact superseded across sessions is recorded as a (old-version, new-version) pair, both retaining their $\mathrm{src}$ pointers. \textsc{flatten} produces four record classes per page: \textsc{fact} (subject-predicate-object-time tuples), \textsc{tag} (normalised entity / topic / time anchors), \textsc{query} (pre-generated plausible questions answered by the evidence), and \textsc{summary} (hub-page aggregates spanning multiple sessions). Algorithm~3 is the query-time fusion that realises Eq.~(\ref{eq:memwiki_merge}): the native retriever runs once over $\mathcal{M}\!\cup\!\mathcal{A}$, every auxiliary hit is expanded back to its source, and the merged pool is reranked. The quota cap $\alpha\!=\!0.30$ prevents auxiliary records from crowding out original evidence; Table~\ref{tab:memwiki_alpha} reports the sweep that confirms it is a structural constraint rather than a tuning knob.

\subsection{Attribution Rules}
\label{app:attr_rules}

Tables~\ref{tab:attr_pos} and~\ref{tab:attr_neg} give the complete task-aware mapping from Examiner states to final defect sets.
For POS generation failures, $\defectset_{\text{gen}}^{+}$ is either $\{\textsc{GF}\}$ or $\{\textsc{GRF}\}$; for low-quality POS retrieval hits, $\defectset_{\text{low}}$ is a non-empty subset of $\{\textsc{LATE},\textsc{NOI}\}$.
For POS, \textsc{Hit-Low} means the evidence is retrieved but ranked too late or diluted by noise; \textsc{Hit-Clean} means the hit passes both checks.
For NEG, \textsc{Miss}/\textsc{Miss}/\textsc{Pass} is the healthy path.
The Generation Examiner uses a comparison schema in implementation: it records both the online answer produced under the system's own retrieved context and the oracle answer produced under $C_{\text{oracle}}$, while the attribution tables use the oracle-side generation state.
Full defect-code names are listed in Table~\ref{tab:defect_codes}.
The POS/NEG matrices below specify the mapping from probe states to defect codes.

\begin{table*}[!t]
\centering
\footnotesize
\setlength{\tabcolsep}{4pt}
\renewcommand{\arraystretch}{1.13}
\begin{adjustbox}{max width=\textwidth}
\begin{tabular}{@{}cllllp{0.30\textwidth}@{}}
\toprule
\textbf{ID} & $S_{\text{enc}}$ & $S_{\text{ret}}$ & $S_{\text{gen}}$ & $\defectset_{\text{total}}$ & \textbf{Attribution meaning} \\
\midrule
\textbf{P1} & \textsc{Miss} & \textsc{Miss} & \textsc{Pass} &
$\{\textsc{EM}\}$ &
Fact absent from store; retrieval miss is shielded. \\
\textbf{P2} & \textsc{Miss} & \textsc{Miss} & \textsc{Fail} &
$\{\textsc{EM}\}\cup\defectset_{\text{gen}}^{+}$ &
Storage failure plus oracle-side generation failure. \\
\midrule
\textbf{P3} & \makecell[l]{\textsc{Corrupt}\\[-1pt]\textsc{Ambig}} & Any & \textsc{Pass} &
$\{\textsc{EA}\}$ &
Stored evidence is referentially ambiguous. \\
\textbf{P4} & \makecell[l]{\textsc{Corrupt}\\[-1pt]\textsc{Ambig}} & Any & \textsc{Fail} &
$\{\textsc{EA}\}\cup\defectset_{\text{gen}}^{+}$ &
Ambiguous evidence plus generation failure. \\
\textbf{P5} & \makecell[l]{\textsc{Corrupt}\\[-1pt]\textsc{Wrong}} & Any & \textsc{Pass} &
$\{\textsc{EW}\}$ &
Stored value is wrong. \\
\textbf{P6} & \makecell[l]{\textsc{Corrupt}\\[-1pt]\textsc{Wrong}} & Any & \textsc{Fail} &
$\{\textsc{EW}\}\cup\defectset_{\text{gen}}^{+}$ &
Wrong value plus generation failure. \\
\midrule
\textbf{P7} & \textsc{Exist} & \textsc{Miss} & \textsc{Pass} &
$\{\textsc{RF}\}$ &
Fact is stored but native retrieval misses it. \\
\textbf{P8} & \textsc{Exist} & \textsc{Miss} & \textsc{Fail} &
$\{\textsc{RF}\}\cup\defectset_{\text{gen}}^{+}$ &
Retrieval failure plus generation failure. \\
\textbf{P9} & \textsc{Exist} & \textsc{Hit-Low} & \textsc{Pass} &
$\defectset_{\text{low}}$ &
Evidence appears, but ranking or SNR is poor. \\
\textbf{P10} & \textsc{Exist} & \textsc{Hit-Low} & \textsc{Fail} &
$\defectset_{\text{low}}\cup\defectset_{\text{gen}}^{+}$ &
Low-quality retrieval plus generation failure. \\
\textbf{P11} & \textsc{Exist} & \textsc{Hit-Clean} & \textsc{Pass} &
$\emptyset$ &
Healthy POS path. \\
\textbf{P12} & \textsc{Exist} & \textsc{Hit-Clean} & \textsc{Fail} &
$\defectset_{\text{gen}}^{+}$ &
Memory path is healthy; failure is generation-side. \\
\bottomrule
\end{tabular}
\end{adjustbox}
\caption{Attribution matrix for POS questions. Retrieval defects are counted only after the source fact is confirmed to exist in memory; otherwise retrieval miss is a consequence of encoding failure.}
\label{tab:attr_pos}
\end{table*}

\begin{table*}[!t]
\centering
\footnotesize
\setlength{\tabcolsep}{4pt}
\renewcommand{\arraystretch}{1.13}
\begin{adjustbox}{max width=\textwidth}
\begin{tabular}{@{}cllllp{0.33\textwidth}@{}}
\toprule
\textbf{ID} & $S_{\text{enc}}$ & $S_{\text{ret}}$ & $S_{\text{gen}}$ & $\defectset_{\text{total}}$ & \textbf{Attribution meaning} \\
\midrule
\textbf{N1} & \textsc{Dirty} & \textsc{Noise} & \textsc{Pass} &
$\{\textsc{DMP}\}$ &
Pseudo-fact is already in memory; retrieval noise is attributed to storage pollution. \\
\textbf{N2} & \textsc{Dirty} & \textsc{Noise} & \textsc{Fail} &
$\{\textsc{DMP},\textsc{GH}\}$ &
Dirty memory plus unsupported generation. \\
\textbf{N3} & \textsc{Dirty} & \textsc{Miss} & \textsc{Pass} &
$\{\textsc{DMP}\}$ &
Store is polluted, although native retrieval does not surface it. \\
\textbf{N4} & \textsc{Dirty} & \textsc{Miss} & \textsc{Fail} &
$\{\textsc{DMP},\textsc{GH}\}$ &
Dirty memory plus unsupported generation without retrieved support. \\
\midrule
\textbf{N5} & \textsc{Miss} & \textsc{Noise} & \textsc{Pass} &
$\{\textsc{NIR}\}$ &
Clean store, but retrieval surfaces a misleading record. \\
\textbf{N6} & \textsc{Miss} & \textsc{Noise} & \textsc{Fail} &
$\{\textsc{NIR},\textsc{GH}\}$ &
Misleading retrieval plus unsupported generation. \\
\textbf{N7} & \textsc{Miss} & \textsc{Miss} & \textsc{Pass} &
$\emptyset$ &
Healthy NEG path: clean memory, clean retrieval, refusal. \\
\textbf{N8} & \textsc{Miss} & \textsc{Miss} & \textsc{Fail} &
$\{\textsc{GH}\}$ &
Model fabricates without memory or retrieval support. \\
\bottomrule
\end{tabular}
\end{adjustbox}
\caption{Attribution matrix for NEG questions. \textsc{DMP} is assigned when an unsupported pseudo-fact is already written into memory; \textsc{NIR} is assigned only when a clean store is followed by misleading retrieval.}
\label{tab:attr_neg}
\end{table*}

\subsection{Defect Codes}
\label{app:defect_codes}

The full taxonomy of $11$ defect codes was summarised in Table~\ref{tab:defect_codes}. We collect their full natural-language definitions here for reproducibility.

\begin{itemize}[leftmargin=1em,itemsep=0.2em,topsep=0em]
\item \textbf{EM (Encoding Missing).} The Encoding Examiner reports \textsc{Miss}: no $(k, v)$ in $\mathcal{M}$ matches $k_{\text{target}}$.
\item \textbf{EA (Encoding Ambiguous).} A matching key exists but $v$ is unresolved (e.g.\ pronoun, date placeholder).
\item \textbf{EW (Encoding Wrong).} A matching key exists but $v\!\neq\!v_{\text{target}}$.
\item \textbf{DMP (Dirty Memory Pollution).} A NEG-side hallucinated fact is in $\mathcal{M}$.
\item \textbf{RF (Retrieval Failure).} $S_{\text{enc}}\!=\!\textsc{Exist}$ but $F_{\text{key}}\not\subseteq C_{\text{original}}$.
\item \textbf{LATE (Retrieval Late).} $F_{\text{key}}\subseteq C_{\text{original}}$ but $\text{rank}(F_{\text{key}})>\tau_{\text{rank}}\!=\!5$.
\item \textbf{NOI (Retrieval Noise).} $F_{\text{key}}\subseteq C_{\text{original}}$ but $\text{snr}(C_{\text{original}})<\tau_{\text{snr}}\!=\!0.20$.
\item \textbf{NIR (Noise-Induced Retrieval).} A NEG query surfaces a relevant-looking but misleading record (e.g.\ stale evidence after rollback).
\item \textbf{GH (Generation Hallucination).} A NEG query is answered (model fails to refuse) under oracle context.
\item \textbf{GF (Generation Faithfulness).} POS query: oracle context contains the answer, but the model uses parametric knowledge instead.
\item \textbf{GRF (Generation Reasoning Failure).} POS query: model reads the oracle context but reasons incorrectly (multi-hop, temporal, etc.).
\end{itemize}

\section{Experimental Results}
\label{app:experiments}

\subsection{Full Cross-Backbone, Cross-Dataset Results}
\label{app:full_matrix}
\label{app:backbone}


\providecommand{\BR}[1]{\textbf{\textcolor{red}{#1}}}
\providecommand{\UB}[1]{\textcolor{blue}{\underline{#1}}}
\definecolor{hmsub}{HTML}{ECECEC}

\begin{table*}[t]
\centering
\scriptsize
\setlength{\tabcolsep}{3pt}
\renewcommand{\arraystretch}{0.98}
\begin{tabular}{l|cccc|cccc|cccc}
\toprule
\multirow{2}{*}{\textbf{System}}
 & \multicolumn{4}{c|}{\textsc{gpt-4o-mini} (default)}
 & \multicolumn{4}{c|}{\textsc{gpt-4.1}}
 & \multicolumn{4}{c}{\textsc{Qwen3-30B-Instruct}} \\
\cmidrule(lr){2-5}\cmidrule(lr){6-9}\cmidrule(lr){10-13}
 & \textbf{Acc} $\uparrow$ & Enc $\downarrow$ & Ret $\downarrow$ & Gen $\downarrow$
 & \textbf{Acc} $\uparrow$ & Enc $\downarrow$ & Ret $\downarrow$ & Gen $\downarrow$
 & \textbf{Acc} $\uparrow$ & Enc $\downarrow$ & Ret $\downarrow$ & Gen $\downarrow$ \\
\midrule
\rowcolor{hmsub}\multicolumn{13}{c}{\textit{Set 1:\ \textsc{LoCoMo}}\quad $N\!=\!1986$}\\
\midrule
TiMem      & \BR{75.3} & 7.1  & \BR{17.0} & \BR{6.0} & \UB{77.8} & 5.5 & \BR{15.4} & \BR{4.1} & \BR{74.2} & 7.8  & \BR{17.6} & \BR{6.5} \\
O-Mem      & \UB{74.9} & \BR{4.9} & \UB{19.6} & \UB{6.1} & \BR{78.1} & \BR{3.7} & \UB{17.5} & \UB{4.2} & \UB{73.8} & \BR{5.5} & \UB{20.2} & \UB{6.7} \\
EverMemOS  & 72.4 & 8.6  & 20.0 & 6.3 & 75.2 & 6.6 & 17.8 & 4.4 & 71.0 & 9.3  & 20.7 & 6.9 \\
MemOS      & 70.1 & 8.1  & 21.0 & 6.5 & 73.1 & 6.1 & 18.7 & 4.7 & 69.2 & 8.7  & 21.8 & 7.1 \\
MemBox     & 63.2 & \UB{5.9} & 25.7 & 6.9 & 65.7 & \UB{5.4} & 24.0 & 5.2 & 61.7 & \UB{6.5} & 26.6 & 7.5 \\
MemoryOS   & 60.8 & 9.3  & 26.0 & 6.6 & 63.8 & 7.5 & 24.4 & 4.8 & 62.3 & 9.9  & 25.3 & 7.2 \\
GAM        & 60.2 & 10.1 & 25.7 & 7.1 & 62.8 & 8.6 & 23.8 & 5.2 & 59.7 & 10.8 & 26.6 & 7.8 \\
\midrule
\textit{mean} & 68.1 & 7.7 & 22.1 & 6.5 & 70.9 & 6.2 & 20.2 & 4.6 & 67.4 & 8.4 & 22.7 & 7.1 \\
\midrule
\rowcolor{hmsub}\multicolumn{13}{c}{\textit{Set 2:\ \textsc{LongMemEval}-S}\quad $N\!=\!500$\,\,(layer totals only; NEG subsample $N_{\text{NEG}}\!=\!30$ too small for $11$-code analysis)}\\
\midrule
EverMemOS  & \BR{82.6} & \BR{5.6} & \BR{12.0} & \BR{4.4} & \BR{86.4} & \BR{4.4} & \BR{9.4}  & \BR{3.4} & \BR{81.0} & \BR{6.0} & \BR{13.0} & \BR{4.8} \\
O-Mem      & \UB{81.4} & \UB{6.0} & \UB{12.8} & \UB{4.8} & \UB{85.6} & \UB{4.6} & \UB{10.0} & \UB{3.6} & \UB{80.2} & \UB{6.4} & \UB{13.6} & \UB{5.0} \\
MemBox     & 77.0 & 7.4  & 15.8 & 5.2 & 81.4 & 6.0 & 13.0 & 4.0 & 75.2 & 8.0  & 16.8 & 5.8 \\
GAM        & 76.6 & 7.6  & 16.0 & 5.4 & 81.0 & 6.2 & 13.2 & 4.2 & 76.0 & 7.8  & 16.8 & 5.6 \\
TiMem      & 76.0 & 7.6  & 16.6 & 5.6 & 80.0 & 6.4 & 13.8 & 4.4 & 74.8 & 8.2  & 17.4 & 6.0 \\
MemOS      & 65.0 & 11.2 & 24.0 & 6.0 & 73.2 & 8.6 & 18.4 & 4.8 & 64.2 & 11.2 & 24.2 & 6.4 \\
MemoryOS   & 55.4 & 14.4 & 30.6 & 6.2 & 58.8 & 13.2 & 28.4 & 5.0 & 56.8 & 14.0 & 29.8 & 6.6 \\
\midrule
\textit{mean} & 73.4 & 8.5 & 18.3 & 5.4 & 78.1 & 7.1 & 15.2 & 4.2 & 72.6 & 8.8 & 18.8 & 5.7 \\
\midrule
\rowcolor{hmsub}\multicolumn{13}{c}{\textit{Set 3:\ \datasetname}\quad $N\!=\!962$}\\
\midrule
TiMem      & \BR{71.2} & \UB{7.6} & \BR{19.0} & \BR{4.3} & \UB{74.5} & \UB{6.0} & \BR{18.0} & \BR{3.2} & \BR{69.6} & \UB{8.1} & \BR{19.9} & \BR{4.9} \\
O-Mem      & \UB{68.9} & \BR{7.5} & \UB{21.5} & \UB{4.5} & \BR{75.1} & \BR{5.5} & \UB{18.5} & \BR{3.0} & \UB{67.7} & \UB{8.1} & \UB{22.3} & \UB{5.1} \\
MemOS      & 62.5 & 9.0  & 25.4 & 4.9 & 65.7 & 8.0  & 24.9 & 3.6 & 60.3 & 9.9  & 26.4 & 5.5 \\
EverMemOS  & 61.0 & 9.6  & 26.1 & 4.7 & 64.3 & 8.5  & 25.5 & 3.4 & 59.0 & 10.3 & 27.1 & 5.3 \\
MemoryOS   & 56.4 & 11.0 & 28.4 & 5.1 & 59.8 & 9.5  & 28.0 & 3.8 & 57.5 & 10.9 & 27.9 & 5.7 \\
MemBox     & 55.9 & 10.0 & 29.1 & 5.5 & 58.9 & 9.0  & 28.5 & 4.3 & 54.2 & 10.7 & 30.1 & 6.1 \\
GAM        & 49.6 & 13.0 & 32.0 & 5.9 & 53.0 & 11.5 & 31.0 & 4.7 & 48.1 & 13.7 & 33.1 & 6.5 \\
\midrule
\textit{mean} & 60.8 & 9.7 & 25.9 & 5.0 & 64.5 & 8.3 & 24.9 & 3.7 & 59.5 & 10.2 & 26.7 & 5.6 \\
\bottomrule
\end{tabular}
\caption{Full $7\!\times\!3\!\times\!3$ matrix of end-to-end accuracy (Acc) and the three layer-aggregate defect rates (Enc, Ret, Gen) across seven memory systems, three datasets (Sets 1--3), and three answer-generation backbones. Defect codes are assigned only to incorrectly answered outputs; all defect rates are sample-level union incidences with the full-query denominator; defect codes may overlap. \BR{Bold red} marks the best per column within each Set (highest Acc, lowest defect); \UB{blue underline} marks the second-best; ties receive the same marker at every tied position. For \textsc{LongMemEval}-S (Set 2) we report only Acc and the three layer totals: with $N_{\text{NEG}}\!=\!30$, the smallest 11-code cell would be unstable beyond statistical reliability (Wilson 95\% CI $\ge\!25$\,pp); the layer-total view remains diagnostic and is the primary use of this dataset as a generalisation check on \textsc{LoCoMo}. Systems within each Set are ordered by Acc under the default backbone (\textsc{gpt-4o-mini}). The companion $11$-code resolution for \textsc{LoCoMo} and \datasetname appears in Figs.~\ref{tab:full_codes_loco}--\ref{tab:full_codes_dyn}.}
\label{tab:full_matrix}
\end{table*}


\begin{figure*}[!t]
\centering
\includegraphics[width=\textwidth]{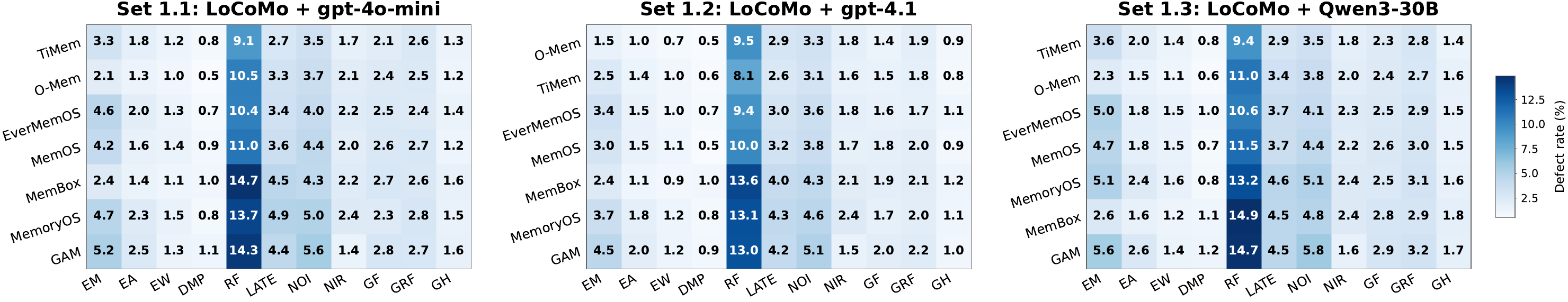}
\caption{Heatmap view of the full $11$-code sub-code breakdown on \textsc{LoCoMo} across three answer-generation backbones (Set 1.1--1.3). Rows are memory systems, columns are defect codes, and each cell shows the sample-level union defect rate (\%) over the full $N\!=\!1986$ POS+NEG query pool; codes are assigned only to incorrectly answered outputs. Color indicates defect magnitude (lighter is lower, darker is higher), while text annotations preserve exact values. Green boxes mark per-column best (lowest defect) within each Set; blue boxes mark second-best (ties are all marked). }
\label{tab:full_codes_loco}
\end{figure*}

\begin{figure*}[!ht]
\centering
\includegraphics[width=\textwidth]{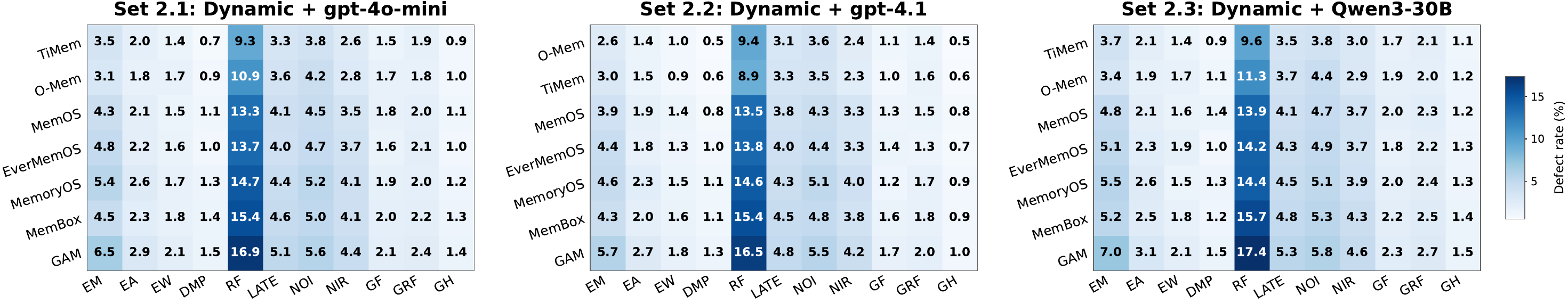}
\caption{Heatmap view of the full $11$-code sub-code breakdown on \datasetname across three answer-generation backbones (Set 2.1--2.3). Values are sample-level union defect rates (\%) over the full $N\!=\!962$ POS+NEG query pool; defect codes are assigned only to incorrectly answered outputs. The NIR column remains visibly darker than in \textsc{LoCoMo}, reflecting the larger share of retrieval defects caused by stale-evidence inference risk. Green and blue boxes denote per-column best and second-best systems within each Set.}
\label{tab:full_codes_dyn}
\end{figure*}

Table~\ref{tab:full_matrix} reports the complete $7\!\times\!3\!\times\!3$ grid: seven memory systems, three datasets (as vertical Sets), and three answer-generation backbones, each with Final Acc and the three layer-aggregate defect rates. The aggregate row at the bottom of each Set is the unweighted mean over the seven systems. Figures~\ref{tab:full_codes_loco}--\ref{tab:full_codes_dyn} (heatmaps) resolve every layer total into its $11$ defect codes on the two primary datasets; we omit \textsc{LongMemEval}-S from the sub-code resolution because its NEG subsample ($N_{\text{NEG}}\!=\!30$) is too small to support reliable per-code analysis (the layer totals remain in Table~\ref{tab:full_matrix}).

\subsection{Native-Embedding Robustness}
\label{app:native_embedding}
The main comparison standardizes the embedding encoder to isolate operation-level differences under a common retrieval representation. To assess sensitivity to released defaults, we rerun the Set 1 protocol with each system's native encoder and use the same encoder in the Encoding-Examiner diagnostic retrieval. As shown in Table~\ref{tab:native_embedding}, systems with a changed encoder shift in accuracy by $1.2$--$6.8$\,pp, while Retrieval remains the most frequently attributed layer; TiMem uses the standardized encoder natively. The rerun changes only the embedding encoder and is not a complete study of every released native configuration.
\begin{table*}[t]
\centering
\scriptsize
\setlength{\tabcolsep}{3pt}
\renewcommand{\arraystretch}{1.02}
\resizebox{0.98\textwidth}{!}{%
\begin{tabular}{l l c c c c}
\toprule
\textbf{System} & \textbf{Released/native embedding} & \textbf{Acc S/N} & $\Delta$\textbf{Acc} & \textbf{Enc/Ret/Gen (S)} & \textbf{Enc/Ret/Gen (N)} \\
\midrule
TiMem      & \makecell[l]{Qwen3-Embedding-0.6B\\(same as standardized)} & $75.3/75.3$ & $0.0$  & $7.1/17.0/6.0$  & $7.1/17.0/6.0$ \\
O-Mem      & all-MiniLM-L6-v2       & $74.9/69.8$ & $-5.1$ & $4.9/19.6/6.1$  & $6.2/23.4/6.7$ \\
EverMemOS  & Qwen3-Embedding-4B    & $72.4/76.0$ & $+3.6$ & $8.6/20.0/6.3$  & $7.5/17.5/5.9$ \\
MemOS      & nomic-embed-text:latest & $70.1/72.9$ & $+2.8$ & $8.1/21.0/6.5$  & $7.2/18.8/6.1$ \\
MemBox     & text-embedding-3-small & $63.2/56.4$ & $-6.8$ & $5.9/25.7/6.9$  & $7.5/30.8/7.6$ \\
MemoryOS   & all-MiniLM-L6-v2       & $60.8/63.2$ & $+2.4$ & $9.3/26.0/6.6$  & $8.5/23.7/6.3$ \\
GAM        & BGE-M3                 & $60.2/61.4$ & $+1.2$ & $10.1/25.7/7.1$ & $9.7/24.6/6.9$ \\
\bottomrule
\end{tabular}%
}
\caption{Native-embedding robustness on \textsc{LoCoMo} with \textsc{gpt-4o-mini}. S/N denotes the standardized/native embedding runs; $\Delta$ is native minus standardized accuracy in percentage points. Only the embedding encoder used by the system's embedding-based retrieval/index and the corresponding Encoding-Examiner diagnostic retrieval is changed; its vectors/index entries are rebuilt as needed, while the answer backbone, Top-$K$, EvalMem judges, memory construction, index type/structure, retrieval logic, and all other settings remain fixed. TiMem uses the standardized encoder natively, and Retrieval remains the most frequently attributed layer in every row. Layer rates are unions assigned only to incorrectly answered outputs, with the full-query denominator. This is an embedding-only ablation, not a complete per-system native-configuration study.}
\label{tab:native_embedding}
\end{table*}

\paragraph{Backbone sensitivity is bounded and structurally consistent.}
Replacing \textsc{gpt-4o-mini} with \textsc{gpt-4.1} reduces mean Final Acc gap by $+2.8$/$+4.7$/$+3.7$\,pp on the three datasets, with the gain absorbed across all three layers (Enc $-1.5$/$-1.4$/$-1.4$\,pp; Ret $-1.9$/$-3.1$/$-1.0$\,pp; Gen $-1.9$/$-1.2$/$-1.3$\,pp). \textsc{Qwen3-30B} sits within $\pm 1.3$\,pp of the default on Acc and within $\pm 1$\,pp on every layer. The descriptive aggregate pattern is preserved across all three backbones: Retrieval has the highest attributed rate in every evaluated (dataset, backbone) cell, under the full-query denominator and the error gate in \S\ref{sec:framework:examiners}.

\paragraph{Ranking stability.}
Spearman rank correlation between any two backbones on a single dataset ranges from $0.929$ to $0.964$. Two near-tied flips appear: TiMem $\leftrightarrow$ O-Mem under \textsc{gpt-4.1} on \textsc{LoCoMo} and \datasetname (both pairs separated by $\le 0.6$\,pp); MemoryOS $\leftrightarrow$ MemBox under \textsc{Qwen3-30B} on \textsc{LoCoMo} (separated by $0.6$\,pp). The replication of the TiMem $\leftrightarrow$ O-Mem flip on the dynamic dataset suggests it is driven by \textsc{gpt-4.1}'s ability to exploit O-Mem's entity-graph index.

\paragraph{Stronger backbones disproportionately help generation.}
On \textsc{LoCoMo}, the largest relative shrinkage from \textsc{gpt-4o-mini} to \textsc{gpt-4.1} is in the Gen layer ($-29\%$, from $6.5$ to $4.6$), then Enc ($-19\%$, from $7.7$ to $6.2$), then Ret ($-9\%$, from $22.1$ to $20.2$). This is consistent with generation being the layer most directly exercised by backbone capability, whereas encoding and retrieval are largely the memory system's own responsibility.

\subsection{Threshold Calibration and Sensitivity}
\label{app:threshold_calibration}
The thresholds $\tau_{\mathrm{rank}}$, $\tau_{\mathrm{snr}}$, and $\tau_H$ affect only the Retrieval subcodes LATE, NOI, and NIR; they do not change Encoding, Generation, or RF assignments. We selected them with system-stratified human calibration on fixed pilot samples before aggregating test results. Table~\ref{tab:threshold_calibration} reports one-at-a-time sweeps and calibration macro-F1; the selected values are $5$, $0.20$, and $0.55$, respectively.
\begin{table*}[t]
\centering
\scriptsize
\setlength{\tabcolsep}{4pt}
\renewcommand{\arraystretch}{1.05}
\begin{tabular}{l l c c c}
\toprule
\textbf{Threshold} & \textbf{Candidates} & \textbf{Controlled sub-code rate (\%)} & \textbf{Calibration macro-F1} & \textbf{Selected} \\
\midrule
$\tau_{\mathrm{rank}}$ & $3/5/7/9$ & $6.3/3.8/2.6/1.7$ & $0.78/0.84/0.76/0.58$ & $5$ \\
$\tau_{\mathrm{snr}}$  & $0.10/0.15/0.20/0.25/0.30$ & $1.8/2.9/4.4/6.0/8.1$ & $0.61/0.75/0.82/0.79/0.74$ & $0.20$ \\
$\tau_H$               & $0.45/0.50/0.55/0.60/0.65$ & $3.2/2.5/2.0/1.6/1.3$ & $0.79/0.82/0.84/0.78/0.69$ & $0.55$ \\
\bottomrule
\end{tabular}
\caption{Threshold sensitivity and calibration under the Set 1 protocol. One threshold is varied at a time; the controlled sub-code rate is the unweighted mean over seven systems. LATE/NOI calibration uses 200 erroneous POS--HIT pairs and NIR calibration uses 200 NEG pairs. Two annotators label independently and a third adjudicates disagreements; the selected value maximizes calibration macro-F1.}
\label{tab:threshold_calibration}
\end{table*}

\subsection{Five-run Diagnostic Stability}
\label{app:diagnostic_stability}
To isolate stochasticity in the evaluator, we hold the baseline systems' \textsc{LoCoMo} outputs fixed and rerun the Encoding, Retrieval, and Generation Examiners together with the Attribution Agent five times under the Set~1 configuration. Table~\ref{tab:diagnostic_stability} reports the mean and standard deviation of each layer rate in percentage points and Fleiss' $\kappa$ across the five run-level diagnoses.
\begin{table*}[t]
\centering
\small
\setlength{\tabcolsep}{7pt}
\renewcommand{\arraystretch}{1.04}
\begin{tabular}{lcccc}
\toprule
\textbf{System} & \textbf{Enc mean$\pm$SD} & \textbf{Ret mean$\pm$SD} & \textbf{Gen mean$\pm$SD} & \textbf{Fleiss' $\kappa$} \\
\midrule
TiMem      & $7.16\!\pm\!0.17$  & $16.84\!\pm\!0.22$ & $6.09\!\pm\!0.11$ & $0.93$ \\
O-Mem      & $4.97\!\pm\!0.14$  & $19.72\!\pm\!0.20$ & $6.05\!\pm\!0.10$ & $0.91$ \\
EverMemOS  & $8.53\!\pm\!0.19$  & $20.17\!\pm\!0.23$ & $6.39\!\pm\!0.11$ & $0.93$ \\
MemOS      & $8.18\!\pm\!0.18$  & $20.91\!\pm\!0.21$ & $6.57\!\pm\!0.13$ & $0.94$ \\
MemBox     & $6.03\!\pm\!0.16$  & $25.86\!\pm\!0.25$ & $6.83\!\pm\!0.12$ & $0.92$ \\
MemoryOS   & $9.37\!\pm\!0.21$  & $26.12\!\pm\!0.24$ & $6.77\!\pm\!0.14$ & $0.89$ \\
GAM        & $10.05\!\pm\!0.20$ & $25.82\!\pm\!0.25$ & $7.16\!\pm\!0.13$ & $0.90$ \\
\bottomrule
\end{tabular}
\caption{Five-run diagnostic stability under the Set~1 configuration (\textsc{LoCoMo} with \textsc{gpt-4o-mini}). Baseline systems' outputs are held fixed while the full EvalMem diagnostic pipeline is rerun five times. Layer values are mean$\pm$SD in percentage points, using the error gate and full-query denominator in \S\ref{sec:framework:examiners}; Fleiss' $\kappa$ measures agreement across the five run-level diagnoses. This is evaluator-run stochasticity, not memory-system training or evaluation-seed variance.}
\label{tab:diagnostic_stability}
\end{table*}

For auditability, Tables~\ref{tab:diagnostic_stability_encoding}--\ref{tab:diagnostic_stability_generation}
resolve the same layer summaries into their constituent defect codes. All entries are
mean$\!\pm\!$SD in percentage points over the five evaluator reruns; the final column in
each table is the corresponding layer total.
\begin{table*}[t]
\centering
\small
\setlength{\tabcolsep}{7pt}
\renewcommand{\arraystretch}{1.04}
\begin{tabular}{lccccc}
\toprule
\textbf{System} & \textbf{EM} & \textbf{EA} & \textbf{EW} & \textbf{DMP} & \textbf{Enc total} \\
\midrule
TiMem      & $3.45\!\pm\!0.21$ & $1.67\!\pm\!0.22$ & $1.31\!\pm\!0.20$ & $0.73\!\pm\!0.15$ & $7.16\!\pm\!0.17$ \\
O-Mem      & $2.11\!\pm\!0.05$ & $1.38\!\pm\!0.19$ & $1.01\!\pm\!0.04$ & $0.47\!\pm\!0.05$ & $4.97\!\pm\!0.14$ \\
EverMemOS  & $4.51\!\pm\!0.16$ & $1.96\!\pm\!0.09$ & $1.34\!\pm\!0.07$ & $0.72\!\pm\!0.04$ & $8.53\!\pm\!0.19$ \\
MemOS      & $4.14\!\pm\!0.14$ & $1.69\!\pm\!0.23$ & $1.54\!\pm\!0.18$ & $0.81\!\pm\!0.16$ & $8.18\!\pm\!0.18$ \\
MemBox     & $2.31\!\pm\!0.10$ & $1.43\!\pm\!0.08$ & $1.21\!\pm\!0.20$ & $1.09\!\pm\!0.19$ & $6.03\!\pm\!0.16$ \\
MemoryOS   & $4.60\!\pm\!0.15$ & $2.33\!\pm\!0.10$ & $1.65\!\pm\!0.16$ & $0.79\!\pm\!0.05$ & $9.37\!\pm\!0.21$ \\
GAM        & $5.18\!\pm\!0.11$ & $2.37\!\pm\!0.23$ & $1.42\!\pm\!0.15$ & $1.09\!\pm\!0.08$ & $10.05\!\pm\!0.20$ \\
\bottomrule
\end{tabular}
\caption{Five-run diagnostic stability for the Encoding layer under Set~1. Each cell is a mean$\!\pm\!$SD in percentage points over the same five evaluator reruns; the four code columns are the resolution of the Enc total. The error gate and full-query denominator are those in \S\ref{sec:framework:examiners}.}
\label{tab:diagnostic_stability_encoding}
\end{table*}

\begin{table*}[t]
\centering
\small
\setlength{\tabcolsep}{7pt}
\renewcommand{\arraystretch}{1.04}
\begin{tabular}{lccccc}
\toprule
\textbf{System} & \textbf{RF} & \textbf{LATE} & \textbf{NOI} & \textbf{NIR} & \textbf{Ret total} \\
\midrule
TiMem      & $9.30\!\pm\!0.22$ & $2.71\!\pm\!0.08$ & $3.30\!\pm\!0.29$ & $1.53\!\pm\!0.18$ & $16.84\!\pm\!0.22$ \\
O-Mem      & $10.56\!\pm\!0.18$ & $3.11\!\pm\!0.27$ & $3.90\!\pm\!0.27$ & $2.15\!\pm\!0.15$ & $19.72\!\pm\!0.20$ \\
EverMemOS  & $10.43\!\pm\!0.13$ & $3.21\!\pm\!0.31$ & $4.13\!\pm\!0.31$ & $2.40\!\pm\!0.27$ & $20.17\!\pm\!0.23$ \\
MemOS      & $11.20\!\pm\!0.23$ & $3.58\!\pm\!0.07$ & $4.21\!\pm\!0.28$ & $1.92\!\pm\!0.24$ & $20.91\!\pm\!0.21$ \\
MemBox     & $14.89\!\pm\!0.32$ & $4.49\!\pm\!0.09$ & $4.33\!\pm\!0.09$ & $2.15\!\pm\!0.10$ & $25.86\!\pm\!0.25$ \\
MemoryOS   & $13.87\!\pm\!0.27$ & $4.70\!\pm\!0.32$ & $4.98\!\pm\!0.08$ & $2.57\!\pm\!0.29$ & $26.12\!\pm\!0.24$ \\
GAM        & $14.46\!\pm\!0.32$ & $4.59\!\pm\!0.34$ & $5.46\!\pm\!0.17$ & $1.31\!\pm\!0.20$ & $25.82\!\pm\!0.25$ \\
\bottomrule
\end{tabular}
\caption{Five-run diagnostic stability for the Retrieval layer under Set~1. Each cell is a mean$\!\pm\!$SD in percentage points over the same five evaluator reruns; the four code columns are the resolution of the Ret total. The error gate and full-query denominator are those in \S\ref{sec:framework:examiners}.}
\label{tab:diagnostic_stability_retrieval}
\end{table*}

\begin{table*}[t]
\centering
\small
\setlength{\tabcolsep}{7pt}
\renewcommand{\arraystretch}{1.04}
\begin{tabular}{lcccc}
\toprule
\textbf{System} & \textbf{GF} & \textbf{GRF} & \textbf{GH} & \textbf{Gen total} \\
\midrule
TiMem      & $2.13\!\pm\!0.08$ & $2.70\!\pm\!0.14$ & $1.26\!\pm\!0.12$ & $6.09\!\pm\!0.11$ \\
O-Mem      & $2.31\!\pm\!0.10$ & $2.59\!\pm\!0.10$ & $1.16\!\pm\!0.11$ & $6.05\!\pm\!0.10$ \\
EverMemOS  & $2.56\!\pm\!0.11$ & $2.43\!\pm\!0.09$ & $1.41\!\pm\!0.05$ & $6.39\!\pm\!0.11$ \\
MemOS      & $2.51\!\pm\!0.11$ & $2.74\!\pm\!0.12$ & $1.32\!\pm\!0.17$ & $6.57\!\pm\!0.13$ \\
MemBox     & $2.70\!\pm\!0.06$ & $2.65\!\pm\!0.12$ & $1.48\!\pm\!0.16$ & $6.83\!\pm\!0.12$ \\
MemoryOS   & $2.22\!\pm\!0.19$ & $2.90\!\pm\!0.13$ & $1.65\!\pm\!0.19$ & $6.77\!\pm\!0.14$ \\
GAM        & $2.67\!\pm\!0.17$ & $2.75\!\pm\!0.09$ & $1.74\!\pm\!0.17$ & $7.16\!\pm\!0.13$ \\
\bottomrule
\end{tabular}
\caption{Five-run diagnostic stability for the Generation layer under Set~1. Each cell is a mean$\!\pm\!$SD in percentage points over the same five evaluator reruns; the three code columns are the resolution of the Gen total. The error gate and full-query denominator are those in \S\ref{sec:framework:examiners}.}
\label{tab:diagnostic_stability_generation}
\end{table*}

\subsection{Diagnostic Token, Call, and Latency Overhead}
\label{app:diagnostic_overhead}
We measure per-example EvalMem diagnostic overhead under the Set~1 protocol; $K$ denotes $10^3$ tokens and p50/p95 are per-example critical-path latencies. Table~\ref{tab:diagnostic_overhead_system} gives per-system totals, while Table~\ref{tab:diagnostic_overhead_component} decomposes the mean into Agentic-RAG (AR), the Encoding judge (EJ), Retrieval, Generation, and Attribution. AR and EJ together form the Encoding path. The three Examiners execute in parallel and Attribution follows them; hence the critical path is approximately $\max(t_E,t_R,t_G)+t_A$, where AR is an internal component of $t_E$, rather than the sum of all listed latencies.
\begin{table*}[t]
\centering
\small
\setlength{\tabcolsep}{6pt}
\renewcommand{\arraystretch}{1.04}
\begin{tabular}{lccccc}
\toprule
\textbf{System} & \textbf{Calls} & \textbf{Input (K)} & \textbf{Output (K)} & \multicolumn{2}{c}{\textbf{Critical latency (s)}} \\
 & & & & \textbf{p50} & \textbf{p95} \\
\midrule
TiMem      & 8.24 & 38.0 & 1.81 & 11.4 & 25.4 \\
O-Mem      & 8.38 & 40.5 & 1.88 & 11.9 & 26.4 \\
EverMemOS  & 8.42 & 44.0 & 1.97 & 12.4 & 27.7 \\
MemOS      & 8.36 & 43.0 & 1.95 & 12.2 & 27.3 \\
MemBox     & 8.64 & 53.0 & 2.18 & 13.8 & 30.8 \\
MemoryOS   & 8.52 & 50.0 & 2.12 & 13.4 & 29.9 \\
GAM        & 8.80 & 62.0 & 2.36 & 15.6 & 34.6 \\
\midrule
\textit{Mean} & 8.48 & 47.2 & 2.04 & 13.0 & 28.9 \\
\bottomrule
\end{tabular}
\caption{Per-example EvalMem diagnostic overhead by memory system under the Set~1 protocol. Input/output values are in $K=10^3$ tokens; calls and token counts are means, and critical latency is the per-example p50/p95 on the parallel-Examiner path followed by Attribution. The mean row is the unweighted mean over seven systems.}
\label{tab:diagnostic_overhead_system}
\end{table*}

\begin{table*}[t]
\centering
\small
\setlength{\tabcolsep}{6pt}
\renewcommand{\arraystretch}{1.04}
\begin{tabular}{lccccc}
\toprule
\textbf{Component} & \textbf{Calls} & \textbf{Input (K)} & \textbf{Output (K)} & \multicolumn{2}{c}{\textbf{Latency (s)}} \\
 & & & & \textbf{p50} & \textbf{p95} \\
\midrule
Agentic-RAG (AR)      & 3.48 & 21.8 & 0.94 & 8.2 & 18.7 \\
Encoding judge (EJ)   & 1.00 & 11.6 & 0.23 & 3.2 & 6.8 \\
Retrieval              & 1.00 & 8.4  & 0.31 & 3.1 & 6.9 \\
Generation             & 2.00 & 3.9  & 0.39 & 4.0 & 8.5 \\
Attribution            & 1.00 & 1.44 & 0.16 & 1.5 & 3.4 \\
\bottomrule
\end{tabular}
\caption{Mean component-level EvalMem diagnostic overhead over seven systems under Set~1. Input/output values are in $K=10^3$ tokens. Rounded component means sum to the per-system mean up to rounding. Agentic-RAG (AR) and the Encoding judge (EJ) form the Encoding path; the three Examiners are parallel, so component p50/p95 values are not additive for the critical path.}
\label{tab:diagnostic_overhead_component}
\end{table*}

\paragraph{Mechanism-level signals from the sub-code view.}
The sub-code resolution makes several mechanism-level shifts visible. On \textsc{LoCoMo} the TiMem $\leftrightarrow$ O-Mem flip under \textsc{gpt-4.1} is driven by O-Mem's lower EM/EA/EW ($1.5$/$1.0$/$0.7$ vs.\ TiMem's $2.5$/$1.4$/$1.0$), not by retrieval differences (TiMem's RF $8.1$ is in fact lower than O-Mem's $9.5$); the rank-1 system flips because the entity-graph encoding savings outweigh the retrieval gap once the backbone is strong enough to exploit them. On \datasetname the NIR code carries a disproportionate share of retrieval defects across every backbone ($2.3$--$4.6\%$, compared with $\le 2.4\%$ on \textsc{LoCoMo}), consistent with rollback / cascade events producing stale but relevant-looking evidence.

\subsection{Adapted \memscan Mechanism Detail}
\label{app:memscan_detail}
\label{app:mechanism_detail}

\begin{figure*}[t]
  \centering
  \includegraphics[width=\textwidth]{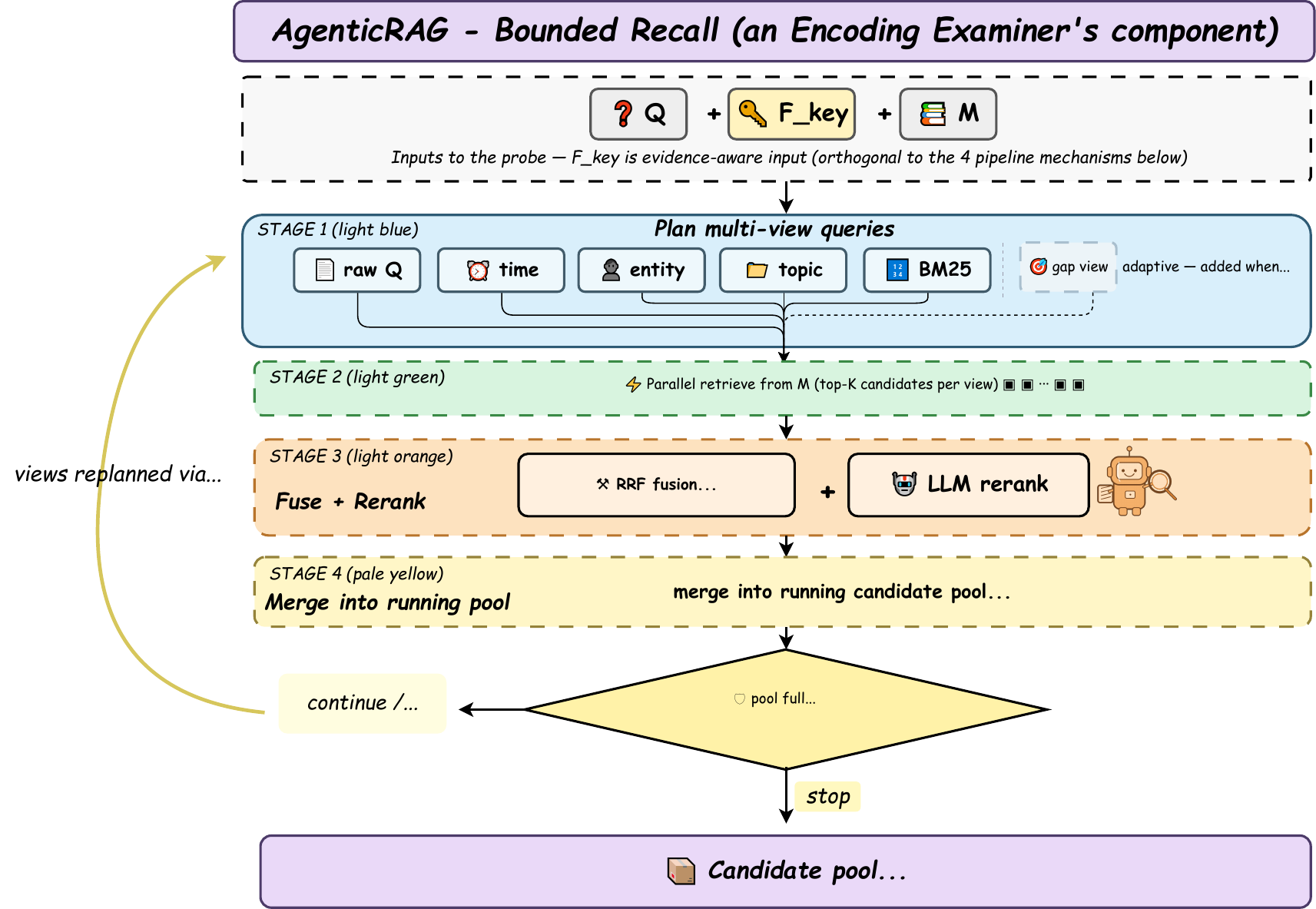}
  \caption{Adapted \memscan{} diagnostic pipeline. For each $(Q, F_{\text{key}})$, the agent plans a set of views (dense, BM25, time, entity, topic, plus an on-demand gap view), routes a granularity-aware top-$K$ per view, fuses with RRF, and reranks by evidence to $F_{\text{key}}$. The loop terminates by pool-saturation, Jaccard stability, or $T_{\max}$, so it bounds observation MISS without false confidence on truly absent facts.}
  \label{fig:memscan}
  \vspace{-4mm}
\end{figure*}

\begin{table}[!t]
\centering
\small
\setlength{\tabcolsep}{4pt}
\renewcommand{\arraystretch}{1.05}
\begin{tabular}{@{}l c c@{}}
\toprule
\textbf{Configuration} & \textbf{MISS\,/\,1540} & \textbf{Rate} \\
\midrule
\multicolumn{3}{c}{\emph{Controlled study (oracle vector store)}}\\
\midrule
$C_1$ Vanilla RAG\,+\,$Q$                          & 459 & 29.8\% \\
$C_2$ \memscan\,+\,$Q$                             & 283 & 18.4\% \\
$C_3$ Vanilla RAG\,+\,$Q$\,+\,$F_{\text{key}}$     & 174 & 11.3\% \\
$C_4$ \memscan\,+\,$Q$\,+\,$F_{\text{key}}$ ($\star$) & \textbf{67}  & \textbf{4.4\%} \\
\midrule
\multicolumn{3}{c}{\emph{Leave-one-out from $C_4$}}\\
\midrule
$-$ evidence-aware query ($\to C_2$)               & 283 & 18.4\% \\
$-$ multi-view (dense $\to 1$)                     & 107 & \phantom{0}6.9\% \\
$-$ LLM rerank                                     & \phantom{0}99 & \phantom{0}6.4\% \\
$-$ multi-round ($T_{\max}\!\to\!1$)               & \phantom{0}89 & \phantom{0}5.8\% \\
$-$ BM25 sparse view                               & \phantom{0}78 & \phantom{0}5.1\% \\
\bottomrule
\end{tabular}
\caption{Adapted \memscan ablation on the \textsc{LoCoMo} POS subset ($N\!=\!1540$). ($\star$)~$C_4$ is the default in all main experiments. The adapted agentic-RAG routine comprises four pipeline mechanisms (multi-view dense retrieval, BM25 sparse view, LLM rerank, multi-round); evidence-aware query is an orthogonal input-level mechanism. Leave-one-out contributions: evidence-aware $+14.0$, multi-view $+2.5$, rerank $+2.0$, multi-round $+1.4$, BM25 $+0.7$\,pp. McNemar $C_1$ vs.\ $C_4$ $\chi^2\!=\!298.4$ ($p\!<\!10^{-3}$); $C_3$ vs.\ $C_4$ $\chi^2\!=\!41.2$ ($p\!<\!10^{-3}$). The four-mechanism leave-one-out sum ($6.6$\,pp) is close to the net pipeline contribution ($C_4$ vs.\ $C_3$, $6.9$\,pp), indicating near-additivity with $0.3$\,pp residual overlap.}
\label{tab:app_memscan}
\end{table}

The controlled study is run on an \emph{oracle vector store}: the full \textsc{LoCoMo10} dialogue is chunked at a fixed turn-level granularity and indexed with Qwen3-Embedding-0.6B, decoupling the diagnostic routine from any tested system's ingestion pipeline so that retrieval-side effects can be isolated from storage-side ones.

Table~\ref{tab:app_memscan} reports the full $C_1$--$C_4$ controlled study together with the leave-one-out breakdown of $C_4$. The four conditions disentangle two orthogonal ingredients: (i)~the adapted agentic-RAG pipeline mechanisms---multi-view dense retrieval, BM25 sparse view, LLM rerank, and multi-round refinement ($C_2, C_4$) vs.\ a vanilla single-view dense retriever ($C_1, C_3$); (ii)~the evidence-aware query that injects the oracle key fact $F_{\text{key}}$ into the query ($C_3, C_4$) vs.\ the raw user question $Q$ alone ($C_1, C_2$). The two ingredients are nearly independent: the pipeline mechanisms contribute $-11.4$\,pp (between $C_1$ and $C_2$) and $-6.9$\,pp (between $C_3$ and $C_4$). McNemar tests confirm both improvements are highly significant.

\paragraph{Leave-one-out near-additivity.}
The leave-one-out section removes each ingredient from $C_4$ in isolation. The evidence-aware query is the single largest contributor at $+14.0$\,pp when removed (which reverts $C_4$ exactly back to the $C_2$ MISS rate). The four adapted agentic-RAG pipeline mechanisms---multi-view dense retrieval, the LLM reranker, the multi-round loop, and the BM25 sparse view---contribute $+2.5$, $+2.0$, $+1.4$, and $+0.7$\,pp respectively. Their sum is $6.6$\,pp; the net contribution of the four mechanisms over $C_3$ is $6.9$\,pp. The $0.3$\,pp gap indicates the four pipeline mechanisms are near-additive, each covering an orthogonal failure mode.

\paragraph{Convergence statistics.}
Across the $1540$ POS queries with $T_{\max}\!=\!4$, the mean number of rounds taken before termination is $1.74$ ($\sigma\!=\!0.81$). $87.9\%$ of queries terminate via convergence ($62.9\%$ by the new-ratio criterion $\epsilon\!<\!0.10$, $25.0\%$ by the Jaccard criterion $\ge\!0.85$ for two consecutive rounds); the remaining $12.1\%$ hit $T_{\max}$ and are concentrated on multi-hop temporal questions where the key fact spans multiple sessions. The pool-full stop does not fire on this set.

\subsection{\memwiki Mechanism Detail}
\label{app:memwiki}

\begin{figure*}[t]
  \centering
  \includegraphics[width=\textwidth]{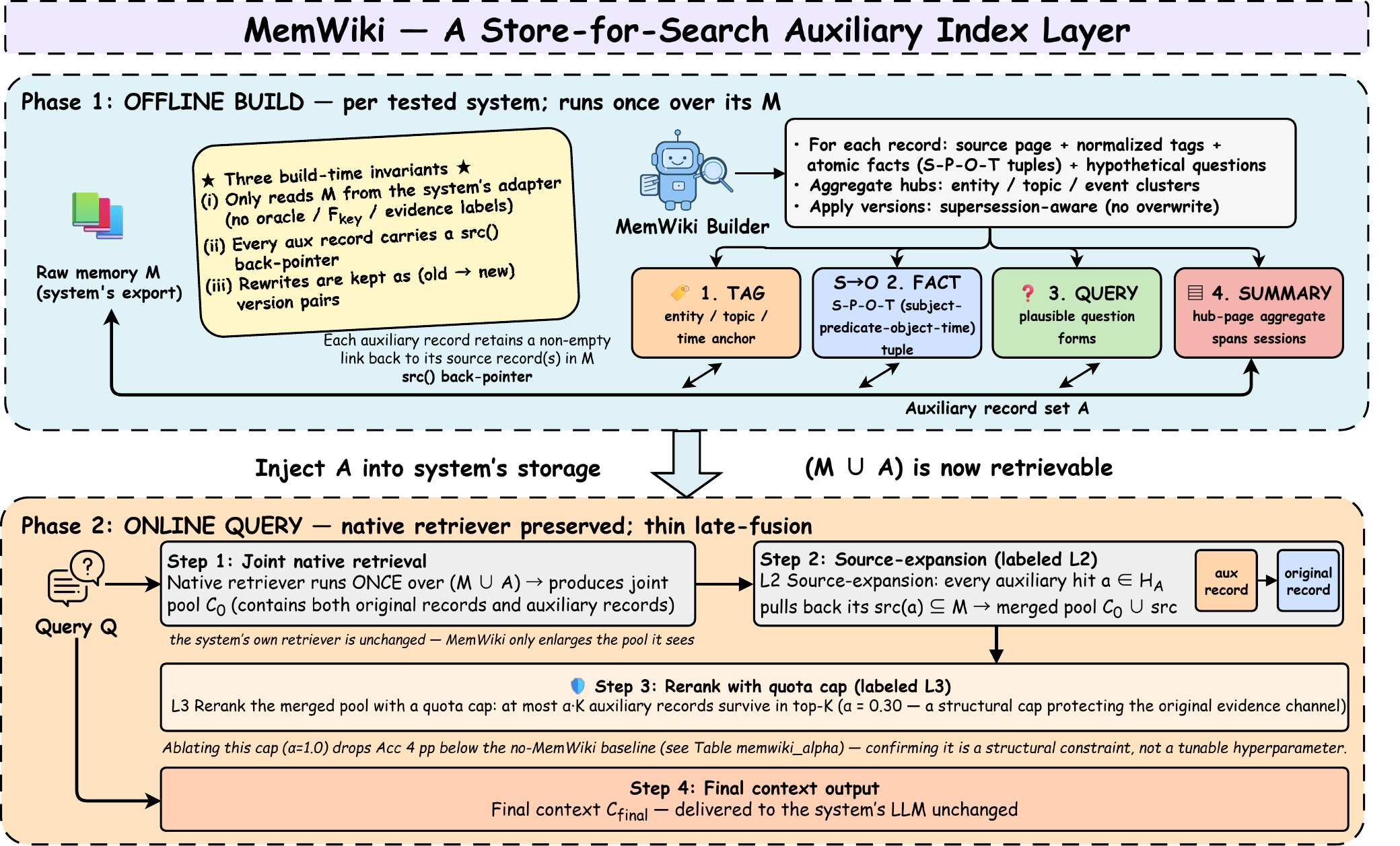}
  \caption{\memwiki{} as a store-for-search auxiliary layer. The raw memory $\mathcal{M}$ is organised into a wiki-style index whose source pages and hub pages are flattened into four AuxRecord classes (factual, tag, query-shaped, aggregate). The system's native retriever runs over $\mathcal{M}\!\cup\!\mathcal{A}$; any auxiliary hit pulls in its source records via $\mathrm{src}(\cdot)$, and a single rerank pass orders the merged pool subject to a $30\%$ auxiliary-record quota.}
  \label{fig:memwiki}
\end{figure*}

\begin{table*}[t]
\centering
\small
\setlength{\tabcolsep}{3pt}
\renewcommand{\arraystretch}{1.05}
\begin{tabular}{lccccc|cc}
\toprule
\textbf{System} & \textsc{Tag} & \textsc{Fact} & \textsc{Query} & \textsc{Sum.} & \textbf{Sum} & \textbf{All} & \textbf{Crowd loss} \\
\midrule
TiMem      & +0.1 & +0.4 & +0.2 & +0.1 & +0.8 & +0.3 & $-62.5\%$ \\
O-Mem      & +0.4 & +0.9 & +0.6 & +0.5 & +2.4 & +1.5 & $-37.5\%$ \\
EverMemOS  & +0.2 & +0.5 & +0.4 & +0.3 & +1.4 & +1.0 & $-28.6\%$ \\
MemOS      & +0.6 & +1.5 & +1.1 & +0.8 & +4.0 & +2.6 & $-35.0\%$ \\
MemBox     & +1.1 & +2.6 & +1.7 & +1.4 & +6.8 & +4.2 & $-38.2\%$ \\
MemoryOS   & +0.8 & +1.8 & +1.3 & +1.0 & +4.9 & +3.0 & $-38.8\%$ \\
GAM        & +1.2 & +2.7 & +1.9 & +1.5 & +7.3 & +4.6 & $-37.0\%$ \\
\bottomrule
\end{tabular}
\caption{\memwiki AuxRecord single-class ablation ($\Delta$ Final Acc, pp). \textsc{Tag}=tag records, \textsc{Fact}=factual records, \textsc{Query}=query-shaped records, \textsc{Sum.}=aggregate summaries. \textbf{Sum}=sum of singletons; \textbf{All}=all four enabled; \textbf{Crowd loss}=$1\!-\!\text{All}/\text{Sum}$, the share of gain dropped to the $30\%$ auxiliary quota.}
\label{tab:memwiki_aux_ablation}
\end{table*}

Table~\ref{tab:memwiki_aux_ablation} reports the four AuxRecord classes turned on one at a time, alongside their sum, the all-four configuration (used in the main paper), and the resulting \emph{crowd loss}: the share of singleton gain that is dropped when all four classes compete for the same $30\%$ quota. Three observations follow.

\paragraph{Factual records dominate.}
For every system, the factual class is the largest singleton contributor (between $+0.4$ and $+2.7$\,pp). Tag and aggregate-summary classes contribute less because they encode coarser information that often already appears (in some form) inside the system's native records.

\paragraph{Crowding is real but modest.}
The crowd loss is $-28.6\%$ to $-38.8\%$ on the six bottom-row systems --- meaningful, but the all-four configuration still recovers the bulk of singleton gain. Only TiMem suffers a much larger $-62.5\%$ crowd loss: its native top-level layer (\emph{L5} persona profile, the highest tier of TiMem's L1--L5 temporal hierarchy that aggregates session content into long-horizon summaries) is already similar to the aggregate-summary class, so the AuxRecord quota largely duplicates information rather than expanding it.

\paragraph{Where \memwiki helps most.}
The systems with the largest empirical \memwiki gains ($\text{MemBox}\!+\!4.2$, $\text{GAM}\!+\!4.6$) are also those with the highest observed native retrieval-defect rates (RF + LATE) in Table~\ref{tab:main_results}. This association is consistent with the auxiliary index being useful when a system's original store lacks query-shaped or factual decomposition.

\providecommand{\BR}[1]{\textbf{\textcolor{red}{#1}}}

\begin{table}[t]
\centering
\small
\setlength{\tabcolsep}{3.5pt}
\renewcommand{\arraystretch}{1.0}
\begin{tabular}{lrrrrrc}
\toprule
\textbf{System} & $b$ & $c$ & $c\!-\!b$ & $\chi^2$ & $p$-value & sig. \\
\midrule
TiMem      &  19 &  25 &    6 & ~~0.82 & 0.366 & --- \\
O-Mem      &  20 &  50 &   30 & 12.86  & $<\!10^{-3}$ & $\star\star\star$ \\
EverMemOS  &  20 &  40 &   20 & ~~6.67 & 0.010        & $\star$ \\
MemOS      &  28 &  80 &   52 & 25.04  & $<\!10^{-3}$ & $\star\star\star$ \\
MemBox     &  47 & 130 &   83 & 38.92  & $<\!10^{-3}$ & $\star\star\star$ \\
MemoryOS   &  38 &  98 &   60 & 26.47  & $<\!10^{-3}$ & $\star\star\star$ \\
GAM        &  58 & 149 &   91 & 40.00  & $<\!10^{-3}$ & $\star\star\star$ \\
\midrule
\textbf{Aggregate} & \textbf{230} & \textbf{572} & \textbf{342} & \textbf{145.84} & $<\!10^{-3}$ & $\star\star\star$ \\
\bottomrule
\end{tabular}
\caption{Per-system McNemar tests for \memwiki on \textsc{LoCoMo} ($N\!=\!1986$ paired observations per system). $b$ = baseline correct $\wedge$ \memwiki wrong (\emph{spoils}); $c$ = baseline wrong $\wedge$ \memwiki correct (\emph{recovers}); $\chi^2\!=\!(b\!-\!c)^2/(b\!+\!c)$ with $df\!=\!1$. Symbols: $\star$: $p<0.05$; $\star\star\star$: $p<10^{-3}$. Six of seven systems show significant improvement; TiMem's marginal $+0.3$\,pp gain is not significant, consistent with its $-62.5\%$ crowd loss (Table~\ref{tab:memwiki_aux_ablation}).}
\label{tab:memwiki_mcnemar}
\end{table}

\paragraph{Per-system significance of \memwiki gains.}
Table~\ref{tab:memwiki_mcnemar} reports per-system McNemar tests between baseline and \memwiki on \textsc{LoCoMo}. For each system, the $1986$ paired observations are partitioned by the four McNemar cells, of which only the discordant pair counts $b$ (\memwiki spoils a previously correct answer) and $c$ (\memwiki recovers a previously wrong answer) contribute to the test statistic $\chi^2\!=\!(b\!-\!c)^2/(b\!+\!c)$. The null hypothesis is $b\!=\!c$, i.e.\ \memwiki neither helps nor hurts on aggregate. Six of seven systems show statistically significant improvement ($p<0.05$, with five at $p<10^{-3}$); TiMem is the only exception ($\chi^2\!=\!0.82$, $p\!=\!0.37$), reflecting that its native top-level (L5) persona-profile layer already covers most of the indexing surface that \memwiki would otherwise add (the same mechanism documented as the $-62.5\%$ crowd loss in Table~\ref{tab:memwiki_aux_ablation}). The Pearson correlation between each system's retrieval defect rate on \textsc{LoCoMo} and its $\log\chi^2$ is $0.92$: \memwiki therefore provides gains that scale with each system's baseline retrieval gap, rather than gains that are uniform across the cohort. The corresponding aggregate $\chi^2$ ($145.8$) is reported in the caption of Table~\ref{tab:memwiki_main}.

\begin{table}[t]
\centering
\small
\setlength{\tabcolsep}{3pt}
\renewcommand{\arraystretch}{1.0}
\begin{tabular}{l|cc|cc}
\toprule
\multirow{2}{*}{\textbf{System}}
 & \multicolumn{2}{c|}{\textbf{Acc}\,$\uparrow$}
 & \multicolumn{2}{c}{\textbf{Ret}\,$\downarrow$} \\
\cmidrule(lr){2-3}\cmidrule(lr){4-5}
 & Bsl & +\,\memwiki & \multicolumn{1}{c}{$\Delta$} & $\Delta$ \\
\midrule
TiMem      & 75.3 & 75.6 & $+0.3$ & $-0.9$ \\
O-Mem      & 74.9 & 76.4 & $+1.5$ & $-2.3$ \\
EverMemOS  & 72.4 & 73.4 & $+1.0$ & $-1.6$ \\
MemOS      & 70.1 & 72.7 & $+2.6$ & $-3.0$ \\
MemBox     & 63.2 & 67.4 & \UB{$+4.2$} & \UB{$-5.0$} \\
MemoryOS   & 60.8 & 63.8 & $+3.0$ & $-4.0$ \\
GAM        & 60.2 & 64.8 & \BR{$+4.6$} & \BR{$-5.8$} \\
\midrule
\textbf{mean} & \textbf{68.1} & \textbf{70.6} & \textbf{$+2.5$} & \textbf{$-3.2$} \\
\bottomrule
\end{tabular}
\caption{\memwiki retrieval-side intervention on \textsc{LoCoMo} with \textsc{gpt-4o-mini}. \memwiki is built per-system from each baseline's own memory export, then re-injected through L1/L2/L3 ($\le\!30\%$ auxiliary quota). Final \textbf{Acc} (\%) and retrieval defect rate \textbf{Ret} (\%) before/after intervention. \BR{Bold red}: largest gain; \UB{blue underline}: second-largest. Aggregate McNemar over $7$ systems $\times 1986$ paired observations: $\chi^2\!=\!145.8$ ($p\!<\!10^{-3}$); per-system breakdown in Appendix~\ref{app:memwiki}, Table~\ref{tab:memwiki_mcnemar}.}
\label{tab:memwiki_main}
\end{table}

\paragraph{Per-system \memwiki intervention.}
Table~\ref{tab:memwiki_main} reports the per-system end-to-end accuracy and retrieval defect rate before and after \memwiki on \textsc{LoCoMo} with \textsc{gpt-4o-mini}. Gains are heterogeneous (GAM $+4.6$ / MemBox $+4.2$ / TiMem $+0.3$\,pp) and track each baseline's retrieval gap; the corresponding $\Delta$ on the retrieval-defect rate is consistently negative for every system. The summary numbers ($+2.5$\,pp mean Acc, $-3.2$\,pp mean Ret defect) appear in the body of \S\ref{sec:exp:mechanism}; this table provides the per-system breakdown.

\providecommand{\BR}[1]{\textbf{\textcolor{red}{#1}}}

\begin{table}[t]
\centering
\small
\setlength{\tabcolsep}{2.8pt}
\renewcommand{\arraystretch}{1.0}
\begin{tabular}{l|c|ccccccc}
\toprule
\multirow{2}{*}{\textbf{System}}
 & \multirow{2}{*}{\textbf{bsl}}
 & \multicolumn{7}{c}{\textbf{Final Acc (\%) under \memwiki quota $\alpha$}} \\
\cmidrule(lr){3-9}
 &      & $.10$ & $.20$ & $.30^{\star}$ & $.40$ & $.50$ & $.70$ & $1.0$ \\
\midrule
TiMem      & 75.3 & 75.5 & \BR{75.7} & 75.6 & 75.3 & 74.8 & 73.4 & 69.8 \\
O-Mem      & 74.9 & 75.4 & 76.2 & \BR{76.4} & 76.1 & 75.5 & 73.9 & 70.4 \\
EverMemOS  & 72.4 & 72.9 & 73.3 & 73.4 & \BR{73.5} & 72.8 & 70.6 & 67.5 \\
MemOS      & 70.1 & 71.0 & 72.1 & \BR{72.7} & 72.6 & 71.9 & 70.2 & 66.5 \\
MemBox     & 63.2 & 64.7 & 66.3 & 67.4 & \BR{67.5} & 66.8 & 64.6 & 60.5 \\
MemoryOS   & 60.8 & 61.8 & 63.0 & 63.8 & \BR{63.9} & 63.1 & 61.0 & 56.9 \\
GAM        & 60.2 & 62.0 & 63.7 & 64.8 & \BR{65.0} & 64.1 & 61.9 & 57.3 \\
\midrule
\textbf{mean} & \textbf{68.1} & 69.0 & 70.0 & \textbf{70.6} & \textbf{70.6} & 69.9 & 67.9 & 64.1 \\
\bottomrule
\end{tabular}
\caption{\memwiki $\alpha$-quota ablation on \textsc{LoCoMo} (gpt-4o-mini). $\alpha$ caps the fraction of \memwiki auxiliary records in the top-$K\!=\!10$ context, so $\alpha K$ is the maximum number of \memwiki slots ($\alpha\!=\!.30$ allows $\le\!3$, $\alpha\!=\!1.0$ removes the cap). ($\star$)~$\alpha\!=\!0.30$ is the default used in all main experiments. \BR{Bold red}: per-system peak. Per-system optima span $\alpha\!\in\!\{0.20, 0.30, 0.40\}$, yet every system at $\alpha\!=\!0.30$ is within $0.2$\,pp of its own peak. The plateau region $\alpha\!\in\![0.30, 0.40]$ gives identical mean Final Acc ($70.6\%$); removing the quota ($\alpha\!=\!1.0$) drops mean Final Acc to $64.1\%$, $4.0$\,pp below the no-\memwiki baseline.}
\label{tab:memwiki_alpha}
\end{table}

\paragraph{$\alpha$-quota robustness.}
Table~\ref{tab:memwiki_alpha} sweeps the auxiliary-record quota $\alpha\!\in\!\{0.10,\ldots,1.00\}$ on \textsc{LoCoMo} with \textsc{gpt-4o-mini}. Per-system optima fall in $\{0.20, 0.30, 0.40\}$, but the variation in Final Acc across this band is at most $0.2$\,pp for every system: the default $\alpha\!=\!0.30$ is therefore within $0.2$\,pp of every per-system peak, and the plateau $\alpha\!\in\![0.30, 0.40]$ gives identical mean Final Acc ($70.6\%$). The $\alpha\!=\!1.0$ row (no quota) drops mean Final Acc to $64.1\%$ --- $4.0$\,pp \emph{below} the no-\memwiki baseline. Without the quota, the rerank promotes auxiliary records past the original evidence, so the retrieval surface becomes net worse. This rules out reading the quota as a tuned hyper-parameter: it is a structural constraint that protects the original evidence channel.

\paragraph{Cross-dataset replication on \textsc{LongMemEval}-S.}
The same \memwiki configuration repeated on \textsc{LongMemEval}-S (Table~\ref{tab:memwiki_longmemeval} in the main text) yields $+2.3$\,pp mean Acc and $-3.2$\,pp retrieval defect, mirroring the \textsc{LoCoMo} result within $0.2$\,pp. Aggregate McNemar $\chi^2\!=\!38.10$ ($p\!<\!10^{-3}$), with five of seven systems significant at $p\!<\!0.05$; the observed association between baseline retrieval defects and accuracy gains is again positive (Pearson $r(\Delta\text{Ret},\Delta\text{Acc})\!=\!0.81$). TiMem is again non-significant in the same direction, consistent with its native L5 redundancy.

\subsection{Top-K Sensitivity Full Data}
\label{app:topk}
\label{app:topk_full}

\providecommand{\BR}[1]{\textbf{\textcolor{red}{#1}}}
\definecolor{kcolsweet}{HTML}{FFF2CC}
\definecolor{kcolmean}{HTML}{ECECEC}

\begin{table*}[t]
\centering
\footnotesize
\setlength{\tabcolsep}{4pt}
\renewcommand{\arraystretch}{1.0}
\begin{tabular}{ll|cccc>{\columncolor{kcolsweet}}c>{\columncolor{kcolsweet}}cccc}
\toprule
\textbf{System} & \textbf{Metric} & $K\!=\!1$ & $K\!=\!2$ & $K\!=\!3$ & $K\!=\!5$ & $K\!=\!8$ & $K\!=\!10$ & $K\!=\!12$ & $K\!=\!15$ & $K\!=\!20$ \\
\midrule
\multirow{2}{*}{TiMem}     & Hit@$K$      & 38.4 & 47.5 & 55.7 & 64.7 & 71.0 & 73.5 & 75.2 & 77.3 & 79.5 \\
                            & Final Acc    & 64.2 & 68.5 & 70.5 & 73.6 & 75.0 & 75.3 & \BR{75.5} & 75.3 & 74.5 \\
\midrule
\multirow{2}{*}{O-Mem}     & Hit@$K$      & 36.4 & 45.5 & 54.3 & 63.2 & 68.7 & 71.1 & 73.0 & 75.2 & 77.7 \\
                            & Final Acc    & 63.8 & 68.0 & 70.2 & 73.0 & 74.7 & \BR{74.9} & 74.8 & 74.0 & 72.6 \\
\midrule
\multirow{2}{*}{EverMemOS} & Hit@$K$      & 34.9 & 43.8 & 52.8 & 61.8 & 67.2 & 69.4 & 71.4 & 73.5 & 76.0 \\
                            & Final Acc    & 61.6 & 65.8 & 67.9 & 70.5 & 72.1 & \BR{72.4} & 72.3 & 71.6 & 70.0 \\
\midrule
\multirow{2}{*}{MemOS}     & Hit@$K$      & 33.7 & 42.4 & 50.7 & 59.7 & 64.8 & 66.7 & 68.7 & 71.0 & 73.3 \\
                            & Final Acc    & 59.7 & 63.7 & 65.9 & 68.5 & \BR{70.4} & 70.1 & 69.8 & 69.0 & 67.7 \\
\midrule
\multirow{2}{*}{MemBox}    & Hit@$K$      & 30.0 & 38.0 & 45.8 & 53.3 & 58.5 & 60.3 & 62.4 & 64.9 & 67.3 \\
                            & Final Acc    & 53.9 & 57.5 & 59.8 & 62.4 & \BR{63.7} & 63.2 & 62.8 & 62.0 & 60.5 \\
\midrule
\multirow{2}{*}{MemoryOS}  & Hit@$K$      & 28.7 & 36.4 & 44.0 & 51.6 & 56.5 & 58.6 & 60.6 & 63.0 & 65.3 \\
                            & Final Acc    & 51.4 & 55.2 & 57.0 & 61.0 & \BR{61.1} & 60.8 & 60.4 & 59.5 & 58.0 \\
\midrule
\multirow{2}{*}{GAM}       & Hit@$K$      & 27.9 & 35.5 & 42.6 & 50.0 & 55.2 & 57.0 & 59.0 & 61.3 & 63.6 \\
                            & Final Acc    & 50.8 & 54.5 & 56.6 & 58.7 & 60.0 & \BR{60.2} & 60.0 & 59.2 & 57.7 \\
\midrule
\rowcolor{kcolmean}\textit{mean} & Hit@$K$ & 32.9 & 41.3 & 49.4 & 57.8 & 63.1 & 65.2 & 67.2 & 69.5 & 71.8 \\
\rowcolor{kcolmean} & Final Acc & 57.9 & 61.9 & 64.0 & 66.8 & \BR{68.1} & \BR{68.1} & 67.9 & 67.2 & 65.9 \\
\bottomrule
\end{tabular}
\caption{Combined top-$K$ sensitivity on \textsc{LoCoMo} with \textsc{gpt-4o-mini}. \textbf{Hit@$K$} (\%, POS subset) is monotone increasing in $K$, whereas \textbf{Final Acc} (\%) peaks at $K\!\in\!\{8, 10\}$ for every system. \BR{Bold red} marks each row's Final Acc peak; the two yellow columns mark the $K\!\in\!\{8, 10\}$ sweet spot shared by Hit@$K$'s elbow and Acc's maximum. Beyond this band, Acc declines although Hit@$K$ keeps rising; the gap is explained by the NOI rate in Table~\ref{tab:topk_ret_layer}. System rankings under both metrics are stable across all $K$ (Spearman $\rho\!\ge\!0.96$ between any two columns).}
\label{tab:topk_combined}
\end{table*}

\begin{table}[t]
\centering
\small
\setlength{\tabcolsep}{3pt}
\renewcommand{\arraystretch}{1.05}
\begin{tabular}{lcccc}
\toprule
$K$ & \textbf{RF} & \textbf{LATE} & \textbf{NOI} & \textbf{Ret. total} \\
\midrule
$1$  & 16.1 & 0.0  & 0.2  & 16.3 \\
$2$  & 14.4 & 0.3  & 0.5  & 15.2 \\
$3$  & 12.5 & 0.7  & 1.0  & 14.2 \\
$5$  & 9.8  & 1.4  & 2.1  & 13.3 \\
$8$  & 6.6  & 2.9  & 3.7  & 13.2 \\
$10$ & 5.0  & 3.4  & 5.5  & 13.9 \\
$12$ & 4.2  & 3.6  & 7.2  & 15.0 \\
$15$ & 3.3  & 3.7  & 9.4  & 16.4 \\
$20$ & 2.6  & 3.7  & 12.5 & 18.8 \\
\bottomrule
\end{tabular}
\caption{Retrieval-layer code rates (\%) on \textsc{LoCoMo} POS as a function of top-$K$ (mean over $7$ systems). \textbf{RF} (\emph{retrieval failure}: key fact never recalled) decays monotonically as more candidates are fetched. \textbf{LATE} (key fact recalled below rank-$5$) and \textbf{NOI} (low signal-to-noise ratio) grow with $K$. The total retrieval defect rate is U-shaped, with a flat minimum at $K\!\in\!\{8, 10\}$ --- the same range that maximises Final Acc in Table~\ref{tab:topk_combined}.}
\label{tab:topk_ret_layer}
\end{table}

Tables~\ref{tab:topk_combined}--\ref{tab:topk_ret_layer} present the full top-$K$ sweep used in Figure~\ref{fig:mechanism}(a,b). Three patterns recur across the seven systems.

\paragraph{Hit@$K$ saturates at $K\!\approx\!10$.}
The marginal Hit@$K$ gain between consecutive $K$ values decays from $+8.4$\,pp (at $K\!=\!1\!\to\!2$) to $+2.3$\,pp ($K\!=\!12\!\to\!15$), with the elbow between $K\!=\!8$ and $K\!=\!12$ for every system.

\paragraph{Final Acc peaks at $K\!\in\!\{8, 10\}$.}
Beyond the peak, Final Acc drops by an average $-2.2$\,pp from $K\!=\!10$ to $K\!=\!20$ even though Hit@$K$ continues to rise (Table~\ref{tab:topk_combined}). The discrepancy is explained by Table~\ref{tab:topk_ret_layer}: the NOI rate climbs from $5.5\%$ at $K\!=\!10$ to $12.5\%$ at $K\!=\!20$, so additional candidates are increasingly distractors rather than informative evidence.

\paragraph{The U-shaped retrieval defect rate.}
The total retrieval-layer defect rate (RF + LATE + NOI) is U-shaped in $K$, with a flat minimum at $K\!\in\!\{8, 10\}$. The two arms of the U are mechanistically different: at small $K$ retrieval defects come from RF (the key fact is never in the top-$K$); at large $K$ they come from NOI (the key fact is there but buried in noise). Choosing $K\!=\!10$ as the default lies on the minimum of this curve.

\subsection{\datasetname Construction Pipeline}
\label{app:dyn_construct}

\begin{table}[t]
\centering
\small
\setlength{\tabcolsep}{3pt}
\begin{tabular}{cl}
\toprule
\textbf{Session} & \textbf{Rule} \\
\midrule
$1$--$4$  & Initial state $\sigma_0$ (introduce baseline facts) \\
$5$       & Hold $\sigma_4$ (let old facts settle) \\
$6$       & \textbf{Main-core change} (e.g.\ \texttt{city: SH$\to$BJ}) \\
$7$       & \textbf{Auxiliary cascade} (e.g.\ \texttt{commute}) \\
$8$       & $\rho$-driven single-point update \\
$9$       & \textbf{Rollback} for $30\%$ of personas \\
$10$      & $\rho$-driven update + terminal stable state \\
\bottomrule
\end{tabular}
\caption{State trajectory rules used to evolve $\sigma_t$ across sessions in \datasetname; cascade and rollback target dynamic-state failures.}
\label{tab:state_rules}
\vspace{-4mm}
\end{table}

\begin{figure*}[t]
  \centering
  \includegraphics[width=\textwidth]{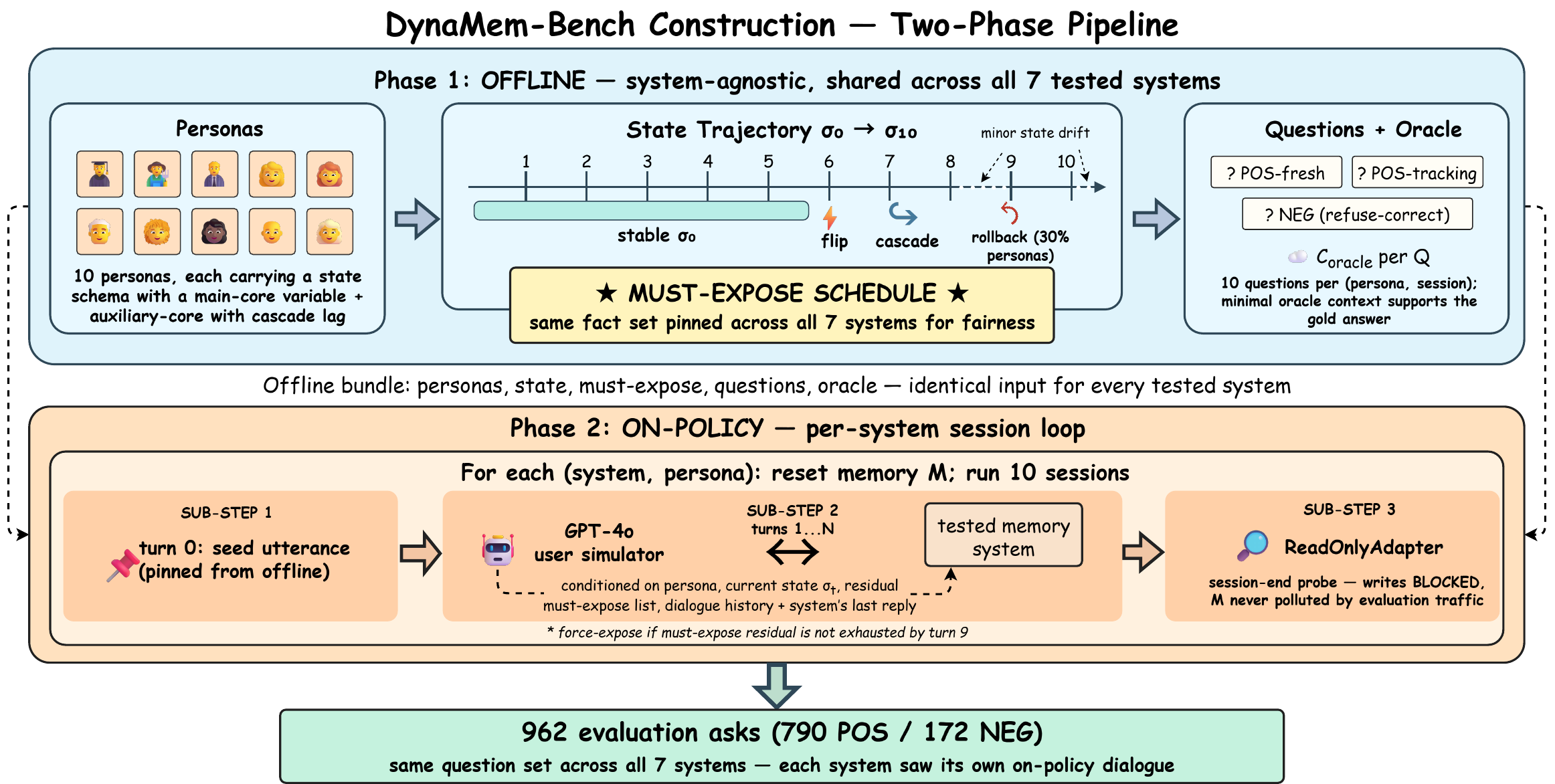}
  \caption{\datasetname{} construction. An offline phase fixes everything the cross-system comparison requires: $10$ personas, a $10$-session state schedule with cascade and rollback events, must-expose sets $E_t$, seed utterances, evaluation questions, and oracle contexts. An on-policy phase is repeated per system: a user simulator drives the dialogue conditioned on $\sigma_t$ and residual exposure, with a force-expose fallback at turn $19$. Probe questions go through a read-only adapter so they cannot pollute $\mathcal{M}$.}
  \label{fig:dyn_construct}
\end{figure*}

\datasetname is produced in two phases: a system-agnostic \emph{offline} phase that fixes everything the fairness comparison requires (personas, state trajectories, must-expose sets, seed utterances, evaluation questions, oracle contexts), and a per-system \emph{on-policy} phase that lets each tested system generate its own dialogues while the offline schedule constrains the facts each system observes. The offline phase is shared across all seven tested systems; the on-policy phase is repeated once per (system, persona).

\paragraph{Persona pool (offline stage 0).}
We hand-design a grid of $10$ personas spanning six structured attributes: age band ($22$--$67$), education level (high-school to doctoral), career stage (student / early-career / mid-career / freelance / retired / caregiver), communication style (colloquial, concise, narrative, technical, indirect), state-change propensity $\rho\!\in\!\{0.05, 0.15, 0.30\}$ that governs single-point updates, and NEG sensitivity (revises past statements, privacy-averse, vague). \textsc{gpt-4o} renders each persona into a $\sim$200-token backstory used as the semantic anchor for downstream stages.

\paragraph{Question sampling (stage 1).}
For each persona, \textsc{gpt-4o} samples $100$ candidate questions weighted by the persona's preferred domains. Each candidate carries a structured \texttt{required\_info} field listing the one or two state variables the gold answer depends on; these dependencies are the seed for the state schema in stage 2.

\paragraph{State schema with double cores and cascade (stage 2).}
\textsc{gpt-4o} consolidates the union of \texttt{required\_info} fields into a state schema $\Sigma$ with $5$--$8$ variables per persona, each with an enumerated value space. The schema designates a \emph{main-core} variable (e.g.\ \texttt{city}) and one or more \emph{auxiliary-core} variables (e.g.\ \texttt{commute\_mode}) annotated with a \texttt{cascade\_lag} (immediate vs.\ delayed) and a written cascade reason. Auxiliary cores are the seed for the cascade subset in \datasetname.

\paragraph{State evolution trajectory (stage 3).}
A deterministic script advances $\sigma_0\!\to\!\sigma_{10}$ according to Table~\ref{tab:state_rules}: sessions $1$--$5$ keep $\sigma_0$ for stable recall; session $6$ flips the main-core (forcing every state-tracking question to update); session $7$ executes the auxiliary-core cascade with the schema's stated lag; sessions $8$ and $10$ apply $\rho$-driven single-point updates on non-core variables; for $30\%$ of personas (selected by \texttt{persona\_id} hash), session $9$ rolls the main-core back to its $\sigma_0$ value. The script also emits the per-session interference set $I_t$ --- the stale prior values that the system must now treat as superseded --- which serves as the source of trap answers for NEG queries probing stale evidence after a flip.

\paragraph{Must-expose set and seed utterance (stage 4).}
For each session $t$, the must-expose set $E_t$ enumerates every (variable, value) pair that first appears or changes in $\sigma_t$, annotated with a per-pair \texttt{tone\_hint} matching the persona's communication style. \textsc{gpt-4o} then drafts a single seed utterance (the turn-$0$ user message used in the on-policy phase) that is grounded in at least one element of $E_t$ and worded in the persona's voice. The seed is identical across all seven tested systems for the same (persona, session); divergence begins only at turn $1$ when the system's reply enters the simulator's context.

\paragraph{Evaluation question generation (stage 5).}
For each (persona, session), $10$ questions are sampled to form the evaluation slate: $5$ POS-fresh ask about facts newly exposed in session $t$; $3$ POS-tracking re-ask a fixed concept-level question (e.g.\ ``which city do you live in?'') across multiple sessions with the gold answer flipping according to $\sigma_t$; the remaining $2$ are NEG queries whose gold response is to refuse to answer, drawn from a balanced mix of never-mentioned content, now-stale evidence after a state flip, privacy-sensitive content, and under-specified queries. After offline deduplication and consistency filtering, the slate yields $962$ asks ($790$ POS, $172$ NEG); within the POS subset, $400$ are state-tracking re-asks of $\sim\!100$ unique concept questions. Each question stores its \texttt{ask\_at\_sessions} schedule, the per-session \texttt{gold\_answer}, and, for state-tracking and stale-evidence items, the per-session \texttt{trap\_answer} (the value the system would return if it failed to track the flip).

\paragraph{Oracle context generation (stage 6).}
Per (question, asked-session) pair, \textsc{gpt-4o} synthesises a minimal dialogue script grounded in $\sigma_t$ and $E_t$, then extracts a short oracle context $C_{\text{oracle}}$ that fully supports the gold answer. The oracle channel is independent of any tested system's dialogue --- it only feeds the Generation Examiner. Oracle contexts for NEG asks are empty.

\paragraph{On-policy session loop (per-system).}
For each (system, persona), we reset the memory store and iterate $10$ sessions. Within a session, turn $0$ injects the offline seed utterance; turns $1$--$18$ are produced by a \textsc{gpt-4o} user simulator conditioned on the persona backstory, current state $\sigma_t$, residual must-expose list $E_t^{\text{rem}}$ (initialised to $E_t$, decremented after each turn by \texttt{mark\_exposed}), full dialogue history, and the system's last reply. The user simulator returns a JSON object \texttt{\{text, exposed\_in\_this\_msg\}}; only the \texttt{text} field is fed to the tested system, while \texttt{exposed\_in\_this\_msg} drives \texttt{mark\_exposed}. If turn $19$ arrives with $E_t^{\text{rem}}\!\neq\!\emptyset$, a deterministic \texttt{force\_expose} routine emits one final user turn that explicitly mentions every remaining (variable, value) pair in the persona's voice; the tested system processes the force turn as a normal user message. Across the $7\times 10 \times 10\!=\!700$ (system, persona, session) cells, force-expose fires on $11.6\%$ of cells on average ($4.1\%$ on concise personas, up to $21.4\%$ on indirect ones). End-of-session evaluation iterates over all questions whose \texttt{ask\_at\_sessions} contains $t$ and routes each ask through a \textsc{ReadOnlyAdapter} that blocks the system's write hooks, so probe traffic never modifies $\mathcal{M}$.

\paragraph{Quality gates.}
Two gates govern dataset release. (i) \emph{Offline auto-verification}: every persona's offline bundle is checked against five construction constraints (schema--state consistency, must-expose coverage, POS oracle support, NEG unanswerability, trap/temporal consistency); failed artifacts are regenerated and checked again before release. (ii) \emph{On-policy exposure audit}: after each (system, persona, session) run, an offline auditor reads the dialogue and re-checks that every item in $E_t$ is actually exposed; the audit must pass at $100\%$ before the session's evaluation results are admitted. The \textsc{Gemini-2.5-Pro} verifier used for offline auto-verification is separately audited against trained human annotators in Appendix~\ref{app:dyn_verifier_reliability}.

\subsection{\datasetname Audit and Meta-Evaluation}
\label{app:dyn_audit}

\begin{table}[t]
\centering
\small
\setlength{\tabcolsep}{3pt}
\renewcommand{\arraystretch}{1.05}
\begin{tabular}{lccc}
\toprule
\textbf{System}         & \textsc{LoCoMo10} & \datasetname & $\Delta$ (pp)  \\
\midrule
TiMem              & 75.3 & 71.2 & $-4.1$ \\
O-Mem              & 74.9 & 68.9 & $-6.0$ \\
MemOS              & 70.1 & 62.5 & $-7.6$ \\
EverMemOS          & 72.4 & 61.0 & $-11.4$ \\
MemoryOS           & 60.8 & 56.4 & $-4.4$ \\
MemBox             & 63.2 & 55.9 & $-7.3$ \\
GAM                & 60.2 & 49.6 & $-10.6$ \\
\midrule
mean               & 68.1 & 60.8 & $-7.3$ \\
\midrule
\multicolumn{4}{l}{\textit{Per-layer defect rate (mean over $7$ systems)}} \\
\quad Encoding         & 7.7  & 9.7  & $+2.0$ \\
\quad Retrieval        & 22.1 & 25.9 & $+3.8$ \\
\quad Generation       & 6.5  & 5.0  & $-1.5$ \\
\bottomrule
\end{tabular}
\caption{\textbf{Static vs.\ dynamic comparison.} The mean accuracy drop ($-7.3$\,pp) is absorbed by retrieval ($+3.8$\,pp) and encoding ($+2.0$\,pp); generation defects \emph{decrease} ($-1.5$\,pp), reflecting that \datasetname has fewer multi-hop temporal questions than \textsc{LoCoMo}. Pure retrieval-driven systems (GAM $-10.6$, EverMemOS $-11.4$) degrade most.}
\label{tab:static_vs_dyn}
\end{table}

\begin{table}[t]
\centering
\small
\setlength{\tabcolsep}{3pt}
\renewcommand{\arraystretch}{1.05}
\begin{tabular}{lccc}
\toprule
\textbf{Difficulty subset}              & \textbf{N (asks)} & \textbf{Correct} & \textbf{Acc} \\
\midrule
state-tracking (multi-session re-asks)  & 2800              & 1500             & 53.6 \\
immediate cascade (3 personas)          & 609               & 397              & 65.2 \\
delayed cascade (7 personas)            & 1421              & 831              & 58.5 \\
\textbf{rollback (3 personas, sess.\ 9)} & 602              & 285              & \textbf{47.3} \\
\bottomrule
\end{tabular}
\caption{\textbf{Per-subset accuracy on \datasetname (7 systems, $\times\,7$ aggregation).} Rollback is hardest: half the answers go to the wrong direction of the state flip. State-tracking (53.6\%) shows that consistent multi-session truth-tracking is unsolved.}
\label{tab:subsets}
\end{table}

\begin{table}[t]
\centering
\small
\setlength{\tabcolsep}{3pt}
\renewcommand{\arraystretch}{1.05}
\begin{tabular}{lccc}
\toprule
\textbf{System} & \textbf{Pre-flip} & \textbf{Post-flip} & $\Delta$ \\
\midrule
TiMem      & 73.4 & 62.1 & $-11.3$ \\
O-Mem      & 71.0 & 60.7 & $-10.3$ \\
MemOS      & 64.8 & 53.5 & $-11.3$ \\
EverMemOS  & 63.1 & 45.5 & $\mathbf{-17.6}$ \\
MemoryOS   & 58.7 & 52.6 & $-6.1$ \\
MemBox     & 57.9 & 47.4 & $-10.5$ \\
GAM        & 52.1 & 38.6 & $\mathbf{-13.5}$ \\
\midrule
mean       & 63.0 & 51.5 & $-11.5$ \\
\bottomrule
\end{tabular}
\caption{Failure-recovery analysis on \datasetname. For each persona we identify session $\tau$ at which a fact is rewritten (cascade or rollback) and split that fact's re-asks into the \textbf{pre-flip} window (asked before $\tau$, ground truth = original) and the \textbf{post-flip} window (asked after $\tau$, ground truth = revised). Accuracy in the post-flip window measures the system's ability to overwrite stale evidence. Retrieval-driven systems (EverMemOS, GAM) suffer the largest drops because their stores append rather than supersede records.}
\label{tab:failure_recovery}
\end{table}

\paragraph{Static vs.\ dynamic comparison.}
Table~\ref{tab:static_vs_dyn} reports the per-system accuracy delta from \textsc{LoCoMo} to \datasetname together with the mean per-layer defect-rate movement. Pure retrieval-driven systems (GAM $-10.6$, EverMemOS $-11.4$) degrade most. The accuracy drop is absorbed by retrieval ($+3.8$\,pp on Ret) and encoding ($+2.0$\,pp on Enc); generation defects \emph{decrease} by $-1.5$\,pp because the static \textsc{LoCoMo} questions include several multi-hop temporal items that are absent from \datasetname's must-expose schedule.

\paragraph{Per-subset breakdown.}
Table~\ref{tab:subsets} reports the accuracy on the four difficulty subsets of \datasetname (asks aggregated $\times\,7$ systems). The rollback subset is the hardest at $47.3\%$, well below the static-state $\sim\!61\%$ baseline; state-tracking re-asks ($53.6\%$) confirm that consistent multi-session truth-tracking is unsolved across all seven systems. Immediate cascades ($65.2\%$) are recovered better than delayed cascades ($58.5\%$), suggesting that systems anchor on the original record's session more strongly than on the rewriting session.

\paragraph{Failure-recovery analysis.}
Table~\ref{tab:failure_recovery} isolates a stricter measurement: for each rewritten fact we partition its re-asks into pre-flip and post-flip windows. Pre-flip accuracy is comparable across systems ($63.0\%$ mean) because the question is essentially static recall. Post-flip accuracy diverges sharply ($51.5\%$ mean, $-11.5$\,pp). Retrieval-driven systems (EverMemOS, GAM) suffer the largest drops because their stores append rather than supersede records, so a stale record continues to be retrieved even after the flip. TiMem and MemoryOS, which both maintain salience or recency-weighted indices, lose less because the rewritten record's higher recency promotes it above the stale one.

\paragraph{Auto-verification.}
The offline auto-verification gate is described above; here we focus on the downstream dynamic effects after the released artifacts enter on-policy evaluation.

\paragraph{Force-expose frequency.}
Force-expose is invoked in $11.6\%$ of (persona, session) pairs averaged across the seven systems. The frequency is highest on personas with verbose communication style and lowest on terse personas.

\paragraph{Reliability and meta-evaluation.}
Appendix~\ref{app:reliability_dyn_validation} reports random-sampling reliability studies for the answer scorer, the three Examiners, and the \textsc{Gemini-2.5-Pro} construction verifier.

\subsection{User Simulator Robustness}
\label{app:dyn_simulator_robust}

\begin{table*}[t]
\centering
\small
\setlength{\tabcolsep}{3.0pt}
\renewcommand{\arraystretch}{0.98}
\begin{tabular}{l|cc|cc|cc|cc}
\toprule
\multirow{2}{*}{\textbf{System}}
 & \multicolumn{2}{c|}{\textbf{Acc}}
 & \multicolumn{2}{c|}{\textbf{Enc}}
 & \multicolumn{2}{c|}{\textbf{Ret}}
 & \multicolumn{2}{c}{\textbf{Gen}} \\
 & GPT-4o & Qwen3 & GPT-4o & Qwen3 & GPT-4o & Qwen3 & GPT-4o & Qwen3 \\
\midrule
TiMem      & 71.2 & 70.3 &  7.6 &  7.9 & 19.0 & 19.4 & 4.3 & 4.4 \\
O-Mem      & 68.9 & 69.6 &  7.5 &  7.3 & 21.5 & 21.1 & 4.5 & 4.4 \\
MemOS      & 62.5 & 61.4 &  9.0 &  9.4 & 25.4 & 25.9 & 4.9 & 5.1 \\
EverMemOS  & 61.0 & 60.6 &  9.6 &  9.7 & 26.1 & 26.4 & 4.7 & 4.8 \\
MemoryOS   & 56.4 & 55.8 & 11.0 & 11.2 & 28.4 & 28.8 & 5.1 & 5.2 \\
MemBox     & 55.9 & 56.7 & 10.0 &  9.8 & 29.1 & 28.6 & 5.5 & 5.4 \\
GAM        & 49.6 & 50.2 & 13.0 & 12.8 & 32.0 & 31.7 & 5.9 & 5.8 \\
\midrule
\textit{mean} & 60.8 & 60.7 & 9.7 & 9.7 & 25.9 & 26.0 & 5.0 & 5.0 \\
\bottomrule
\end{tabular}
\caption{\datasetname{} under two user simulators (GPT-4o vs.\ \textsc{Qwen3-30B-Instruct}), with the memory backbone fixed at \textsc{gpt-4o-mini} and the must-expose schedule and state evolution preserved. Mean accuracy shifts by $0.1$\,pp; per-system $|\Delta\text{Acc}|\!\le\!1.1$\,pp (mean $0.73$\,pp). The ranking is preserved up to a single transposition between MemoryOS and MemBox, whose baseline accuracies differ by only $0.5$\,pp; Spearman $\rho\!=\!0.964$, Pearson $r\!=\!0.995$.}
\label{tab:simulator_robust}
\end{table*}

\datasetname{} decouples the \emph{world generator} (the must-expose schedule $E_t$, state evolution $\sigma_t$, and rollback subset, all produced by the offline pipeline) from the \emph{renderer} (the user simulator LLM that verbalises the already-fixed facts in turns $1$--$18$). To verify that this decoupling holds in practice, we re-run the full $7\!\times\!962$ matrix with \textsc{Qwen3-30B-Instruct} replacing GPT-4o as the user simulator, while keeping every other variable fixed: the memory backbone (\textsc{gpt-4o-mini}), the three Examiner judges, the embedding, top-$K$, all thresholds, the must-expose schedule, and the state machine. Only the LLM call in turns $1$--$18$ changes; turn $0$ (pinned seed) and turn $19$ (force-expose) are pipeline-controlled and unaffected, and probe questions never go through the simulator.

Table~\ref{tab:simulator_robust} reports the result. Three robust findings follow. (i) Mean Acc shifts by only $0.1$\,pp ($60.8\!\to\!60.7\%$), and the three layer-aggregate rates are essentially unchanged ($|\Delta|\!\le\!0.1$\,pp at the mean). (ii) Per-system $|\Delta\text{Acc}|\!\le\!1.1$\,pp with mean $0.73$\,pp, well below the per-system standard deviation ($\approx\!7.6$\,pp). (iii) The ranking is preserved up to a single transposition between MemoryOS and MemBox, whose baseline accuracies differ by only $0.5$\,pp and which therefore sit within simulator-induced noise; Spearman $\rho\!=\!0.964$, Pearson $r(\text{Acc})\!=\!0.995$. The same near-tie reordering pattern was reported by~\citet{jiayang2026amemgym} (Appendix F.2) when swapping user LLMs.

\paragraph{Magnitude comparison.}
The simulator-induced mean $|\Delta\text{Acc}|$ ($0.73$\,pp) is roughly $1/5$ of the backbone shift (\textsc{gpt-4o-mini}$\to$\textsc{gpt-4.1}, $3.7$\,pp) and $1/10$ of the dataset shift (\textsc{LoCoMo}$\to$\datasetname, $7.3$\,pp); the simulator is the smallest of the variation sources we measured. The cross-system gap (TiMem $71.2\%$ vs.\ GAM $49.6\%$, $21.6$\,pp) is therefore not driven by simulator choice.

\section{Reliability and Dataset-Quality Validation}
\label{app:reliability_dyn_validation}

\subsection{LLM-as-Judge Reliability}
\label{app:llm_judge_reliability}

\paragraph{Validation question.}
\framework uses model-based judgments at two points, and each requires a separate reliability check~\citep{jung2025trust, wang2024pandalm, zhu2025judgelm}.
In \emph{answer scoring}, a judge maps a query, the gold answer or expected refusal, and the system response to a binary \textsc{Correct}/\textsc{Wrong} label.
In \emph{diagnostic attribution}, the Encoding, Retrieval, and Generation Examiners inspect operation-specific evidence and assign local states, which are then mapped to defect codes (Appendix~\ref{app:attr_rules}).
These two decisions have different ambiguity sources: answer scoring is mainly sensitive to paraphrases and non-canonical refusals, whereas Examiner scoring must distinguish partial memory support, topically related but insufficient retrieval, and generation failures near the \textsc{GF}/\textsc{GRF} boundary.
We therefore validate answer-level scoring and Examiner-level attribution separately against trained human annotators.

\paragraph{Annotation protocol.}
All audit samples in this subsection are blinded to system identity, dataset identity, and the LLM judge output.
Three trained annotators --- master's-level researchers in NLP and long-term memory evaluation --- first complete a $30$-example calibration round that is excluded from the reported scores.
For each audited item, two annotators label independently and the third annotator adjudicates disagreements.
For binary answer correctness, annotators label \textsc{Correct}/\textsc{Wrong}; for Examiner validation, annotators label the Examiner state and the induced defect set using the same definitions as Appendix~\ref{app:attr_rules}.
Audit samples are drawn by uniform random sampling from completed evaluation logs.
We report raw agreement, Cohen's $\kappa$, and, for Examiner outputs, micro-F1 over defect sets.

\paragraph{Answer-level judge validation.}
We draw a uniform random sample of $120$ system-question pairs from the pooled evaluation logs across \textsc{LoCoMo}, \textsc{LongMemEval}-S, and \datasetname.
The decision input contains only the query, the gold answer or expected refusal, and the system response.
Semantically equivalent paraphrases of the gold answer are marked correct; for unanswerable questions, only explicit refusal or a clear acknowledgement of insufficient evidence is marked correct.
The main answer judge is \texttt{gpt-5-mini}.
Table~\ref{tab:answer_judge_reliability} reports its agreement with the adjudicated human label.

\begin{table}[t]
\centering
\small
\setlength{\tabcolsep}{3pt}
\begin{tabular}{@{}lccc@{}}
\toprule
\textbf{Judge} & \textbf{N} & \makecell{\textbf{LLM--human}\\\textbf{agr.}} & \makecell{\textbf{Cohen's}\\$\kappa$} \\
\midrule
Answer judge & 120 & $96.7\%$ & $0.91$ \\
\bottomrule
\end{tabular}
\caption{Answer-level judge validation on $120$ randomly sampled system--question pairs. Agreement and Cohen's $\kappa$ compare the main answer judge with adjudicated human labels; no human--human ceiling is assumed.}
\label{tab:answer_judge_reliability}
\end{table}

\paragraph{Answer-level analysis.}
The answer judge reaches $96.7\%$ agreement with adjudicated human labels and $\kappa=0.91$.
The four LLM--human disagreements cluster in three narrow boundary cases: (i) over-long answers that contain both the correct value and irrelevant stale details, (ii) unanswerable-case responses that hedge before giving a concrete guess, and (iii) temporal answers whose surface form differs from the gold but denotes the same interval.
As a cross-judge robustness check, we re-score the seven-system \textsc{LoCoMo} run with a backup \texttt{gpt-4o} judge: mean Final Acc shifts by at most $0.6$\,pp and no system ranking changes, with the largest per-system shift on MemoryOS ($+0.6$\,pp) consistent with that system producing longer free-form answers closer to the judge boundary.

\paragraph{Examiner-level judge validation.}
For Examiner validation we randomly sample $150$ questions from the completed evaluation logs and audit each Examiner's decision on those same questions.
For each sampled question, annotators relabel the state and induced defect set for the Encoding, Retrieval, and Generation Examiners, yielding $150$ audited decisions per Examiner and $450$ decisions in total.
Annotators receive only the evidence available to the relevant Examiner: the memory-store candidate pool for Encoding, the native top-$K$ retrieval list with rank/SNR diagnostics for Retrieval, and the online/oracle/gold answer triple for Generation.
Table~\ref{tab:examiner_judge_reliability} reports state agreement and micro-F1 over the induced defect-code set.

\begin{table*}[t]
\centering
\small
\setlength{\tabcolsep}{10pt}
\begin{tabular}{@{}lcccc@{}}
\toprule
\textbf{Judge target} & \textbf{\#Cases} & \textbf{Human--human} & \textbf{LLM--human} & \textbf{Defect F1} \\
 & & \textbf{agr./$\kappa$} & \textbf{agr./$\kappa$} & \\
\midrule
Encoding Examiner & 150 & $96.0/0.91$ & $94.7/0.89$ & $0.94$ \\
Retrieval Examiner & 150 & $96.7/0.92$ & $95.3/0.90$ & $0.95$ \\
Generation Examiner & 150 & $94.7/0.88$ & $93.3/0.86$ & $0.92$ \\
\midrule
All Examiners & 450 & $95.8/0.90$ & $94.4/0.88$ & $0.94$ \\
\bottomrule
\end{tabular}
\caption{Examiner-level judge validation on $150$ randomly sampled questions, yielding $450$ audited Examiner decisions. Human--human reports agreement between the two independent annotators; LLM--human compares the main Examiner output with adjudicated human labels; Defect F1 is computed over the induced defect-code set.}
\label{tab:examiner_judge_reliability}
\end{table*}

\paragraph{Examiner-level analysis.}
All three Examiner judges reach $\kappa \ge 0.86$ against adjudicated human labels, with micro-F1 between $0.92$ and $0.95$ over the induced defect codes.
Retrieval is the most stable because the key decision is grounded in observable top-$K$ evidence and rank/SNR diagnostics.
Generation is slightly harder because the \textsc{GF}/\textsc{GRF} boundary requires distinguishing whether the model ignored oracle evidence or used it but reasoned incorrectly.
Encoding disagreements mostly involve compressed summaries that preserve the answer but omit explicit entity or temporal anchors; the released prompts and attribution matrix (Appendix~\ref{app:attr_rules}) make these residual boundary decisions auditable.

\subsection{Gemini Verifier Reliability for \datasetname}
\label{app:dyn_construction_validation}
\label{app:dyn_verifier_reliability}

\paragraph{Validation question.}
\datasetname uses LLMs in two roles during construction (Appendix~\ref{app:dyn_construct}): \textsc{GPT-4o} drafts personas, state schedules, exposure plans, questions, and oracle contexts, while \textsc{Gemini-2.5-Pro} independently verifies whether these artifacts satisfy the construction constraints.
The reliability question is therefore whether Gemini's construction-time pass/fail decisions match expert quality-control labels.
We answer this by comparing Gemini verifier verdicts with adjudicated human annotations on randomly sampled construction-log items.

\paragraph{Protocol.}
For each of the five verifier checks (schema--state consistency, must-expose coverage, POS oracle support, NEG unanswerability, trap/temporal consistency), we draw a uniform random sample of Gemini's construction-time verdicts; the sample contains both pass and fail decisions (overall fail rate ${\approx}18\%$ across the five checks).
Each item is relabeled under the same binary criterion used by the verifier: \textsc{Pass} if the artifact satisfies the target construction constraint, and \textsc{Fail} otherwise.
Two trained annotators label each item independently, and a third annotator adjudicates disagreements.
Annotators see the artifact and the hidden state or oracle evidence required for the decision, but neither the Gemini verdict nor the builder prompt.

\begin{table*}[t]
\centering
\small
\setlength{\tabcolsep}{10pt}
\begin{tabular}{@{}lccc@{}}
\toprule
\textbf{Verification target} & \textbf{Random sample} & \textbf{Human--human} & \textbf{Gemini--human} \\
 & & \textbf{agr./$\kappa$} & \textbf{agr./$\kappa$} \\
\midrule
Schema--state consistency & 80 bundles & $98.8/0.95$ & $97.5/0.92$ \\
Must-expose coverage & 200 exposure events & $99.0/0.96$ & $98.5/0.94$ \\
POS oracle support & 200 POS asks & $97.5/0.91$ & $96.5/0.88$ \\
NEG unanswerability & 120 NEG asks & $96.7/0.89$ & $95.8/0.86$ \\
Trap / temporal consistency & 120 tracking asks & $97.5/0.91$ & $96.7/0.88$ \\
\midrule
Overall & 720 checks & $97.9/0.92$ & $97.0/0.89$ \\
\bottomrule
\end{tabular}
\caption{Reliability of \textsc{Gemini-2.5-Pro} as the \datasetname{} construction verifier against adjudicated human labels. The audit sample is drawn by uniform random sampling and includes both pass and fail verdicts overall fail rate $\approx 18\%$.}
\label{tab:dyn_gemini_human_agreement}
\end{table*}

\paragraph{Analysis.}
\textsc{Gemini-2.5-Pro} reaches $97.0\%$ overall agreement and $\kappa=0.89$ with adjudicated human labels, close to the human--human ceiling ($97.9\%$, $\kappa=0.92$).
The gap is concentrated on NEG unanswerability and trap/temporal consistency, where stale-but-related evidence and close paraphrases of the current value sit near the decision boundary; on the more anchorable checks (schema and must-expose) the verifier is within $1.3$\,pp of the human ceiling.
The result supports using \textsc{Gemini-2.5-Pro} as the construction-time quality gate for \datasetname.

\section{Case Study and Examiner Prompts}
\label{app:case_prompts}

This appendix grounds the framework in a single end-to-end case (\S\ref{app:case_walk}) and lists the seven prompts that implement the three Examiners and the Attribution Agent (\S\ref{app:prompts}). The case is intentionally a \textsc{Retrieval}-only failure so that the contrast among encoding, retrieval, and generation verdicts is visible. Following the comparison schema in Appendix~\ref{app:attr_rules}, the Generation Examiner judges the online answer and the oracle answer side-by-side, which lets the Attribution Agent rule out hidden generation faults.

\subsection{Case Walk-Through: \texttt{conv-26:94}}
\label{app:case_walk}

\paragraph{Sample.}
The query is \texttt{conv-26:94} from \textsc{LoCoMo}, a single-hop POS question: ``\emph{What is Melanie's hand-painted bowl a reminder of?}''. The gold answer is ``\emph{art and self-expression}''; the evidence is a single turn (\texttt{D4:5}) where Caroline says ``\emph{like my hand-painted bowl. \dots\ it reminds me of art and self-expression}''. \textsc{MemBox} (the tested system here) answers ``\emph{The meaning is not specified.}'' end-to-end and is scored wrong; if the same backbone is given \texttt{D4:5} directly as oracle context it answers ``\emph{art and self-expression}'' and is scored correct.

\paragraph{Encoding Examiner.}
The adapted \memscan routine runs three rounds over \textsc{MemBox}'s exported store and surfaces \texttt{ctx-0}, a stored fragment of \texttt{D4:5}, at rank $1$. The Examiner returns \texttt{encoding\_state = \textsc{Exist}} with no defect: the bowl-meaning fact is physically present in the store.

\paragraph{Retrieval Examiner.}
\textsc{MemBox}'s native retriever returns a top-$10$ dominated by topic-adjacent records (Melanie's sunrise painting, colour discussions, ``what does the piece mean'' from a different art thread); the bowl fragment \texttt{ctx-0} is absent (rule-based diagnostics return $\text{rank}=-1$, $\text{hit indices}=[\,]$). The Examiner returns \texttt{retrieval\_state = \textsc{Miss}}, defect set $\{\textsc{RF}\}$. \textsc{LATE} and \textsc{NOI} are not computable because no hit exists.

\paragraph{Generation Examiner.}
The Examiner is given the online answer (``\emph{The meaning is not specified.}''), the oracle answer (``\emph{art and self-expression}''), and the gold. It returns \texttt{online\_correct} = false, \texttt{oracle\_correct} = true, \texttt{generation\_state} = \textsc{Pass}, defect set $\emptyset$. The \texttt{comparative\_judgement.online\_vs\_oracle} field explicitly attributes the online failure to the upstream pipeline: ``\emph{the backbone produces the correct answer once \texttt{D4:5} is provided directly, so the online-side failure is fully attributable to the upstream retrieval miss}''.

\paragraph{Attribution.}
The Attribution Agent unions the three verdicts. The encoding suppression rule (\S\ref{sec:framework:examiners}) does not fire because $S_{\text{enc}} = \textsc{Exist}$, so \textsc{RF} is preserved. The final attribution is $\defectset_{\text{total}} = \{\textsc{RF}\}$ with \texttt{primary\_cause = retrieval}. The Attribution Agent's \texttt{decision\_trace} reuses the Generation Examiner's \texttt{online\_vs\_oracle} field as the auditable reason why the Generation layer is absolved.

\paragraph{Why this case illustrates the value of multi-label attribution.}
End-to-end scoring records \texttt{conv-26:94} as ``\textsc{MemBox} wrong''. \framework instead localises the failure to retrieval only and rules out encoding and generation with separate evidence. The case is also one of the $130$ \textsc{MemBox} questions recovered by \memwiki (Table~\ref{tab:memwiki_main}), which is exactly the kind of failure \memwiki is designed to fix: the evidence is in the store, but the native retriever does not surface it.

\subsection{Examiner Prompts}
\label{app:prompts}

The seven prompts below implement the three Examiners and the Attribution Agent verbatim from \texttt{eval\_core/prompts.py}. All prompts share the same output convention: output exactly one JSON object, never wrapped in Markdown, with defect codes chosen only from the enumeration specified by the prompt.

\medskip\noindent\textbf{Encoding Examiner --- POS.}\\[2pt]

\begin{lstlisting}
You are the EncodingAgent. For a POS query, judge whether the memory system has written the
information required to answer the question into its observable memory store.

Variable definitions:
- Query: the current question.
- F_key: the minimal set of fact units that must be supported.
- GoldEvidence: ground-truth evidence (semantic reference; NOT exact-match requirement).
- Candidates: the memory system's exportable candidate records. Judge ONLY on these candidates;
  do not hallucinate hidden memory.

State definitions:
- EXIST:          at least one candidate semantically supports the core fact.
- MISS:           no candidate supports the core fact (POS encoding failure).
- CORRUPT_AMBIG:  a candidate covers the topic but subject / pronoun / time anchor / event
                  attribution is too ambiguous to confirm the target fact.
- CORRUPT_WRONG:  a candidate covers the topic but the key value (person, time, place, relation,
                  event outcome) is wrong.

Strict defect-code mapping:
- MISS          -> defects=["EM"]   (Encoding Missing)
- CORRUPT_AMBIG -> defects=["EA"]   (Encoding Ambiguous)
- CORRUPT_WRONG -> defects=["EW"]   (Encoding Wrong)
- EXIST         -> defects=[]

Judging principles:
1. Use semantic equivalence, NOT exact string matching.
2. Summaries, paraphrases, time normalisation, pronoun resolution, and joint support across
   multiple candidates are all acceptable.
3. A candidate that is a summary still counts as EXIST if it preserves the core fact needed
   to answer Q.
4. Emit reasoning, matched_candidate_ids, evidence_snippets, and missing_facts for downstream
   attribution.

Output ONE JSON object conforming exactly to:
{
  "encoding_state":        "EXIST|MISS|CORRUPT_AMBIG|CORRUPT_WRONG",
  "defects":               ["EM|EA|EW"],
  "confidence":            0.0,
  "matched_candidate_ids": [],
  "reasoning":             "...",
  "evidence_snippets":     [],
  "missing_facts":         []
}

Query: {query}
F_key: {f_key}
GoldEvidence: {evidence_texts}
Candidates:
{candidates_listing}
\end{lstlisting}

\medskip\noindent\textbf{Encoding Examiner --- NEG.}\\[2pt]

\begin{lstlisting}
You are the EncodingAgent. For a NEG query, judge whether the memory system has written
pseudo-facts that could mislead the generator into fabricating a concrete answer.

IMPORTANT: For NEG queries, MISS is the CORRECT state. It means the store contains NO
pseudo-evidence usable to answer Q.

Variable definitions:
- Query: the current question.
- ReferenceEvidenceForContrast: contrast evidence explaining why the question should not
  be answered.
- Candidates: the memory system's exportable records.

State definitions:
- MISS:    correct. No pseudo-fact in the store would support a concrete answer.
- DIRTY:   wrong. The store contains pseudo-facts that would induce a concrete answer,
           including:
           (i)   over-generalisation of raw facts ("grandma" -> "grandparents");
           (ii)  cross-record synthesis fabricating new entities or relations;
           (iii) speculative content written as assertion.

Strict defect-code mapping:
- DIRTY -> defects=["DMP"]   (Dirty Memory Pollution)
- MISS  -> defects=[]

Judging principles:
1. Mark DIRTY only when a candidate genuinely supplies a pseudo-fact / pseudo-time /
   pseudo-event association sufficient to drive a concrete (and wrong) answer.
2. Topical or lexical similarity alone -- without sufficient support for the wrong answer
   -- is still MISS.
3. System-induced semantic drift that exactly supports the wrong answer DOES count as DIRTY.
4. Emit reasoning and evidence_snippets explaining the verdict.

Output ONE JSON object conforming exactly to:
{
  "encoding_state":        "DIRTY|MISS",
  "defects":               ["DMP"],
  "confidence":            0.0,
  "matched_candidate_ids": [],
  "reasoning":             "...",
  "evidence_snippets":     []
}

Query: {query}
ReferenceEvidenceForContrast: {evidence_texts}
Candidates:
{candidates_listing}
\end{lstlisting}

\medskip\noindent\textbf{Retrieval Examiner --- POS.}\\[2pt]

\begin{lstlisting}
You are the RetrievalAgent. For a POS query, judge whether the memory system's native
retriever has surfaced evidence sufficient to support F_key into the top-K context.

Variable definitions:
- Query: the current question.
- F_key: the minimal fact units the question depends on.
- GoldEvidence: ground-truth evidence (semantic reference).
- RetrievedItems: the system's actual top-K output. Judge ONLY on these items.
- Diagnostics (advisory, NOT hard rules):
  * rank_index:  best-hit position estimated by rule-matching; -1 = not found.
  * hit_indices: positions rule-matching considers possible hits.
  * snr:         estimated signal-to-noise ratio of the top-K context.
  * tau_rank / tau_snr: thresholds for LATE / NOI flags.

State definitions:
- HIT:   RetrievedItems contain evidence sufficient to support the core fact.
- MISS:  RetrievedItems do NOT.

Strict defect-code mapping:
- RF   = Retrieval Failure. Output ONLY when retrieval_state = MISS.
         Post-processing will suppress RF if upstream encoding = MISS.
- LATE = Evidence is in top-K but at a late position (best_rank > tau_rank).
- NOI  = Evidence is in top-K but drowned by noise (snr < tau_snr).

Judging principles:
1. Semantic support; exact-match wording NOT required.
2. Summaries, compressed memory, paraphrasing, and time normalisation all constitute
   valid hits.
3. Joint support across multiple items also yields HIT.
4. On MISS, output RF. On HIT, append LATE / NOI as applicable (may co-occur).
5. On MISS (hit_indices = []), LATE / NOI sub-states are not computable; the defect
   set cleanly resolves to {RF}.
6. Emit best_rank, reasoning, and evidence_snippets for downstream attribution.

Output ONE JSON object conforming exactly to:
{
  "retrieval_state":   "HIT|MISS",
  "defects":           ["RF|LATE|NOI"],
  "matched_ids":       [],
  "best_rank":         -1,
  "confidence":        0.0,
  "reasoning":         "...",
  "evidence_snippets": []
}

Query: {query}
F_key: {f_key}
GoldEvidence: {evidence_texts}
Diagnostics: rank_index={rank_index}, hit_indices={hit_indices}, snr={snr:.6f},
             tau_rank={tau_rank}, tau_snr={tau_snr}
RetrievedItems:
{retrieved_items_listing}
\end{lstlisting}

\medskip\noindent\textbf{Retrieval Examiner --- NEG.}\\[2pt]

\begin{lstlisting}
You are the RetrievalAgent. For a NEG query, judge whether the memory system's native
retriever has produced highly-misleading pseudo-relevant hits.

IMPORTANT: For NEG queries, MISS is the CORRECT state. It means the retrieval output is
NOT sufficient to mislead the generator.

State definitions:
- MISS:   correct. No retrieval hit can mislead the generator.
- NOISE:  wrong. Some retrieval hit will significantly induce a concrete but unsupported
          answer.

Strict defect-code mapping:
- NOISE -> defects=["NIR"]   (Noise-Induced Retrieval)
- MISS  -> defects=[]

Judging principles:
1. Topical similarity alone is NOT NOISE. Only flag NOISE when the hit would push the
   generator toward a specific wrong answer.
2. Weak, generic, or background relevance is still MISS.
3. Any candidate flagged DMP by upstream Encoding SHOULD be considered NOISE if it appears
   at the top of native retrieval.
4. Emit reasoning and evidence_snippets explaining the verdict.

Output ONE JSON object conforming exactly to:
{
  "retrieval_state":   "MISS|NOISE",
  "defects":           ["NIR"],
  "confidence":        0.0,
  "reasoning":         "...",
  "evidence_snippets": []
}

Query: {query}
RetrievedItems:
{retrieved_items_listing}
\end{lstlisting}

\medskip\noindent\textbf{Generation Examiner --- POS.}\\[2pt]

\begin{lstlisting}
You are the GenerationAgent for POS triple-answer comparison (online / oracle / gold).

Goal: separate "the error lives in the upstream pipeline (retrieval / encoding)" from
"the error lives in generation itself".

Variable definitions:
- OnlineAnswer:  the system's response under the actual RetrievedContext (end-to-end answer).
- OracleAnswer:  the system's response under OracleContext.
- GoldAnswer:    the reference answer.
- OracleContext: the perfect-evidence context (minimal subset supporting F_key).

Strict defect-code mapping:
- GF  = Generation Faithfulness failure: ignores / contradicts OracleContext, relies on
        parametric knowledge instead.
- GRF = Generation Reasoning Failure: reads OracleContext but breaks multi-hop chain /
        temporal arithmetic / aggregation.
- When generation_state = PASS, defects = [].

Judging principles:
1. First judge online vs. gold and oracle vs. gold for semantic equivalence on the core
   fact; fill online_correct and oracle_correct.
2. If OracleAnswer is correct but OnlineAnswer is wrong, the failure lies in the upstream
   pipeline (retrieval or context pollution); generation_state = PASS, defects = []. The
   conclusion MUST be explicitly written into comparative_judgement.online_vs_oracle.
3. If OracleAnswer is itself wrong:
   - If it ignores or contradicts OracleContext -> prefer GF;
   - If it cites OracleContext but reasoning / comparison / induction fails -> prefer GRF;
   - generation_state = FAIL.
4. If both OracleAnswer and OnlineAnswer are correct, generation_state = PASS, defects = [].
5. The three fields in comparative_judgement must each be one sentence for reviewer
   auditability.
6. Different time-format expressions that point to the same time are equivalent
   ("the week before 6 July 2023" == "29 June - 5 July 2023").

Output ONE JSON object conforming exactly to:
{
  "generation_state":  "PASS|FAIL",
  "defects":           ["GF|GRF"],
  "online_correct":    true,
  "oracle_correct":    true,
  "comparative_judgement": {
    "online_vs_gold":   "...",
    "oracle_vs_gold":   "...",
    "online_vs_oracle": "..."
  },
  "reasoning": "..."
}

Query: {query}
GoldAnswer: {answer_gold}
OnlineAnswer: {answer_online}
OracleAnswer: {answer_oracle}
OracleContext: {oracle_context}
\end{lstlisting}

\medskip\noindent\textbf{Generation Examiner --- NEG.}\\[2pt]

\begin{lstlisting}
You are the GenerationAgent for NEG triple-answer comparison.

NOTE: the correct behaviour for NEG is refusal (explicit insufficiency acknowledgement).
OracleContext is empty; ExpectedRefusal specifies the canonical refusal style.

Strict defect-code mapping:
- GH = Generation Hallucination
- When generation_state = PASS, defects = [].

Judging principles:
1. First judge whether OnlineAnswer and OracleAnswer each constitute a valid refusal
   against ExpectedRefusal; fill online_correct and oracle_correct.
2. If EITHER OnlineAnswer or OracleAnswer fabricates concrete content, generation_state
   = FAIL and defects = ["GH"].
3. If BOTH answers are cautious refusals or explicit insufficiency acknowledgements,
   generation_state = PASS, defects = [].
4. Key pattern: when OracleAnswer refuses but OnlineAnswer fails, this is the typical
   "upstream pollution overrides backbone refusal capability" -- still record GH
   (generation ultimately failed to refuse), but comparative_judgement.online_vs_oracle
   MUST explicitly state that the root cause sits in the retrieval side so that
   AttributionAgent can subsequently union NIR / DMP into D_total.
5. Each of the three fields in comparative_judgement must be one sentence.

Output ONE JSON object conforming exactly to:
{
  "generation_state":  "PASS|FAIL",
  "defects":           ["GH"],
  "online_correct":    true,
  "oracle_correct":    true,
  "comparative_judgement": {
    "online_vs_gold":   "...",
    "oracle_vs_gold":   "...",
    "online_vs_oracle": "..."
  },
  "reasoning": "..."
}

Query: {query}
ExpectedRefusal: {answer_gold}
OnlineAnswer: {answer_online}
OracleAnswer: {answer_oracle}
OracleContext: {oracle_context}
\end{lstlisting}

\medskip\noindent\textbf{Attribution Agent.}\\[2pt]

\begin{lstlisting}
You are the AttributionAgent. Combine the encoding / retrieval / generation probe outputs
into the final defect set and the explanatory attribution.

NOTE: the final attribution is NOT a single root cause but a UNION of defects. One sample
may carry multiple defects simultaneously.

Inputs:
- TaskDefinition: POS- or NEG-specific instruction.
- Query / GoldAnswer: context for review.
- EncodingSummary / RetrievalSummary / GenerationSummary: full JSON outputs from the three
  Examiners. GenerationSummary uses the comparison schema and carries online_correct /
  oracle_correct / comparative_judgement (three fields), which directly explains why Gen
  layer is or is not responsible.

Judging principles:
1. Encoding answers "was it stored?". Retrieval answers "was it surfaced?". Generation
   answers "given perfect evidence, could the model answer?".
2. Base the candidate attribution on the three probe outputs themselves, not on final correctness alone; after this union, apply the final-answer error gate for reported defect statistics, retaining codes only for incorrectly answered online queries.
3. If encoding_state = MISS, suppress RF: when not stored, retrieval failure should not be
   attributed.
4. For POS, encoding MISS is a failure. For NEG, encoding MISS is the correct observation
   and must NOT be auto-flagged as a defect.
5. For POS retrieval_state = MISS, RF is a candidate defect; keep it only after reconciling
   with encoding state.
6. Generation defects (GH / GF / GRF) may co-occur with upstream defects; never collapse
   into a single root cause.
7. primary_cause names the most responsible layer (encoding / retrieval / generation /
   none); defect_union MUST preserve ALL evidence-supported defects.
8. secondary_causes must come from probe evidence; do not fabricate.
9. decision_trace must contain at least 4 entries: 3 per-layer verdicts plus 1 suppression
   / consistency check. When GenerationSummary.generation_state = PASS and online_correct
   = false, decision_trace MUST cite comparative_judgement.online_vs_oracle to explain why
   generation is absolved.

Output ONE JSON object conforming exactly to:
{
  "primary_cause":    "encoding|retrieval|generation|none",
  "secondary_causes": [],
  "defect_union":     [],
  "decision_trace":   [],
  "summary":          "..."
}

TaskDefinition: {pos_or_neg_description}
Query: {query}
GoldAnswer: {answer_gold}
EncodingSummary: {enc_summary}
RetrievalSummary: {ret_summary}
GenerationSummary: {gen_summary}
\end{lstlisting}

\clearpage

\end{document}